\documentclass[runningheads]{llncs}
\usepackage[a4paper,left=3cm,right=3cm]{geometry}

\usepackage[T1]{fontenc}
\usepackage{graphicx}
\usepackage{amsfonts}
\usepackage{amsmath}
\usepackage{algorithm}
\usepackage{algpseudocode}
\usepackage{censor,caption}
\usepackage{multirow}
 
\usepackage[T1]{fontenc}
\usepackage{graphicx,verbatim}
\usepackage{color}

\usepackage{booktabs}
\usepackage{array}
\usepackage{longtable}
\usepackage{tabularx}
\usepackage{threeparttable}
\usepackage{caption}
\usepackage{xurl}
\usepackage{amssymb} 

\usepackage[colorlinks=true,linkcolor=blue,citecolor=blue,urlcolor=blue]{hyperref}
\makeatletter
\AtBeginDocument{\let\dpd@doi\doi\renewcommand{\doi}[1]{\in@{https://doi.org/}{#1}\ifin@\href{\detokenize{#1}}{\url{#1}}\else\dpd@doi{#1}\fi}}
\makeatother

\begin{document}
\title{The pretraining domain outweighs the training objective in setting the privacy-utility trade-off of differentially private medical image analysis}

\author{
Soroosh Tayebi Arasteh\inst{1,2,3,4}$^{\ast}$ \and
Mina Farajiamiri\inst{5} \and
Mahshad Lotfinia\inst{1,2,6} \and
Behrus Hinrichs-Puladi\inst{7,8} \and
Jonas Bienzeisler\inst{8} \and
Mohamed Alhaskir\inst{8} \and
Mirabela Rusu\inst{3,4} \and
Christiane Kuhl\inst{2} \and
Sven Nebelung\inst{1,2} \and
Daniel Truhn\inst{1,2}
}

\institute{
Lab for AI in Medicine, RWTH Aachen University, Aachen, Germany \and
Department of Diagnostic and Interventional Radiology, University Hospital RWTH Aachen, Aachen, Germany \and
Department of Urology, Stanford University, Stanford, CA, USA \and
Department of Radiology, Stanford University, Stanford, CA, USA \and
School of Business and Economics, RWTH Aachen University, Aachen, Germany \and
Pattern Recognition Lab, Friedrich-Alexander-Universit\"at Erlangen-N\"urnberg, Erlangen, Germany \and
Department of Oral and Maxillofacial Surgery, University Hospital RWTH Aachen, Aachen, Germany \and
Institute of Medical Informatics, University Hospital RWTH Aachen, Aachen, Germany
}

\maketitle 
{\footnotesize
\noindent$^{\ast}$Correspondence to: Soroosh Tayebi Arasteh (\email{soroosh.arasteh@rwth-aachen.de})
}

\begin{abstract}
Differential privacy limits how much each training image can influence a medical imaging model but lowers its diagnostic accuracy. Pretrained weights reduce this loss. It is unknown whether the pretraining objective or domain improves accuracy more. Here we train chest radiograph classifiers under differential privacy from five starting points on 591{,}884 radiographs from five datasets. We show that supervised pretraining on chest radiographs ranks first in 24 of 25 combinations of dataset and privacy setting. Measured by the area under the receiver operating characteristic curve, its lead over supervised pretraining on natural images is 2.6\% without privacy and 15.3\% at the strictest budget. From that starting point, changing the domain improves this area 2.2 to 3.7 times more than changing the objective. Private pretraining on chest radiographs lowers this area by about 5\% under privacy but outperforms the three starting points without patient data at the strictest budget. Under privacy, supervised chest radiograph pretraining leads in the worst-performing demographic subgroup.
\end{abstract}

\section*{Introduction}

Deep learning models for medical imaging improve with the size and diversity of their training data. However, medical images are among the most sensitive patient data~\cite{Kaissis2021EndToEnd}. Training images can also be reconstructed from the released weights of a model~\cite{Balle2022Reconstructing}. Differential privacy (DP) bounds how much any single training example can influence a model~\cite{Dwork2006Differential}. For deep networks, it is realized by differentially private stochastic gradient descent (DP-SGD), which clips per-sample gradients and adds calibrated noise~\cite{Abadi2016Deep}. DP-SGD has been applied in medical imaging~\cite{Ziller2021Medical,TayebiArasteh2024Preserving}, but it lowers diagnostic accuracy. A recent review of 74 studies found that models remain clinically useful at moderate budgets near $\epsilon=10$~\cite{Mohammadi2026Differential}. The same review reported that accuracy commonly decreases at strict budgets near $\epsilon=1$ and decreases most in small and heterogeneous cohorts. Reconciling privacy with accuracy remains a main obstacle to deploying private medical artificial intelligence (AI)~\cite{Ziller2024Reconciling}.

Among the methods that reduce this loss, the initialization has been among the most effective. Beginning private training from pretrained weights instead of from scratch recovers much of the lost accuracy, an effect documented on natural images~\cite{De2022Unlocking,Kurakin2022Toward}, in language~\cite{Li2022Large}, and in radiography~\cite{TayebiArasteh2024Preserving}. Self-supervised pretraining produces representations that transfer more broadly and are more robust than those of supervised pretraining~\cite{Caron2021Emerging,Oquab2023DINOv2,Hendrycks2019Using}. It has proven useful in medical imaging~\cite{Krishnan2022SelfSupervised,Azizi2023Robust,TayebiArasteh2024Enhancing,Lotfinia2025Boosting}. Features from self-supervised pretraining also improve accuracy under DP~\cite{Tramer2021Differentially}. Recent DINOv3 models provide self-supervised pretrained weights for convolutional backbones~\cite{Simeoni2025DINOv3}. Separately, pretraining on the target modality is long-standing practice in chest radiography, although its benefit over generic pretraining is contested without privacy~\cite{Ke2021CheXtransfer,Khader2022Artificial,Mei2022RadImageNet}. Foundation models for chest radiography are pretrained on radiographs, either with self-supervision~\cite{PerezGarcia2025Exploring,Yao2025EVAX} or with expert labels~\cite{Ma2025Fully}. It has not been established whether the objective or the domain drives performance under privacy.

\begin{figure*}[p]
\centering
\footnotesize
\includegraphics[width=\textwidth]{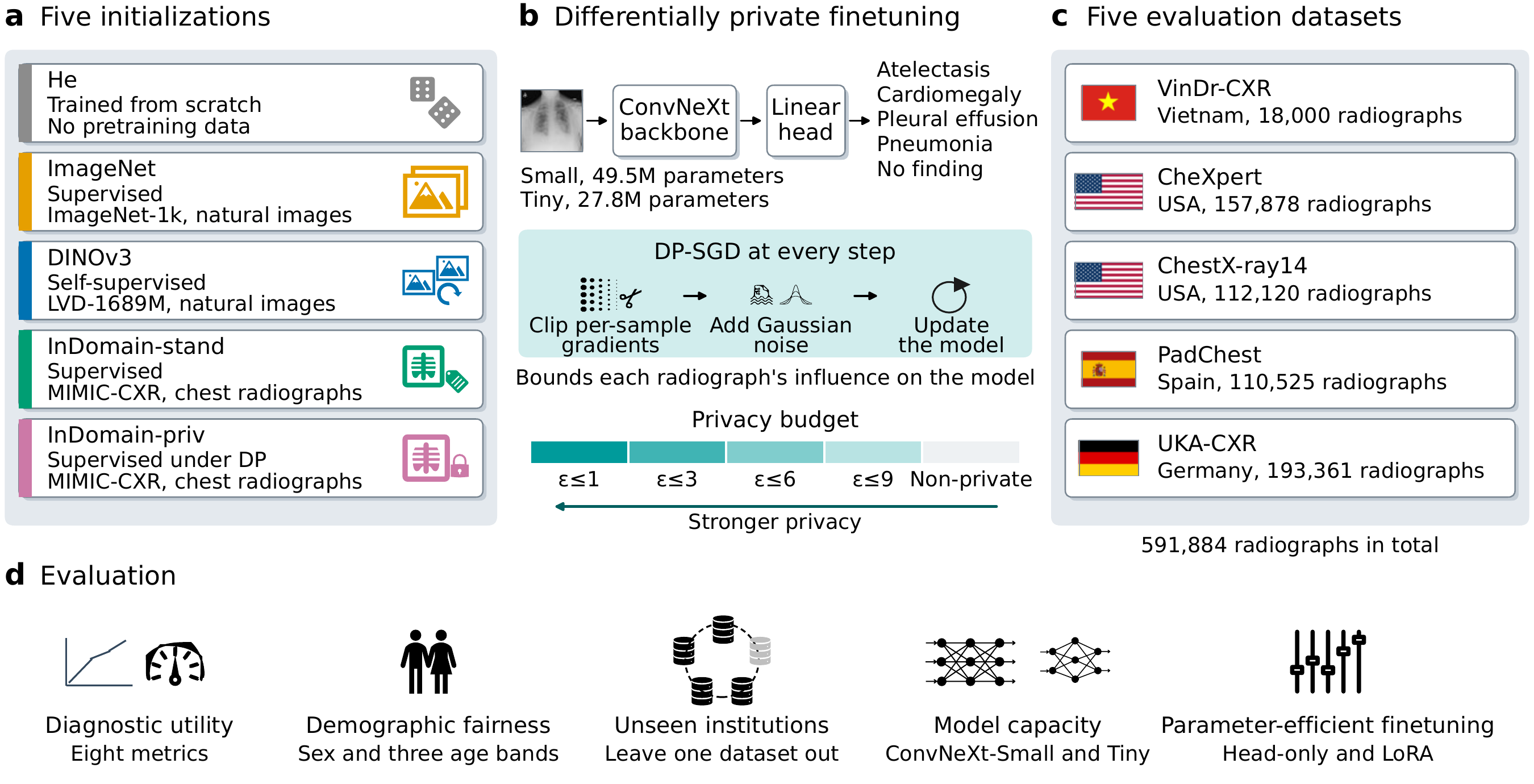}
\caption{Study design. Colors mark the five initializations. \textbf{a}, The initializations of the ConvNeXt~\cite{Liu2022ConvNet} classifier. He initialization~\cite{He2015Delving} trains the network from scratch. ImageNet~\cite{Deng2009ImageNet} was pretrained with labels on ImageNet-1k, and DINOv3~\cite{Simeoni2025DINOv3} was pretrained without labels on LVD-1689M. Both datasets contain natural images. InDomain-stand and InDomain-priv were pretrained with labels on the 153{,}255 frontal radiographs of the MIMIC-CXR~\cite{Johnson2019MIMIC} training split. InDomain-priv was pretrained under differential privacy. \textbf{b}, Every initialization was finetuned to predict five findings from a chest radiograph, once without privacy and once at each of four privacy budgets. Private finetuning used DP-SGD~\cite{Abadi2016Deep}, which clips the gradient of every training radiograph and adds Gaussian noise at every step. The ConvNeXt backbone has 49.5 million parameters (Small) or 27.8 million (Tiny). \textbf{c}, The five datasets, VinDr-CXR~\cite{Nguyen2022VinDr}, CheXpert~\cite{Irvin2019CheXpert}, ChestX-ray14~\cite{Wang2017ChestXray8}, PadChest~\cite{Bustos2020PadChest}, and UKA-CXR~\cite{Khader2022Artificial}, each with its country and the number of radiographs used. \textbf{d}, Evaluations of diagnostic utility in eight metrics, fairness across sex and three age bands, generalization to an institution held out of training, model capacity with ConvNeXt-Small and ConvNeXt-Tiny, and the parameter-efficient schemes of head-only finetuning and LoRA~\cite{Hu2022LoRA}. The radiograph in \textbf{b} is from ChestX-ray14. DP, differential privacy; DP-SGD, differentially private stochastic gradient descent; ImageNet-1k, the 1{,}000-class ImageNet dataset; LoRA, low-rank adaptation; LVD-1689M, the 1.7 billion web images of DINOv3 pretraining; M, million; $\epsilon$, privacy budget.}
\label{fig:overview}
\end{figure*}

Here we contrast each property while holding the other fixed. Using a fixed ConvNeXt backbone~\cite{Liu2022ConvNet} and a common DP-SGD pipeline, we compare five initializations that differ in the pretraining objective and the pretraining domain (Fig.~\ref{fig:overview}). Three of them are random initialization with the He scheme~\cite{He2015Delving}, supervised ImageNet pretraining~\cite{Deng2009ImageNet}, and self-supervised DINOv3 pretraining on natural images~\cite{Simeoni2025DINOv3}. The other two are supervised pretraining on chest radiographs from MIMIC-CXR~\cite{Johnson2019MIMIC}, once conventionally and once under DP-SGD. Supplementary Table~\ref{stab:glossary} defines every term of the study. Pretraining under DP-SGD addresses a second problem. Private finetuning conventionally treats the pretraining corpus as public. This assumption is hard to defend when the pretraining corpus consists of patient data~\cite{Tramer2024Position}. Private pretraining extends the guarantee to the pretraining data and has been studied on hundreds of millions of web images~\cite{Yu2024ViP}. We evaluate every initialization at four privacy budgets and without privacy on 591{,}884 radiographs from five datasets~\cite{Nguyen2022VinDr,Irvin2019CheXpert,Wang2017ChestXray8,Bustos2020PadChest,Khader2022Artificial}. These datasets come from four countries and from routine and intensive care. Further experiments compare three finetuning schemes, including low-rank adaptation (LoRA)~\cite{Hu2022LoRA}, and two model capacities. We also test generalization to unseen institutions by holding out one dataset at a time. Finally, we measure performance within demographic subgroups, in which the effect of privacy is known to be uneven~\cite{Bagdasaryan2019Differential}. To our knowledge, this is the first study to compare the pretraining objective with the pretraining domain under DP at this scale, and the first in medical imaging to measure the downstream cost of protecting the pretraining corpus.

We expected the noise added to every gradient to penalize a model in proportion to how much it still has to learn. An initialization that already encodes the target modality should then lose the least. We test this expectation on diagnostic accuracy, on the privacy spent before a model converges, on transfer to institutions not seen in training, and on the demographic subgroup with the lowest performance. Pretraining on chest radiographs leads in every dataset at every privacy budget. The pretraining domain contributes more to this lead than the pretraining objective. Making that pretraining private keeps it ahead of generic pretraining in most private settings. Under privacy, the domain-matched initialization also converges with less privacy spent and gives the highest performance in the worst-performing subgroup. Supplementary Table~\ref{stab:findings} summarizes the findings. These findings indicate that encoders pretrained on the target modality, including encoders pretrained under DP, may be a more effective route to private medical AI than generic self-supervised encoders.

\section*{Results}

Metrics are percentages, and the percent sign is dropped. The main metric is the area under the receiver operating characteristic curve (AUROC) averaged over the five findings shared by all datasets, written macro AUROC. A difference between two metrics is given in percent with the metric named. Each value is the mean $\pm$ standard deviation over bootstrap resamples of the test set, with the 95\% confidence interval (CI) in brackets. Initializations are compared on macro AUROC with a two-sided paired bootstrap test at the 0.05 level, without correction for multiple comparisons. Private runs used four privacy budgets, written $\epsilon\le1$, $\epsilon\le3$, $\epsilon\le6$, and $\epsilon\le9$, beside a non-private reference. Rounded to one decimal, every private run spent at most the upper end of its privacy budget. Every final model comes from its converged epoch, the epoch with the highest validation macro AUROC. The achieved $\epsilon$ of a run is the budget spent up to that epoch (Supplementary Tables~\ref{stab:accounting} and~\ref{stab:steps}).

\subsection*{In-domain pretraining sets the privacy-utility frontier}

\begin{figure}[p]
\centering
\footnotesize
\includegraphics[width=\textwidth]{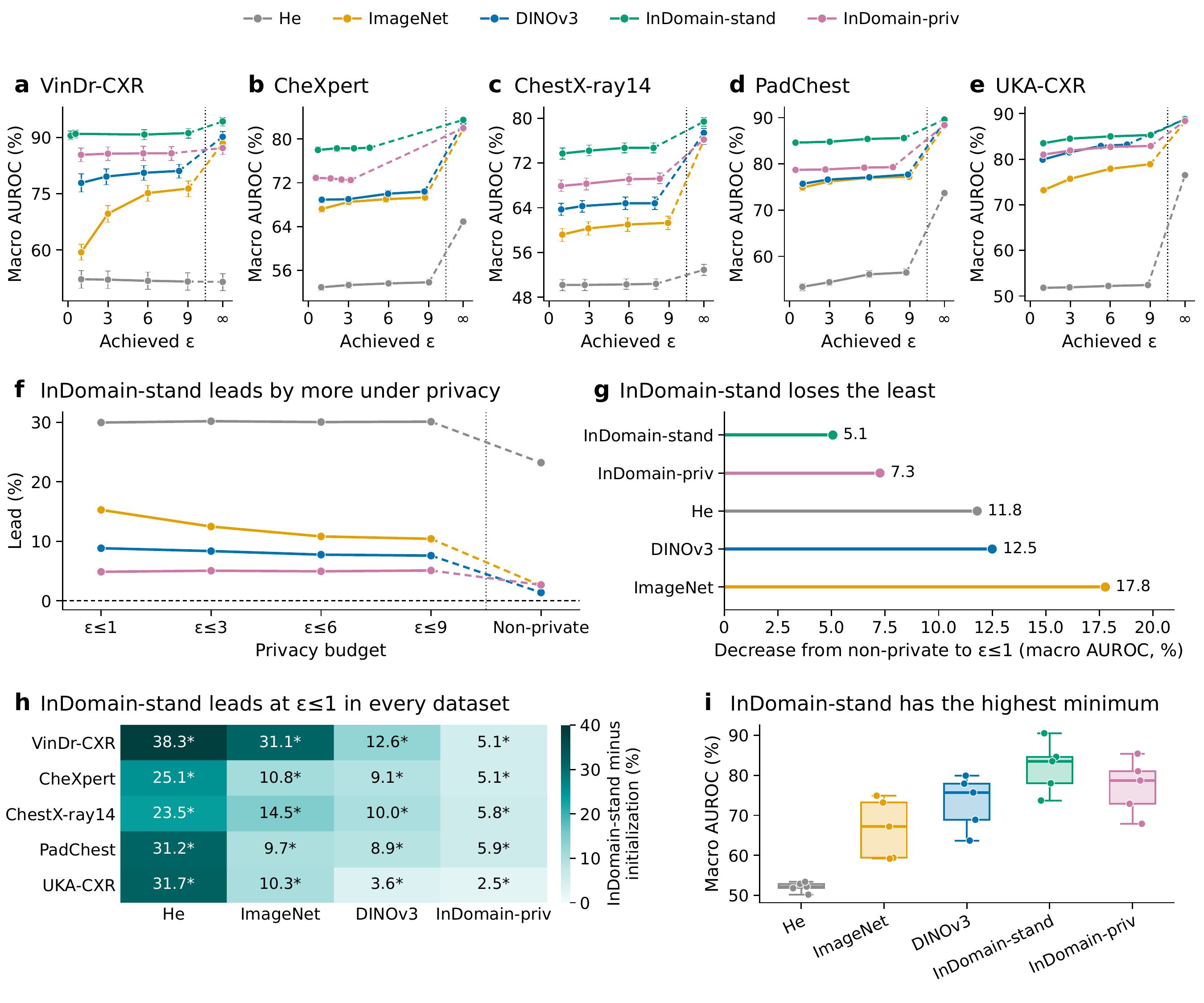}
\caption{Macro AUROC of ConvNeXt-Small~\cite{Liu2022ConvNet} classifiers finetuned with differentially private stochastic gradient descent~\cite{Abadi2016Deep} from five initializations at four privacy budgets and without privacy on the test sets of five datasets ($n = 120{,}068$ test radiographs). Colors mark the initializations. In \textbf{a}--\textbf{e}, a marker is the mean of 10{,}000 bootstrap resamples of the test set, and the error bar is the 95\% confidence interval, from the 2.5th to the 97.5th percentile. No other panel has an error bar. Counts per panel: 25 runs (\textbf{a}--\textbf{e}), values that are means over the five datasets (\textbf{f}, \textbf{g}), 20 comparisons (\textbf{h}), and 5 datasets per initialization (\textbf{i}). \textbf{a}--\textbf{e}, Macro AUROC on the test sets of VinDr-CXR~\cite{Nguyen2022VinDr} ($n = 3{,}000$), CheXpert~\cite{Irvin2019CheXpert} ($n = 29{,}321$), ChestX-ray14~\cite{Wang2017ChestXray8} ($n = 25{,}596$), PadChest~\cite{Bustos2020PadChest} ($n = 22{,}045$), and UKA-CXR~\cite{Khader2022Artificial} ($n = 40{,}106$) against the achieved $\epsilon$ of each run, with dashed lines to the non-private reference at $\infty$ beyond the dotted break. \textbf{f}, Macro AUROC of InDomain-stand minus that of each other initialization. \textbf{g}, Decrease in macro AUROC from non-private training to $\epsilon\le1$. \textbf{h}, InDomain-stand minus each other initialization at $\epsilon\le1$. An asterisk marks $p<0.05$ in a two-sided paired bootstrap test with 10{,}000 resamples and no correction for multiple comparisons. \textbf{i}, Macro AUROC at $\epsilon\le1$. A dot is one dataset, and a box spans the 25th to the 75th percentile with the median as a line and whiskers to the smallest and largest value. AUROC, area under the receiver operating characteristic curve; $\epsilon$, privacy budget; He, He initialization~\cite{He2015Delving}; ImageNet, supervised pretraining on ImageNet~\cite{Deng2009ImageNet}; DINOv3, self-supervised DINOv3 pretraining~\cite{Simeoni2025DINOv3}; InDomain-stand, supervised pretraining on MIMIC-CXR~\cite{Johnson2019MIMIC}; InDomain-priv, differentially private pretraining on MIMIC-CXR.}
\label{fig:frontier}
\end{figure}

He initialization trains the network from scratch. Its starting weights are drawn from the He (Kaiming) normal distribution instead of being taken from a trained model. ImageNet was pretrained on natural images with labels, and DINOv3 was pretrained on natural images without labels. InDomain-stand was pretrained with labels on chest radiographs from MIMIC-CXR, and InDomain-priv was pretrained in the same way under DP. On five chest radiograph datasets, we finetuned a ConvNeXt-Small classifier from each of these five initializations at four privacy budgets and without privacy. InDomain-stand had the highest macro AUROC in 24 of the 25 combinations of dataset and privacy setting (Table~\ref{tab:frontier}, Fig.~\ref{fig:frontier}a--e). The exception was UKA-CXR without privacy, where InDomain-stand and DINOv3 had the same macro AUROC (88.8 each, $p=0.562$).

\begin{table}[p]
\centering
\footnotesize
\caption{Macro AUROC of five initializations at four privacy budgets and without privacy on the test sets of five datasets ($n = 120{,}068$ test radiographs). Every model is a ConvNeXt-Small~\cite{Liu2022ConvNet} finetuned with differentially private stochastic gradient descent~\cite{Abadi2016Deep}, except in the rows without privacy. Each entry is the AUROC averaged over the five findings in percent, as the mean $\pm$ standard deviation [95\% confidence interval] of 10{,}000 bootstrap resamples. Rows are grouped by privacy setting and ordered from no pretraining to in-domain pretraining. The test sets have 3{,}000 (VinDr-CXR~\cite{Nguyen2022VinDr}), 29{,}321 (CheXpert~\cite{Irvin2019CheXpert}), 25{,}596 (ChestX-ray14~\cite{Wang2017ChestXray8}), 22{,}045 (PadChest~\cite{Bustos2020PadChest}), and 40{,}106 (UKA-CXR~\cite{Khader2022Artificial}) radiographs. AUROC, area under the receiver operating characteristic curve; $\epsilon$, privacy budget; He, He initialization~\cite{He2015Delving}; ImageNet, supervised pretraining on ImageNet~\cite{Deng2009ImageNet}; DINOv3, self-supervised DINOv3 pretraining~\cite{Simeoni2025DINOv3}; InDomain-stand and InDomain-priv, supervised pretraining on MIMIC-CXR~\cite{Johnson2019MIMIC} without and with differential privacy.}
\label{tab:frontier}
\scriptsize
\setlength{\tabcolsep}{5pt}
\renewcommand{\arraystretch}{0.81}
\begin{tabular}{lccccc}
\toprule
Initialization & VinDr-CXR & CheXpert & ChestX-ray14 & PadChest & UKA-CXR \\
\midrule
\multicolumn{6}{l}{\textbf{$\epsilon\le1$}} \\
He & \begin{tabular}[t]{@{}c@{}}52.2 $\pm$ 1.2\\{[}49.8, 54.5{]}\end{tabular} & \begin{tabular}[t]{@{}c@{}}52.9 $\pm$ 0.3\\{[}52.3, 53.4{]}\end{tabular} & \begin{tabular}[t]{@{}c@{}}50.2 $\pm$ 0.5\\{[}49.2, 51.2{]}\end{tabular} & \begin{tabular}[t]{@{}c@{}}53.4 $\pm$ 0.4\\{[}52.6, 54.1{]}\end{tabular} & \begin{tabular}[t]{@{}c@{}}51.8 $\pm$ 0.3\\{[}51.2, 52.5{]}\end{tabular} \\
ImageNet & \begin{tabular}[t]{@{}c@{}}59.4 $\pm$ 1.0\\{[}57.3, 61.5{]}\end{tabular} & \begin{tabular}[t]{@{}c@{}}67.2 $\pm$ 0.3\\{[}66.6, 67.8{]}\end{tabular} & \begin{tabular}[t]{@{}c@{}}59.2 $\pm$ 0.6\\{[}58.0, 60.3{]}\end{tabular} & \begin{tabular}[t]{@{}c@{}}74.9 $\pm$ 0.3\\{[}74.2, 75.5{]}\end{tabular} & \begin{tabular}[t]{@{}c@{}}73.2 $\pm$ 0.3\\{[}72.6, 73.8{]}\end{tabular} \\
DINOv3 & \begin{tabular}[t]{@{}c@{}}77.9 $\pm$ 1.2\\{[}75.5, 80.3{]}\end{tabular} & \begin{tabular}[t]{@{}c@{}}68.9 $\pm$ 0.3\\{[}68.3, 69.4{]}\end{tabular} & \begin{tabular}[t]{@{}c@{}}63.7 $\pm$ 0.5\\{[}62.6, 64.8{]}\end{tabular} & \begin{tabular}[t]{@{}c@{}}75.7 $\pm$ 0.3\\{[}75.0, 76.3{]}\end{tabular} & \begin{tabular}[t]{@{}c@{}}79.9 $\pm$ 0.2\\{[}79.4, 80.3{]}\end{tabular} \\
InDomain-stand & \begin{tabular}[t]{@{}c@{}}90.5 $\pm$ 0.6\\{[}89.4, 91.7{]}\end{tabular} & \begin{tabular}[t]{@{}c@{}}78.0 $\pm$ 0.3\\{[}77.5, 78.6{]}\end{tabular} & \begin{tabular}[t]{@{}c@{}}73.7 $\pm$ 0.5\\{[}72.7, 74.7{]}\end{tabular} & \begin{tabular}[t]{@{}c@{}}84.6 $\pm$ 0.3\\{[}84.1, 85.1{]}\end{tabular} & \begin{tabular}[t]{@{}c@{}}83.5 $\pm$ 0.2\\{[}83.1, 83.9{]}\end{tabular} \\
InDomain-priv & \begin{tabular}[t]{@{}c@{}}85.4 $\pm$ 0.9\\{[}83.6, 87.2{]}\end{tabular} & \begin{tabular}[t]{@{}c@{}}72.9 $\pm$ 0.3\\{[}72.3, 73.5{]}\end{tabular} & \begin{tabular}[t]{@{}c@{}}67.9 $\pm$ 0.5\\{[}66.9, 69.0{]}\end{tabular} & \begin{tabular}[t]{@{}c@{}}78.7 $\pm$ 0.3\\{[}78.1, 79.3{]}\end{tabular} & \begin{tabular}[t]{@{}c@{}}81.0 $\pm$ 0.2\\{[}80.5, 81.4{]}\end{tabular} \\
\addlinespace
\multicolumn{6}{l}{\textbf{$\epsilon\le3$}} \\
He & \begin{tabular}[t]{@{}c@{}}52.1 $\pm$ 1.2\\{[}49.7, 54.4{]}\end{tabular} & \begin{tabular}[t]{@{}c@{}}53.3 $\pm$ 0.3\\{[}52.7, 53.9{]}\end{tabular} & \begin{tabular}[t]{@{}c@{}}50.2 $\pm$ 0.5\\{[}49.2, 51.2{]}\end{tabular} & \begin{tabular}[t]{@{}c@{}}54.4 $\pm$ 0.4\\{[}53.6, 55.1{]}\end{tabular} & \begin{tabular}[t]{@{}c@{}}51.9 $\pm$ 0.3\\{[}51.3, 52.6{]}\end{tabular} \\
ImageNet & \begin{tabular}[t]{@{}c@{}}69.7 $\pm$ 1.1\\{[}67.4, 71.9{]}\end{tabular} & \begin{tabular}[t]{@{}c@{}}68.5 $\pm$ 0.3\\{[}67.9, 69.1{]}\end{tabular} & \begin{tabular}[t]{@{}c@{}}60.3 $\pm$ 0.6\\{[}59.1, 61.5{]}\end{tabular} & \begin{tabular}[t]{@{}c@{}}76.2 $\pm$ 0.3\\{[}75.5, 76.8{]}\end{tabular} & \begin{tabular}[t]{@{}c@{}}75.7 $\pm$ 0.3\\{[}75.1, 76.2{]}\end{tabular} \\
DINOv3 & \begin{tabular}[t]{@{}c@{}}79.6 $\pm$ 1.1\\{[}77.4, 81.6{]}\end{tabular} & \begin{tabular}[t]{@{}c@{}}69.0 $\pm$ 0.3\\{[}68.5, 69.6{]}\end{tabular} & \begin{tabular}[t]{@{}c@{}}64.3 $\pm$ 0.5\\{[}63.2, 65.3{]}\end{tabular} & \begin{tabular}[t]{@{}c@{}}76.6 $\pm$ 0.3\\{[}76.0, 77.2{]}\end{tabular} & \begin{tabular}[t]{@{}c@{}}81.5 $\pm$ 0.2\\{[}81.1, 81.9{]}\end{tabular} \\
InDomain-stand & \begin{tabular}[t]{@{}c@{}}91.0 $\pm$ 0.5\\{[}89.9, 92.0{]}\end{tabular} & \begin{tabular}[t]{@{}c@{}}78.3 $\pm$ 0.3\\{[}77.7, 78.8{]}\end{tabular} & \begin{tabular}[t]{@{}c@{}}74.2 $\pm$ 0.5\\{[}73.3, 75.2{]}\end{tabular} & \begin{tabular}[t]{@{}c@{}}84.8 $\pm$ 0.3\\{[}84.3, 85.3{]}\end{tabular} & \begin{tabular}[t]{@{}c@{}}84.5 $\pm$ 0.2\\{[}84.1, 84.9{]}\end{tabular} \\
InDomain-priv & \begin{tabular}[t]{@{}c@{}}85.7 $\pm$ 0.9\\{[}83.9, 87.5{]}\end{tabular} & \begin{tabular}[t]{@{}c@{}}72.8 $\pm$ 0.3\\{[}72.2, 73.3{]}\end{tabular} & \begin{tabular}[t]{@{}c@{}}68.3 $\pm$ 0.5\\{[}67.2, 69.3{]}\end{tabular} & \begin{tabular}[t]{@{}c@{}}78.8 $\pm$ 0.3\\{[}78.2, 79.3{]}\end{tabular} & \begin{tabular}[t]{@{}c@{}}81.9 $\pm$ 0.2\\{[}81.4, 82.3{]}\end{tabular} \\
\addlinespace
\multicolumn{6}{l}{\textbf{$\epsilon\le6$}} \\
He & \begin{tabular}[t]{@{}c@{}}51.8 $\pm$ 1.2\\{[}49.5, 54.1{]}\end{tabular} & \begin{tabular}[t]{@{}c@{}}53.6 $\pm$ 0.3\\{[}53.1, 54.2{]}\end{tabular} & \begin{tabular}[t]{@{}c@{}}50.3 $\pm$ 0.5\\{[}49.4, 51.3{]}\end{tabular} & \begin{tabular}[t]{@{}c@{}}56.1 $\pm$ 0.4\\{[}55.4, 56.9{]}\end{tabular} & \begin{tabular}[t]{@{}c@{}}52.2 $\pm$ 0.3\\{[}51.5, 52.9{]}\end{tabular} \\
ImageNet & \begin{tabular}[t]{@{}c@{}}75.2 $\pm$ 1.1\\{[}73.0, 77.3{]}\end{tabular} & \begin{tabular}[t]{@{}c@{}}69.0 $\pm$ 0.3\\{[}68.4, 69.6{]}\end{tabular} & \begin{tabular}[t]{@{}c@{}}61.0 $\pm$ 0.6\\{[}59.8, 62.2{]}\end{tabular} & \begin{tabular}[t]{@{}c@{}}77.0 $\pm$ 0.3\\{[}76.3, 77.6{]}\end{tabular} & \begin{tabular}[t]{@{}c@{}}77.9 $\pm$ 0.3\\{[}77.4, 78.4{]}\end{tabular} \\
DINOv3 & \begin{tabular}[t]{@{}c@{}}80.6 $\pm$ 1.0\\{[}78.6, 82.5{]}\end{tabular} & \begin{tabular}[t]{@{}c@{}}70.0 $\pm$ 0.3\\{[}69.4, 70.6{]}\end{tabular} & \begin{tabular}[t]{@{}c@{}}64.8 $\pm$ 0.5\\{[}63.7, 65.9{]}\end{tabular} & \begin{tabular}[t]{@{}c@{}}77.1 $\pm$ 0.3\\{[}76.5, 77.7{]}\end{tabular} & \begin{tabular}[t]{@{}c@{}}82.9 $\pm$ 0.2\\{[}82.5, 83.3{]}\end{tabular} \\
InDomain-stand & \begin{tabular}[t]{@{}c@{}}90.8 $\pm$ 0.7\\{[}89.4, 92.1{]}\end{tabular} & \begin{tabular}[t]{@{}c@{}}78.3 $\pm$ 0.3\\{[}77.8, 78.9{]}\end{tabular} & \begin{tabular}[t]{@{}c@{}}74.7 $\pm$ 0.5\\{[}73.8, 75.6{]}\end{tabular} & \begin{tabular}[t]{@{}c@{}}85.4 $\pm$ 0.3\\{[}84.9, 85.9{]}\end{tabular} & \begin{tabular}[t]{@{}c@{}}85.0 $\pm$ 0.2\\{[}84.6, 85.3{]}\end{tabular} \\
InDomain-priv & \begin{tabular}[t]{@{}c@{}}85.8 $\pm$ 0.9\\{[}83.9, 87.7{]}\end{tabular} & \begin{tabular}[t]{@{}c@{}}72.6 $\pm$ 0.3\\{[}72.0, 73.2{]}\end{tabular} & \begin{tabular}[t]{@{}c@{}}69.1 $\pm$ 0.5\\{[}68.1, 70.1{]}\end{tabular} & \begin{tabular}[t]{@{}c@{}}79.2 $\pm$ 0.3\\{[}78.6, 79.8{]}\end{tabular} & \begin{tabular}[t]{@{}c@{}}82.7 $\pm$ 0.2\\{[}82.3, 83.1{]}\end{tabular} \\
\addlinespace
\multicolumn{6}{l}{\textbf{$\epsilon\le9$}} \\
He & \begin{tabular}[t]{@{}c@{}}51.6 $\pm$ 1.2\\{[}49.3, 53.9{]}\end{tabular} & \begin{tabular}[t]{@{}c@{}}53.8 $\pm$ 0.3\\{[}53.2, 54.4{]}\end{tabular} & \begin{tabular}[t]{@{}c@{}}50.4 $\pm$ 0.5\\{[}49.4, 51.3{]}\end{tabular} & \begin{tabular}[t]{@{}c@{}}56.5 $\pm$ 0.4\\{[}55.8, 57.3{]}\end{tabular} & \begin{tabular}[t]{@{}c@{}}52.4 $\pm$ 0.3\\{[}51.8, 53.0{]}\end{tabular} \\
ImageNet & \begin{tabular}[t]{@{}c@{}}76.4 $\pm$ 1.1\\{[}74.2, 78.4{]}\end{tabular} & \begin{tabular}[t]{@{}c@{}}69.3 $\pm$ 0.3\\{[}68.7, 69.8{]}\end{tabular} & \begin{tabular}[t]{@{}c@{}}61.3 $\pm$ 0.6\\{[}60.1, 62.5{]}\end{tabular} & \begin{tabular}[t]{@{}c@{}}77.2 $\pm$ 0.3\\{[}76.6, 77.9{]}\end{tabular} & \begin{tabular}[t]{@{}c@{}}78.9 $\pm$ 0.2\\{[}78.4, 79.4{]}\end{tabular} \\
DINOv3 & \begin{tabular}[t]{@{}c@{}}81.1 $\pm$ 0.9\\{[}79.2, 82.9{]}\end{tabular} & \begin{tabular}[t]{@{}c@{}}70.4 $\pm$ 0.3\\{[}69.8, 71.0{]}\end{tabular} & \begin{tabular}[t]{@{}c@{}}64.8 $\pm$ 0.5\\{[}63.7, 65.9{]}\end{tabular} & \begin{tabular}[t]{@{}c@{}}77.7 $\pm$ 0.3\\{[}77.1, 78.3{]}\end{tabular} & \begin{tabular}[t]{@{}c@{}}83.2 $\pm$ 0.2\\{[}82.8, 83.6{]}\end{tabular} \\
InDomain-stand & \begin{tabular}[t]{@{}c@{}}91.2 $\pm$ 0.7\\{[}89.8, 92.4{]}\end{tabular} & \begin{tabular}[t]{@{}c@{}}78.4 $\pm$ 0.3\\{[}77.9, 78.9{]}\end{tabular} & \begin{tabular}[t]{@{}c@{}}74.7 $\pm$ 0.5\\{[}73.8, 75.6{]}\end{tabular} & \begin{tabular}[t]{@{}c@{}}85.6 $\pm$ 0.3\\{[}85.1, 86.1{]}\end{tabular} & \begin{tabular}[t]{@{}c@{}}85.3 $\pm$ 0.2\\{[}84.9, 85.6{]}\end{tabular} \\
InDomain-priv & \begin{tabular}[t]{@{}c@{}}85.8 $\pm$ 1.0\\{[}83.9, 87.6{]}\end{tabular} & \begin{tabular}[t]{@{}c@{}}72.5 $\pm$ 0.3\\{[}71.9, 73.1{]}\end{tabular} & \begin{tabular}[t]{@{}c@{}}69.2 $\pm$ 0.5\\{[}68.3, 70.2{]}\end{tabular} & \begin{tabular}[t]{@{}c@{}}79.3 $\pm$ 0.3\\{[}78.7, 79.9{]}\end{tabular} & \begin{tabular}[t]{@{}c@{}}82.9 $\pm$ 0.2\\{[}82.5, 83.3{]}\end{tabular} \\
\addlinespace
\multicolumn{6}{l}{\textbf{Non-private}} \\
He & \begin{tabular}[t]{@{}c@{}}51.5 $\pm$ 1.2\\{[}49.2, 53.7{]}\end{tabular} & \begin{tabular}[t]{@{}c@{}}64.9 $\pm$ 0.3\\{[}64.3, 65.5{]}\end{tabular} & \begin{tabular}[t]{@{}c@{}}52.9 $\pm$ 0.5\\{[}51.9, 53.9{]}\end{tabular} & \begin{tabular}[t]{@{}c@{}}73.7 $\pm$ 0.4\\{[}73.0, 74.3{]}\end{tabular} & \begin{tabular}[t]{@{}c@{}}76.5 $\pm$ 0.3\\{[}75.9, 77.0{]}\end{tabular} \\
ImageNet & \begin{tabular}[t]{@{}c@{}}88.5 $\pm$ 0.8\\{[}86.9, 90.0{]}\end{tabular} & \begin{tabular}[t]{@{}c@{}}81.8 $\pm$ 0.3\\{[}81.3, 82.3{]}\end{tabular} & \begin{tabular}[t]{@{}c@{}}76.1 $\pm$ 0.4\\{[}75.3, 76.9{]}\end{tabular} & \begin{tabular}[t]{@{}c@{}}88.2 $\pm$ 0.2\\{[}87.7, 88.6{]}\end{tabular} & \begin{tabular}[t]{@{}c@{}}88.2 $\pm$ 0.2\\{[}87.9, 88.6{]}\end{tabular} \\
DINOv3 & \begin{tabular}[t]{@{}c@{}}90.2 $\pm$ 0.7\\{[}88.7, 91.6{]}\end{tabular} & \begin{tabular}[t]{@{}c@{}}83.0 $\pm$ 0.2\\{[}82.5, 83.4{]}\end{tabular} & \begin{tabular}[t]{@{}c@{}}77.4 $\pm$ 0.4\\{[}76.6, 78.2{]}\end{tabular} & \begin{tabular}[t]{@{}c@{}}89.2 $\pm$ 0.2\\{[}88.7, 89.6{]}\end{tabular} & \begin{tabular}[t]{@{}c@{}}88.8 $\pm$ 0.2\\{[}88.5, 89.1{]}\end{tabular} \\
InDomain-stand & \begin{tabular}[t]{@{}c@{}}94.3 $\pm$ 0.5\\{[}93.1, 95.3{]}\end{tabular} & \begin{tabular}[t]{@{}c@{}}83.5 $\pm$ 0.2\\{[}83.0, 83.9{]}\end{tabular} & \begin{tabular}[t]{@{}c@{}}79.4 $\pm$ 0.4\\{[}78.6, 80.1{]}\end{tabular} & \begin{tabular}[t]{@{}c@{}}89.6 $\pm$ 0.2\\{[}89.2, 90.0{]}\end{tabular} & \begin{tabular}[t]{@{}c@{}}88.8 $\pm$ 0.2\\{[}88.4, 89.1{]}\end{tabular} \\
InDomain-priv & \begin{tabular}[t]{@{}c@{}}87.2 $\pm$ 0.8\\{[}85.5, 88.8{]}\end{tabular} & \begin{tabular}[t]{@{}c@{}}82.0 $\pm$ 0.3\\{[}81.5, 82.5{]}\end{tabular} & \begin{tabular}[t]{@{}c@{}}76.2 $\pm$ 0.5\\{[}75.3, 77.1{]}\end{tabular} & \begin{tabular}[t]{@{}c@{}}88.4 $\pm$ 0.2\\{[}87.9, 88.8{]}\end{tabular} & \begin{tabular}[t]{@{}c@{}}88.4 $\pm$ 0.2\\{[}88.1, 88.7{]}\end{tabular} \\
\addlinespace
\bottomrule
\end{tabular}
\end{table}

The size of this lead depended on the privacy budget. Without privacy, the four pretrained initializations averaged 84.4 to 87.1 over the five datasets. InDomain-stand led ImageNet by 2.6\% and DINOv3 by 1.4\% macro AUROC. At $\epsilon\le1$, its leads were 15.3\% over ImageNet and 8.8\% over DINOv3 (Fig.~\ref{fig:frontier}f). Moving from non-private training to $\epsilon\le1$ lowered the macro AUROC by 5.1\% on average for InDomain-stand, 7.3\% for InDomain-priv, 12.5\% for DINOv3, and 17.8\% for ImageNet (Fig.~\ref{fig:frontier}g). He initialization had a macro AUROC of 50.2 to 56.5 at every privacy budget. Its non-private average of 63.9 decreased by 11.8\% at $\epsilon\le1$, where InDomain-stand led it by 30.0\%. The lead over ImageNet was largest on VinDr-CXR, whose training set of 12{,}000 radiographs is the smallest of the five. There, InDomain-stand led ImageNet by 31.1\% macro AUROC at $\epsilon\le1$ and by 5.8\% without privacy. On the four datasets with 77{,}870 to 137{,}902 training radiographs, its lead was 9.7\% to 14.5\% at $\epsilon\le1$ and 0.6\% to 3.3\% without privacy.

At $\epsilon\le1$, InDomain-stand ranked first in every dataset and averaged 82.1 over the five datasets. Its macro AUROC ranged from 73.7 $\pm$ 0.5 [72.7, 74.7] on ChestX-ray14 to 90.5 $\pm$ 0.6 [89.4, 91.7] on VinDr-CXR (Fig.~\ref{fig:frontier}i). InDomain-priv ranked second in every dataset. All 60 paired comparisons of InDomain-stand with He initialization, ImageNet, and DINOv3 under privacy favored InDomain-stand at $p<0.0001$ (Fig.~\ref{fig:frontier}h). At this budget, InDomain-stand also had the best value of the area under the precision-recall curve, the mean average precision, accuracy, sensitivity, specificity, and the Brier score (Table~\ref{tab:metrics}). Supplementary Figs.~\ref{sfig:metrics} and~\ref{sfig:metricsnp} plot these metrics for each dataset. Its expected calibration error (ECE) of 8.4 was lower than those of DINOv3 and InDomain-priv and higher than those of ImageNet (7.5) and He initialization (5.8). InDomain-stand ranked first in all 25 combinations of finding and dataset at $\epsilon\le1$ (Supplementary Fig.~\ref{sfig:findings}).

\begin{table*}[p]
\centering
\footnotesize
\caption{Eight metrics of five initializations at $\epsilon\le1$ and without privacy on the test sets of five datasets ($n = 120{,}068$ test radiographs). Each entry is the metric averaged over the five findings in percent, as the mean $\pm$ standard deviation over the five datasets. Lower is better for the Brier score and ECE and higher for the other metrics. Accuracy, sensitivity, and specificity use the thresholds selected on the validation set. Rows are grouped by privacy setting. The models are those of Table~\ref{tab:frontier}. AUPRC, area under the precision-recall curve; AUROC, area under the receiver operating characteristic curve; ECE, expected calibration error; $\epsilon$, privacy budget; He, He initialization~\cite{He2015Delving}; ImageNet, supervised pretraining on ImageNet~\cite{Deng2009ImageNet}; DINOv3, self-supervised DINOv3 pretraining~\cite{Simeoni2025DINOv3}; InDomain-stand and InDomain-priv, supervised pretraining on MIMIC-CXR~\cite{Johnson2019MIMIC} without and with differential privacy; mAP, mean average precision.}
\label{tab:metrics}
\footnotesize
\setlength{\tabcolsep}{4pt}
\begin{tabular}{lccccc}
\toprule
Metric & He & ImageNet & DINOv3 & InDomain-stand & InDomain-priv \\
\midrule
\multicolumn{6}{l}{\textbf{$\epsilon\le1$}} \\
AUROC & 52.1 $\pm$ 1.2 & 66.8 $\pm$ 7.4 & 73.2 $\pm$ 6.7 & 82.1 $\pm$ 6.4 & 77.2 $\pm$ 6.9 \\
AUPRC & 18.7 $\pm$ 5.0 & 27.8 $\pm$ 9.4 & 35.6 $\pm$ 11.1 & 49.1 $\pm$ 13.2 & 41.5 $\pm$ 12.6 \\
mAP & 18.7 $\pm$ 5.0 & 27.9 $\pm$ 9.4 & 35.8 $\pm$ 11.1 & 49.4 $\pm$ 13.5 & 41.7 $\pm$ 12.7 \\
Accuracy & 54.6 $\pm$ 7.0 & 58.3 $\pm$ 11.9 & 63.8 $\pm$ 15.3 & 72.4 $\pm$ 9.4 & 66.5 $\pm$ 14.3 \\
Sensitivity & 50.7 $\pm$ 7.8 & 66.6 $\pm$ 11.8 & 72.9 $\pm$ 5.2 & 77.9 $\pm$ 4.4 & 75.7 $\pm$ 5.2 \\
Specificity & 52.6 $\pm$ 9.7 & 59.1 $\pm$ 13.0 & 63.4 $\pm$ 15.4 & 72.1 $\pm$ 8.8 & 65.9 $\pm$ 14.4 \\
Brier score & 12.7 $\pm$ 4.1 & 11.7 $\pm$ 2.8 & 11.5 $\pm$ 3.1 & 10.2 $\pm$ 3.2 & 11.2 $\pm$ 3.2 \\
ECE & 5.8 $\pm$ 3.8 & 7.5 $\pm$ 3.7 & 8.7 $\pm$ 2.6 & 8.4 $\pm$ 2.4 & 8.9 $\pm$ 2.2 \\
\addlinespace
\multicolumn{6}{l}{\textbf{Non-private}} \\
AUROC & 63.9 $\pm$ 11.5 & 84.6 $\pm$ 5.5 & 85.7 $\pm$ 5.4 & 87.1 $\pm$ 5.8 & 84.4 $\pm$ 5.3 \\
AUPRC & 26.8 $\pm$ 12.7 & 51.6 $\pm$ 13.0 & 54.4 $\pm$ 13.0 & 57.1 $\pm$ 13.7 & 52.3 $\pm$ 13.0 \\
mAP & 26.9 $\pm$ 12.7 & 51.9 $\pm$ 13.1 & 54.7 $\pm$ 13.1 & 57.4 $\pm$ 13.9 & 52.6 $\pm$ 13.1 \\
Accuracy & 57.0 $\pm$ 10.8 & 76.1 $\pm$ 10.1 & 78.0 $\pm$ 10.0 & 78.1 $\pm$ 10.8 & 76.5 $\pm$ 11.0 \\
Sensitivity & 68.3 $\pm$ 4.4 & 78.7 $\pm$ 2.9 & 77.6 $\pm$ 5.5 & 81.0 $\pm$ 3.8 & 77.1 $\pm$ 5.5 \\
Specificity & 52.3 $\pm$ 13.7 & 75.5 $\pm$ 9.2 & 77.7 $\pm$ 9.2 & 78.1 $\pm$ 9.9 & 76.1 $\pm$ 10.2 \\
Brier score & 22.7 $\pm$ 4.2 & 14.6 $\pm$ 5.8 & 14.3 $\pm$ 6.1 & 14.6 $\pm$ 6.9 & 14.2 $\pm$ 5.5 \\
ECE & 30.5 $\pm$ 8.4 & 16.7 $\pm$ 7.7 & 16.3 $\pm$ 8.2 & 17.2 $\pm$ 9.6 & 16.0 $\pm$ 7.6 \\
\addlinespace
\bottomrule
\end{tabular}
\end{table*}

\subsection*{The pretraining domain outweighs the training objective}

\begin{figure*}[p]
\centering
\footnotesize
\includegraphics[width=\textwidth]{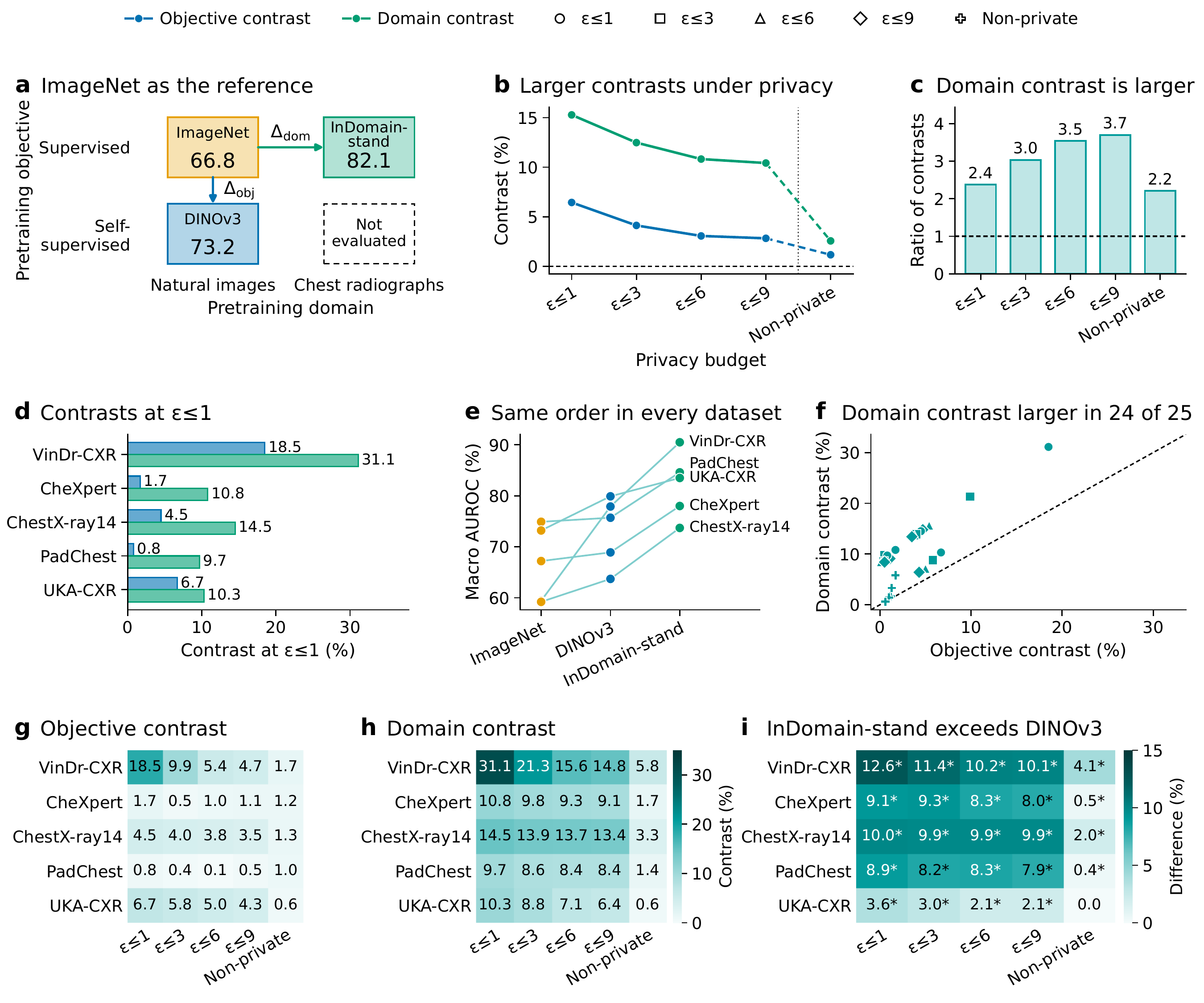}
\caption{Contrasts of the pretraining objective and the pretraining domain on the test sets of VinDr-CXR~\cite{Nguyen2022VinDr}, CheXpert~\cite{Irvin2019CheXpert}, ChestX-ray14~\cite{Wang2017ChestXray8}, PadChest~\cite{Bustos2020PadChest}, and UKA-CXR~\cite{Khader2022Artificial} ($n = 120{,}068$ test radiographs). The objective contrast is the macro AUROC of DINOv3 minus that of ImageNet, which were both pretrained on natural images. The domain contrast is InDomain-stand minus ImageNet, which were both pretrained with labels. Blue marks the objective contrast, and green marks the domain contrast. Marker shapes mark the privacy settings in \textbf{f}. No panel has an error bar. Counts per panel: 3 initializations (\textbf{a}), 5 privacy settings (\textbf{b}, \textbf{c}), 5 datasets (\textbf{d}, \textbf{e}), and 25 combinations of dataset and privacy setting (\textbf{f}--\textbf{i}). \textbf{a}, The three evaluated initializations by pretraining objective and domain, with their mean macro AUROC over the five datasets at $\epsilon\le1$ and the two contrasts as arrows from ImageNet. The fourth combination was not evaluated. \textbf{b}, The two contrasts averaged over the five datasets. \textbf{c}, Ratio of the two averaged contrasts, with a dashed line at 1. \textbf{d}, The two contrasts in each dataset at $\epsilon\le1$. \textbf{e}, Macro AUROC of the three initializations at $\epsilon\le1$, with a line per dataset. \textbf{f}, Domain contrast against objective contrast, with a dashed line of equality. \textbf{g},\textbf{h}, The two contrasts on a shared color scale. \textbf{i}, InDomain-stand minus DINOv3, which equals the domain contrast minus the objective contrast (Eq.~\ref{eq:contrasts}). An asterisk marks $p<0.05$ in an uncorrected two-sided paired bootstrap test with 10{,}000 resamples. AUROC, area under the receiver operating characteristic curve; $\epsilon$, privacy budget; ImageNet, supervised pretraining on ImageNet~\cite{Deng2009ImageNet}; DINOv3, self-supervised DINOv3 pretraining~\cite{Simeoni2025DINOv3}; InDomain-stand, supervised pretraining on MIMIC-CXR~\cite{Johnson2019MIMIC}.}
\label{fig:disentangle}
\end{figure*}

The pretrained initializations differ in their pretraining objective, which is supervised or self-supervised, and in the domain of their pretraining images, which is natural images or chest radiographs. ImageNet, DINOv3, and InDomain-stand cover three of the four combinations of objective and domain (Fig.~\ref{fig:disentangle}a). No public model pretrained with self-supervision on chest radiographs was available for the ConvNeXt backbone. The fourth combination was therefore not evaluated. We compare the two properties through two pairwise contrasts that share supervised ImageNet pretraining as their reference. The objective contrast is DINOv3 minus ImageNet, which were both pretrained on natural images. The domain contrast is InDomain-stand minus ImageNet, which were both pretrained with a supervised objective. Neither contrast is a factorial main effect. With three conditions, the interaction between objective and domain cannot be estimated.

Both contrasts were positive in all 25 combinations of dataset and privacy setting. Averaged over the five datasets, the objective contrast was 1.2\% macro AUROC without privacy and 6.4\% at $\epsilon\le1$. The domain contrast was 2.6\% and 15.3\% (Fig.~\ref{fig:disentangle}b). In every dataset, the domain contrast was larger at each privacy budget than without privacy and largest at $\epsilon\le1$ (Fig.~\ref{fig:disentangle}h). The domain contrast exceeded the objective contrast in 24 of the 25 combinations. The exception was UKA-CXR without privacy, where both contrasts were 0.6\% (Fig.~\ref{fig:disentangle}f). The mean domain contrast over the five datasets was 2.2 times the mean objective contrast without privacy and 2.4 to 3.7 times under privacy (Fig.~\ref{fig:disentangle}c). At $\epsilon\le1$, the domain contrast ranged from 9.7\% on PadChest to 31.1\% on VinDr-CXR. The objective contrast ranged from 0.8\% to 18.5\% (Fig.~\ref{fig:disentangle}d,e). Because both contrasts subtract the same ImageNet run, their difference equals InDomain-stand minus DINOv3. We tested this difference directly with the paired bootstrap. Under privacy, InDomain-stand exceeded DINOv3 by 2.1\% to 12.6\% macro AUROC in every dataset at every budget (Fig.~\ref{fig:disentangle}i). All 20 of these comparisons had $p<0.0001$. Without privacy, InDomain-stand exceeded DINOv3 on four datasets and equaled it on UKA-CXR.

\subsection*{A domain-matched initialization converges with less privacy spent}

\begin{figure*}[p]
\centering
\footnotesize
\includegraphics[width=\textwidth]{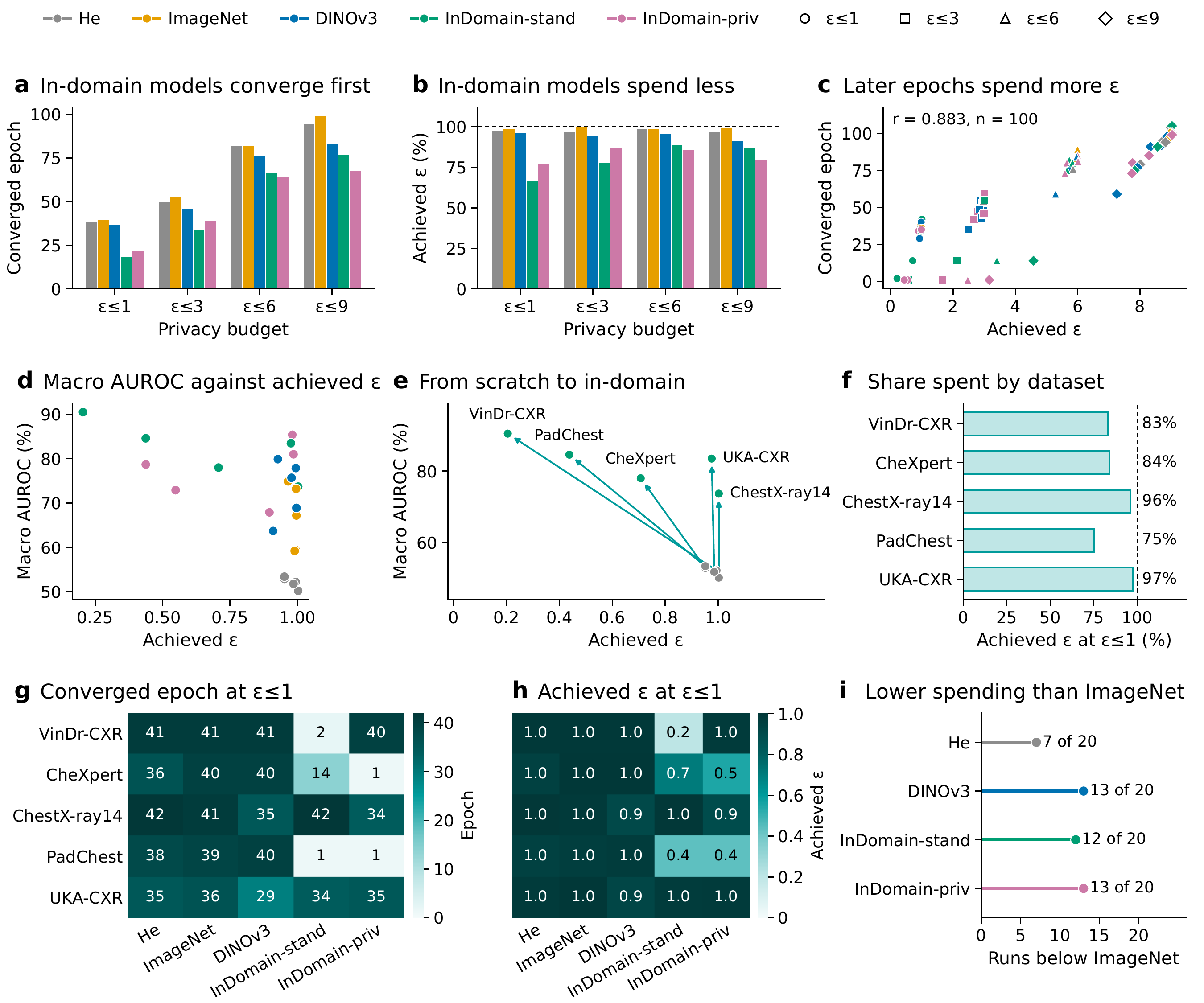}
\caption{Converged epoch and achieved $\epsilon$ of the 100 private runs of the main experiment on the test sets of VinDr-CXR~\cite{Nguyen2022VinDr}, CheXpert~\cite{Irvin2019CheXpert}, ChestX-ray14~\cite{Wang2017ChestXray8}, PadChest~\cite{Bustos2020PadChest}, and UKA-CXR~\cite{Khader2022Artificial} ($n = 120{,}068$ test radiographs). Every run finetuned a ConvNeXt-Small~\cite{Liu2022ConvNet} with differentially private stochastic gradient descent~\cite{Abadi2016Deep}. The converged epoch is the epoch with the highest validation macro AUROC. The achieved $\epsilon$ is the privacy budget spent up to the converged epoch. Colors mark the initializations, and marker shapes mark the privacy budgets in \textbf{c}. No panel has an error bar. Counts per panel: 20 bars, each a mean over the five datasets (\textbf{a}, \textbf{b}), 100 runs (\textbf{c}), 25 runs at $\epsilon\le1$ (\textbf{d}, \textbf{g}, \textbf{h}), 10 runs (\textbf{e}), 5 bars, each a mean over the five initializations (\textbf{f}), and 20 runs per initialization (\textbf{i}). \textbf{a}, Converged epoch. \textbf{b}, Achieved $\epsilon$ as a percentage of the upper end of the budget, with a dashed line at 100\%. \textbf{c}, Converged epoch against achieved $\epsilon$, with the Pearson correlation $r$. \textbf{d}, Macro AUROC against achieved $\epsilon$ at $\epsilon\le1$. \textbf{e}, Arrows from He initialization to InDomain-stand at $\epsilon\le1$, one per dataset. \textbf{f}, Achieved $\epsilon$ at $\epsilon\le1$ as a percentage of 1, with a dashed line at 100\%. \textbf{g},\textbf{h}, Converged epoch (\textbf{g}) and achieved $\epsilon$ (\textbf{h}) at $\epsilon\le1$, with the datasets as the rows of both. \textbf{i}, Number of private runs of each initialization with a lower achieved $\epsilon$ than the ImageNet run on the same dataset at the same budget. Both values were rounded to one decimal before the comparison. AUROC, area under the receiver operating characteristic curve; $\epsilon$, privacy budget; He, He initialization~\cite{He2015Delving}; ImageNet, supervised pretraining on ImageNet~\cite{Deng2009ImageNet}; DINOv3, self-supervised DINOv3 pretraining~\cite{Simeoni2025DINOv3}; InDomain-stand, supervised pretraining on MIMIC-CXR~\cite{Johnson2019MIMIC}; InDomain-priv, differentially private pretraining on MIMIC-CXR.}
\label{fig:budget}
\end{figure*}

The converged epoch of a private run could come long before its training stopped at the end of the budget. At $\epsilon\le1$, the converged epoch averaged 18.6 for InDomain-stand, 22.2 for InDomain-priv, 37.0 for DINOv3, 38.4 for He initialization, and 39.4 for ImageNet (Fig.~\ref{fig:budget}a). At every budget, the two in-domain initializations converged before DINOv3, and DINOv3 converged before He initialization and ImageNet. At $\epsilon\le6$ and $\epsilon\le9$, InDomain-priv converged before InDomain-stand. Seven of the 100 private runs of the main experiment kept the model of their first epoch. All seven started from InDomain-stand or InDomain-priv, which begin finetuning with a classification head already trained on the same five findings. These models therefore start as working chest radiograph classifiers. We attribute their early converged epoch to this starting point. Under privacy, further training can move such a model away from it faster than the clipped and noisy gradient improves it.

Because the spent budget grows with every optimization step, the converged epoch and the achieved $\epsilon$ were correlated across the 100 private runs (Pearson $r=0.883$; Fig.~\ref{fig:budget}c). Earlier convergence therefore gave a lower achieved $\epsilon$. Averaged over the five datasets, the achieved $\epsilon$ at $\epsilon\le1$ was 67\% of the upper end of the budget for InDomain-stand, 77\% for InDomain-priv, 96\% for DINOv3, 98\% for He initialization, and 99\% for ImageNet (Fig.~\ref{fig:budget}b). On VinDr-CXR, InDomain-stand had a macro AUROC of 90.5 at an achieved $\epsilon$ of 0.2, and He initialization had 52.2 at 1.0. On PadChest, InDomain-stand had 84.6 at 0.4, and DINOv3 had 75.7 at 1.0 (Fig.~\ref{fig:budget}d,e). On ChestX-ray14 and UKA-CXR, InDomain-stand had an achieved $\epsilon$ of 1.0 at $\epsilon\le1$, the same as ImageNet and He initialization (Fig.~\ref{fig:budget}h). Rounded to one decimal, the achieved $\epsilon$ of InDomain-stand was lower than that of ImageNet in 12 of the 20 private combinations of dataset and budget. The corresponding counts were 13 for DINOv3 and for InDomain-priv and 7 for He initialization (Fig.~\ref{fig:budget}i).

\subsection*{Private in-domain pretraining outperforms generic pretraining under privacy}

In-domain pretraining needs a corpus of medical images. Such a corpus is patient data and has the same privacy requirements as the data used downstream. InDomain-priv was pretrained on the same MIMIC-CXR radiographs with the same supervised objective. Its pretraining ran under DP-SGD and satisfies a guarantee of $\epsilon=7.93$ at $\delta=6\times10^{-6}$. Private pretraining lowered macro AUROC by about 5\% at every privacy budget. InDomain-priv trailed InDomain-stand by 4.9\% at $\epsilon\le1$ and by 5.0\% to 5.1\% at the looser budgets. Without privacy downstream, the difference was 2.7\%. All 25 comparisons of InDomain-priv with InDomain-stand had $p<0.0001$ (Table~\ref{tab:privatize}).

\begin{table}[p]
\centering
\footnotesize
\caption{Macro AUROC of the two in-domain initializations, whose pretraining on MIMIC-CXR~\cite{Johnson2019MIMIC} differs only in being differentially private or not, on the test sets of five datasets ($n = 120{,}068$ test radiographs). Values are in percent, as the mean $\pm$ standard deviation [95\% confidence interval] of 10{,}000 bootstrap resamples. $\Delta$ is InDomain-stand minus InDomain-priv. Each of the 25 differences has $p<0.0001$ in a two-sided paired bootstrap test with 10{,}000 resamples and no correction for multiple comparisons. Rows are grouped by the privacy setting of the finetuning. The test sets are those of Table~\ref{tab:frontier}. AUROC, area under the receiver operating characteristic curve; $\epsilon$, privacy budget; InDomain-stand and InDomain-priv, supervised pretraining on MIMIC-CXR without and with differential privacy.}
\label{tab:privatize}
\footnotesize
\setlength{\tabcolsep}{5pt}
\begin{tabular}{lccc}
\toprule
Dataset & InDomain-stand & InDomain-priv & $\Delta$ \\
\midrule
\multicolumn{4}{l}{\textbf{$\epsilon\le1$}} \\
VinDr-CXR & 90.5 $\pm$ 0.6 [89.4, 91.7] & 85.4 $\pm$ 0.9 [83.6, 87.2] & +5.1 \\
CheXpert & 78.0 $\pm$ 0.3 [77.5, 78.6] & 72.9 $\pm$ 0.3 [72.3, 73.5] & +5.1 \\
ChestX-ray14 & 73.7 $\pm$ 0.5 [72.7, 74.7] & 67.9 $\pm$ 0.5 [66.9, 69.0] & +5.8 \\
PadChest & 84.6 $\pm$ 0.3 [84.1, 85.1] & 78.7 $\pm$ 0.3 [78.1, 79.3] & +5.9 \\
UKA-CXR & 83.5 $\pm$ 0.2 [83.1, 83.9] & 81.0 $\pm$ 0.2 [80.5, 81.4] & +2.5 \\
\addlinespace
\multicolumn{4}{l}{\textbf{$\epsilon\le3$}} \\
VinDr-CXR & 91.0 $\pm$ 0.5 [89.9, 92.0] & 85.7 $\pm$ 0.9 [83.9, 87.5] & +5.3 \\
CheXpert & 78.3 $\pm$ 0.3 [77.7, 78.8] & 72.8 $\pm$ 0.3 [72.2, 73.3] & +5.5 \\
ChestX-ray14 & 74.2 $\pm$ 0.5 [73.3, 75.2] & 68.3 $\pm$ 0.5 [67.2, 69.3] & +5.9 \\
PadChest & 84.8 $\pm$ 0.3 [84.3, 85.3] & 78.8 $\pm$ 0.3 [78.2, 79.3] & +6.0 \\
UKA-CXR & 84.5 $\pm$ 0.2 [84.1, 84.9] & 81.9 $\pm$ 0.2 [81.4, 82.3] & +2.6 \\
\addlinespace
\multicolumn{4}{l}{\textbf{$\epsilon\le6$}} \\
VinDr-CXR & 90.8 $\pm$ 0.7 [89.4, 92.1] & 85.8 $\pm$ 0.9 [83.9, 87.7] & +5.0 \\
CheXpert & 78.3 $\pm$ 0.3 [77.8, 78.9] & 72.6 $\pm$ 0.3 [72.0, 73.2] & +5.7 \\
ChestX-ray14 & 74.7 $\pm$ 0.5 [73.8, 75.6] & 69.1 $\pm$ 0.5 [68.1, 70.1] & +5.6 \\
PadChest & 85.4 $\pm$ 0.3 [84.9, 85.9] & 79.2 $\pm$ 0.3 [78.6, 79.8] & +6.2 \\
UKA-CXR & 85.0 $\pm$ 0.2 [84.6, 85.3] & 82.7 $\pm$ 0.2 [82.3, 83.1] & +2.3 \\
\addlinespace
\multicolumn{4}{l}{\textbf{$\epsilon\le9$}} \\
VinDr-CXR & 91.2 $\pm$ 0.7 [89.8, 92.4] & 85.8 $\pm$ 1.0 [83.9, 87.6] & +5.4 \\
CheXpert & 78.4 $\pm$ 0.3 [77.9, 78.9] & 72.5 $\pm$ 0.3 [71.9, 73.1] & +5.9 \\
ChestX-ray14 & 74.7 $\pm$ 0.5 [73.8, 75.6] & 69.2 $\pm$ 0.5 [68.3, 70.2] & +5.5 \\
PadChest & 85.6 $\pm$ 0.3 [85.1, 86.1] & 79.3 $\pm$ 0.3 [78.7, 79.9] & +6.3 \\
UKA-CXR & 85.3 $\pm$ 0.2 [84.9, 85.6] & 82.9 $\pm$ 0.2 [82.5, 83.3] & +2.4 \\
\addlinespace
\multicolumn{4}{l}{\textbf{Non-private}} \\
VinDr-CXR & 94.3 $\pm$ 0.5 [93.1, 95.3] & 87.2 $\pm$ 0.8 [85.5, 88.8] & +7.1 \\
CheXpert & 83.5 $\pm$ 0.2 [83.0, 83.9] & 82.0 $\pm$ 0.3 [81.5, 82.5] & +1.5 \\
ChestX-ray14 & 79.4 $\pm$ 0.4 [78.6, 80.1] & 76.2 $\pm$ 0.5 [75.3, 77.1] & +3.2 \\
PadChest & 89.6 $\pm$ 0.2 [89.2, 90.0] & 88.4 $\pm$ 0.2 [87.9, 88.8] & +1.2 \\
UKA-CXR & 88.8 $\pm$ 0.2 [88.4, 89.1] & 88.4 $\pm$ 0.2 [88.1, 88.7] & +0.4 \\
\addlinespace
\bottomrule
\end{tabular}
\end{table}

At $\epsilon\le1$, InDomain-priv led DINOv3 by 4.0\%, ImageNet by 10.4\%, and He initialization by 25.1\% macro AUROC on average. Each of these comparisons had $p<0.0001$ in every dataset. Over the 20 combinations of dataset and privacy budget, InDomain-priv exceeded DINOv3 in 18 and ImageNet in all 20. The two exceptions were UKA-CXR at $\epsilon\le6$ and $\epsilon\le9$, where DINOv3 was higher by 0.2\% and 0.3\%. Without privacy, InDomain-priv trailed DINOv3 in every dataset and by 1.3\% on average. Its advantage over DINOv3 therefore appeared only under private finetuning.

\subsection*{Low-rank adaptation improves the pretrained initializations under privacy}

\begin{figure*}[p]
\centering
\footnotesize
\includegraphics[width=\textwidth]{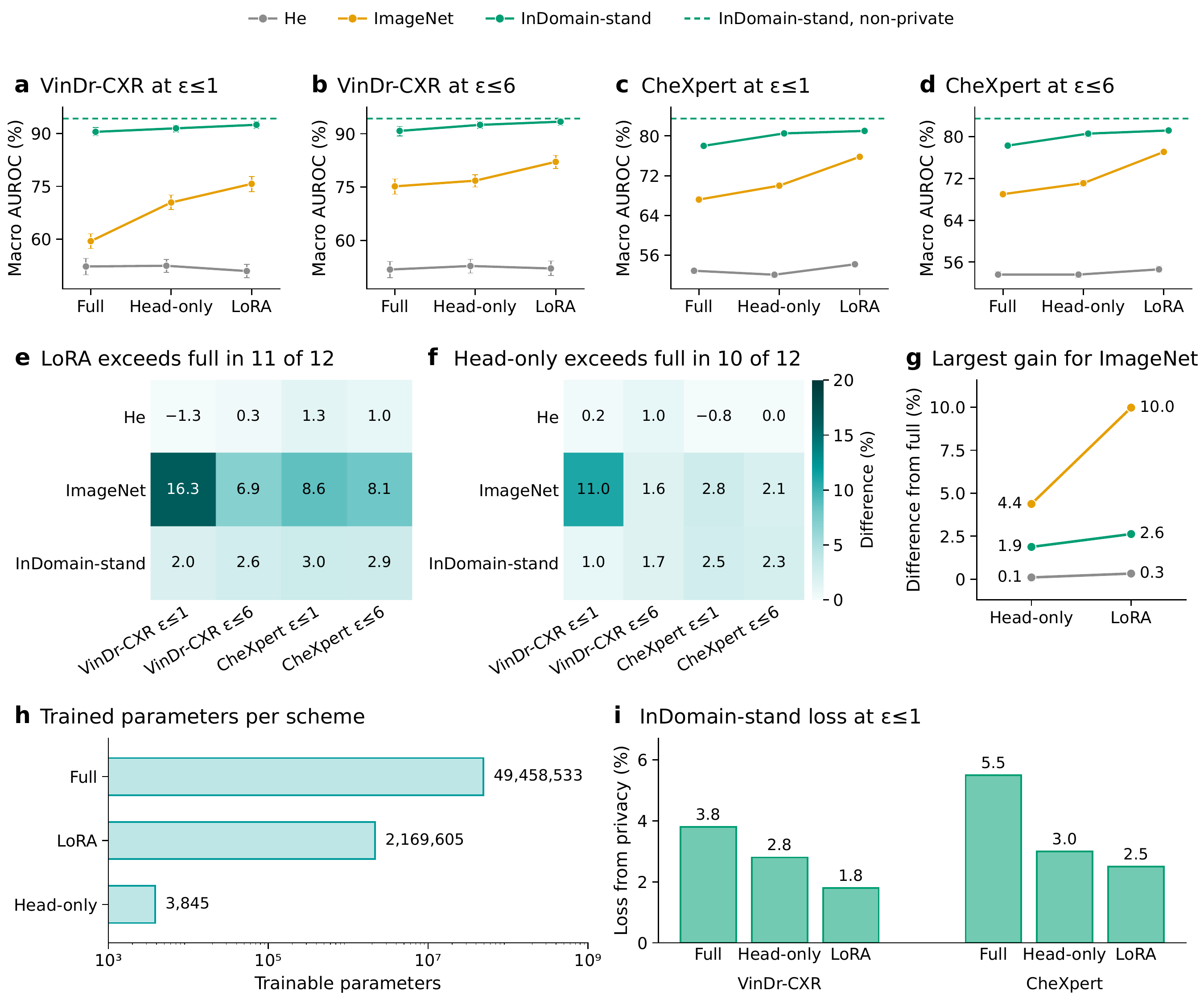}
\caption{Macro AUROC of three finetuning schemes at $\epsilon\le1$ and $\epsilon\le6$ on the test sets of VinDr-CXR~\cite{Nguyen2022VinDr} and CheXpert~\cite{Irvin2019CheXpert} ($n = 32{,}321$ test radiographs). Every model is a ConvNeXt-Small~\cite{Liu2022ConvNet} finetuned with differentially private stochastic gradient descent~\cite{Abadi2016Deep}. Full finetuning trains all parameters, head-only finetuning trains only the classification head, and LoRA~\cite{Hu2022LoRA} trains low-rank adapters and the head. Colors mark the initializations. The dashed line in \textbf{a}--\textbf{d} marks the non-private macro AUROC of InDomain-stand with full finetuning. In \textbf{a}--\textbf{d}, a marker is the mean of 10{,}000 bootstrap resamples of the test set, and the error bar is the 95\% confidence interval, from the 2.5th to the 97.5th percentile. No other panel has an error bar. Counts per panel: 9 runs and 1 non-private reference (\textbf{a}--\textbf{d}), 12 pairs of runs (\textbf{e}, \textbf{f}), 6 values that are means over 4 pairs of runs (\textbf{g}), 3 schemes (\textbf{h}), and 6 private runs against 2 non-private references (\textbf{i}). \textbf{a}--\textbf{d}, Macro AUROC on the test sets of VinDr-CXR ($n = 3{,}000$) at $\epsilon\le1$ (\textbf{a}) and $\epsilon\le6$ (\textbf{b}) and of CheXpert ($n = 29{,}321$) at $\epsilon\le1$ (\textbf{c}) and $\epsilon\le6$ (\textbf{d}). \textbf{e},\textbf{f}, Macro AUROC with LoRA (\textbf{e}) and with head-only finetuning (\textbf{f}) minus that with full finetuning. The two panels share one color scale. A value below 0 takes the color of 0. \textbf{g}, Differences of \textbf{e} and \textbf{f} averaged over the two datasets and the two budgets. \textbf{h}, Number of trainable parameters of ConvNeXt-Small with each scheme, on a logarithmic axis. \textbf{i}, Non-private macro AUROC of InDomain-stand with full finetuning minus its macro AUROC at $\epsilon\le1$ with each scheme. AUROC, area under the receiver operating characteristic curve; $\epsilon$, privacy budget; He, He initialization~\cite{He2015Delving}; ImageNet, supervised pretraining on ImageNet~\cite{Deng2009ImageNet}; InDomain-stand, supervised pretraining on MIMIC-CXR~\cite{Johnson2019MIMIC}; LoRA, low-rank adaptation.}
\label{fig:peft}
\end{figure*}

DP-SGD adds Gaussian noise to the update of every trainable parameter. The norm of the noise therefore grows with the number of trained parameters. We compared full finetuning with two parameter-efficient schemes, head-only finetuning, which trains only the classification head, and LoRA (Fig.~\ref{fig:peft}a--d,h). The comparison covered He initialization, ImageNet, and InDomain-stand on VinDr-CXR and CheXpert at $\epsilon\le1$ and $\epsilon\le6$. LoRA exceeded full finetuning in 11 of the 12 configurations (Fig.~\ref{fig:peft}e). The mean difference over all 12 was 4.3\% macro AUROC, and the largest difference was 16.3\%. The exception was He initialization on VinDr-CXR at $\epsilon\le1$, where LoRA was 1.3\% lower. Head-only finetuning exceeded full finetuning in 10 of the 12 configurations (Fig.~\ref{fig:peft}f). Its mean difference over all 12 was 2.1\%. The other two configurations were He initialization on CheXpert, where head-only finetuning was 0.8\% lower at $\epsilon\le1$ and equal at $\epsilon\le6$.

The gain from LoRA averaged 10.0\% for ImageNet, 2.6\% for InDomain-stand, and 0.3\% for He initialization (Fig.~\ref{fig:peft}g). He initialization remained close to chance with every scheme. In all four combinations of dataset and budget, the best ImageNet model with any scheme was 1.2\% to 14.8\% below InDomain-stand with full finetuning. At $\epsilon\le1$, InDomain-stand with LoRA had the highest macro AUROC of all schemes and initializations on both datasets. Its macro AUROC was 92.5 $\pm$ 0.5 [91.5, 93.5] on VinDr-CXR and 81.0 $\pm$ 0.3 [80.5, 81.5] on CheXpert. These values were 1.8\% and 2.5\% below InDomain-stand with full finetuning without privacy. With full finetuning, the differences were 3.8\% and 5.5\% (Fig.~\ref{fig:peft}i). With InDomain-stand, LoRA therefore reduced the loss in macro AUROC under strict privacy by about half on both datasets.

\subsection*{The initialization advantage persists at unseen institutions and exceeds the effect of model capacity}

\begin{figure*}[p]
\centering
\footnotesize
\includegraphics[width=\textwidth]{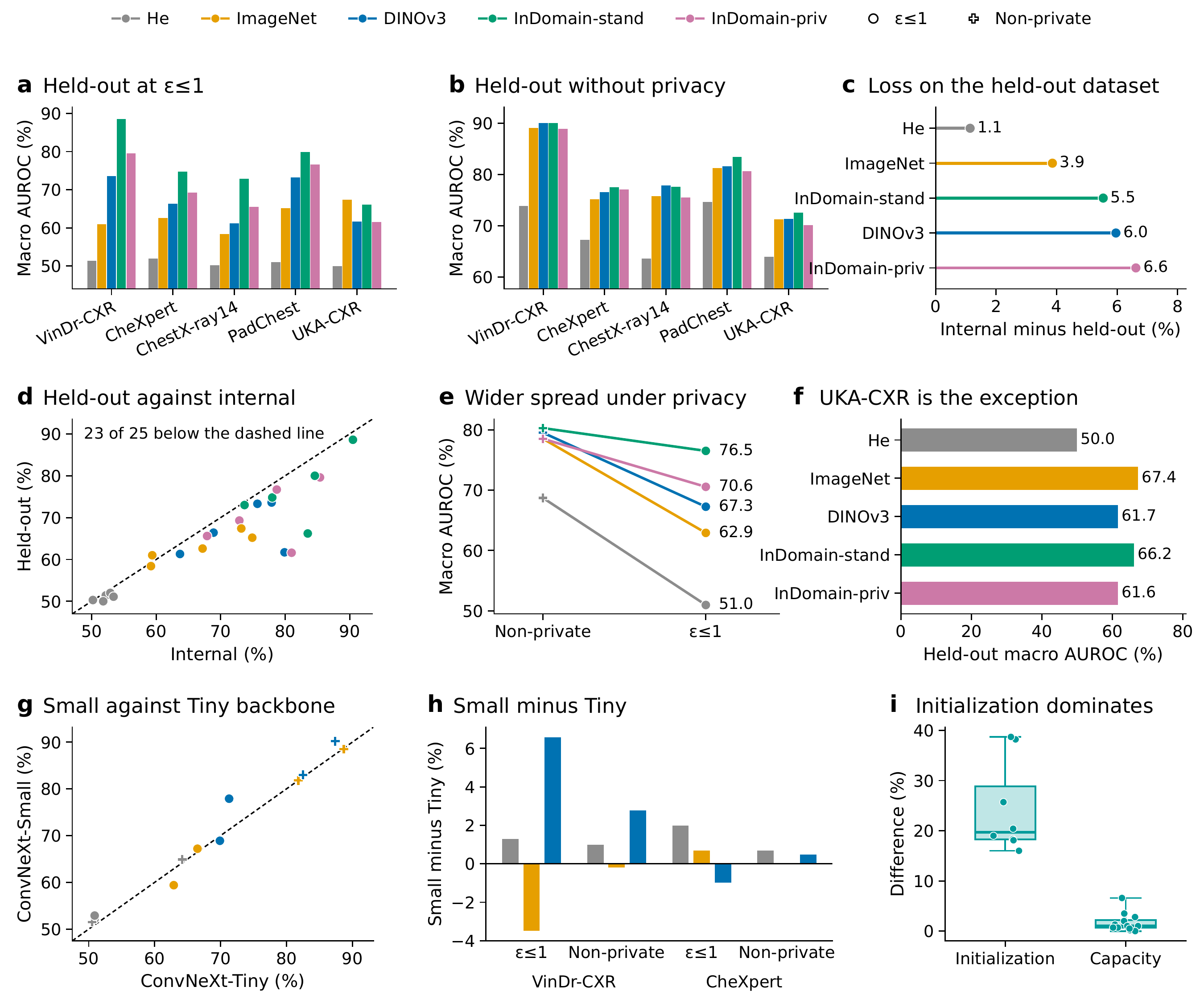}
\caption{Generalization to unseen institutions and effect of backbone capacity on the test sets of VinDr-CXR~\cite{Nguyen2022VinDr}, CheXpert~\cite{Irvin2019CheXpert}, ChestX-ray14~\cite{Wang2017ChestXray8}, PadChest~\cite{Bustos2020PadChest}, and UKA-CXR~\cite{Khader2022Artificial} ($n = 120{,}068$ test radiographs). A ConvNeXt-Small~\cite{Liu2022ConvNet} was finetuned on the other four datasets for the held-out value of a dataset and on the dataset itself for its internal value. Private finetuning used differentially private stochastic gradient descent~\cite{Abadi2016Deep}. ConvNeXt-Tiny was trained from He initialization, ImageNet, and DINOv3 on VinDr-CXR and CheXpert. Colors mark the initializations, and marker shapes mark the privacy settings. No panel has an error bar. Counts per panel: 25 runs (\textbf{a}, \textbf{b}, \textbf{d}), means over the five datasets (\textbf{c}, \textbf{e}), 5 runs (\textbf{f}), 12 pairs of runs (\textbf{g}, \textbf{h}), and 8 and 12 values (dots in \textbf{i}). \textbf{a},\textbf{b}, Macro AUROC on each held-out dataset at $\epsilon\le1$ (\textbf{a}) and without privacy (\textbf{b}). \textbf{c}, Internal minus held-out macro AUROC at $\epsilon\le1$. \textbf{d}, Held-out against internal at $\epsilon\le1$, with a dashed line of equality. \textbf{e}, Held-out macro AUROC. \textbf{f}, Macro AUROC with UKA-CXR held out at $\epsilon\le1$. \textbf{g}, ConvNeXt-Small against ConvNeXt-Tiny, with a dashed line of equality. \textbf{h}, ConvNeXt-Small minus ConvNeXt-Tiny. \textbf{i}, Range across the three initializations within each backbone, and absolute difference between the two backbones for each initialization, per dataset and privacy setting. A box spans the 25th to the 75th percentile with the median as a line and whiskers to the smallest and largest value. AUROC, area under the receiver operating characteristic curve; $\epsilon$, privacy budget; He, He initialization~\cite{He2015Delving}; ImageNet, supervised pretraining on ImageNet~\cite{Deng2009ImageNet}; DINOv3, self-supervised DINOv3 pretraining~\cite{Simeoni2025DINOv3}; InDomain-stand, supervised pretraining on MIMIC-CXR~\cite{Johnson2019MIMIC}; InDomain-priv, differentially private pretraining on MIMIC-CXR.}
\label{fig:lodo}
\end{figure*}

Deployment usually means running a model at a new institution. To test this setting, we held out each dataset in turn, trained on the other four, and measured macro AUROC on the held-out dataset. At $\epsilon\le1$, the macro AUROC of the four pretrained initializations on a held-out dataset was 3.9\% to 6.6\% lower than after training on that dataset. He initialization lost 1.1\%, because it was close to chance in both evaluations (Fig.~\ref{fig:lodo}a,c,d and Supplementary Table~\ref{stab:lodo}). At $\epsilon\le1$, InDomain-stand led on four of the five held-out datasets. The held-out averages were 76.5 for InDomain-stand, 70.6 for InDomain-priv, 67.3 for DINOv3, 62.9 for ImageNet, and 51.0 for He initialization (Fig.~\ref{fig:lodo}a,e). Without privacy, the averages of the four pretrained initializations were 78.5 to 80.3. InDomain-stand led on three of the five held-out datasets, had the same macro AUROC as DINOv3 on VinDr-CXR (90.1), and was 0.2\% below DINOv3 on ChestX-ray14 (Fig.~\ref{fig:lodo}b,e). With UKA-CXR held out at $\epsilon\le1$, the macro AUROC was 67.4 for ImageNet, 66.2 for InDomain-stand, and 61.7 for DINOv3 (Fig.~\ref{fig:lodo}f). UKA-CXR is the only dataset collected primarily in intensive care and the only dataset labeled by structured clinical reporting. The in-domain benefit may therefore depend on how closely the pretraining corpus resembles the deployment setting, beyond the imaging modality.

The two in-domain initializations exist for ConvNeXt-Small only. The comparison of backbone capacities therefore covers He initialization, ImageNet, and DINOv3. Moving from ConvNeXt-Small to the smaller ConvNeXt-Tiny changed macro AUROC by 1.7\% on average and by at most 6.6\% (Fig.~\ref{fig:lodo}g,h and Supplementary Table~\ref{stab:capacity}). Within ConvNeXt-Tiny, the initializations differed by up to 20.4\% on VinDr-CXR and 19.0\% on CheXpert at $\epsilon\le1$ (Fig.~\ref{fig:lodo}i). At a fixed privacy budget, the initialization therefore changed macro AUROC more than the capacity of the backbone did.

\subsection*{In-domain pretraining gives the highest macro AUROC in the worst subgroup under privacy}

\begin{figure*}[p]
\centering
\footnotesize
\includegraphics[width=\textwidth]{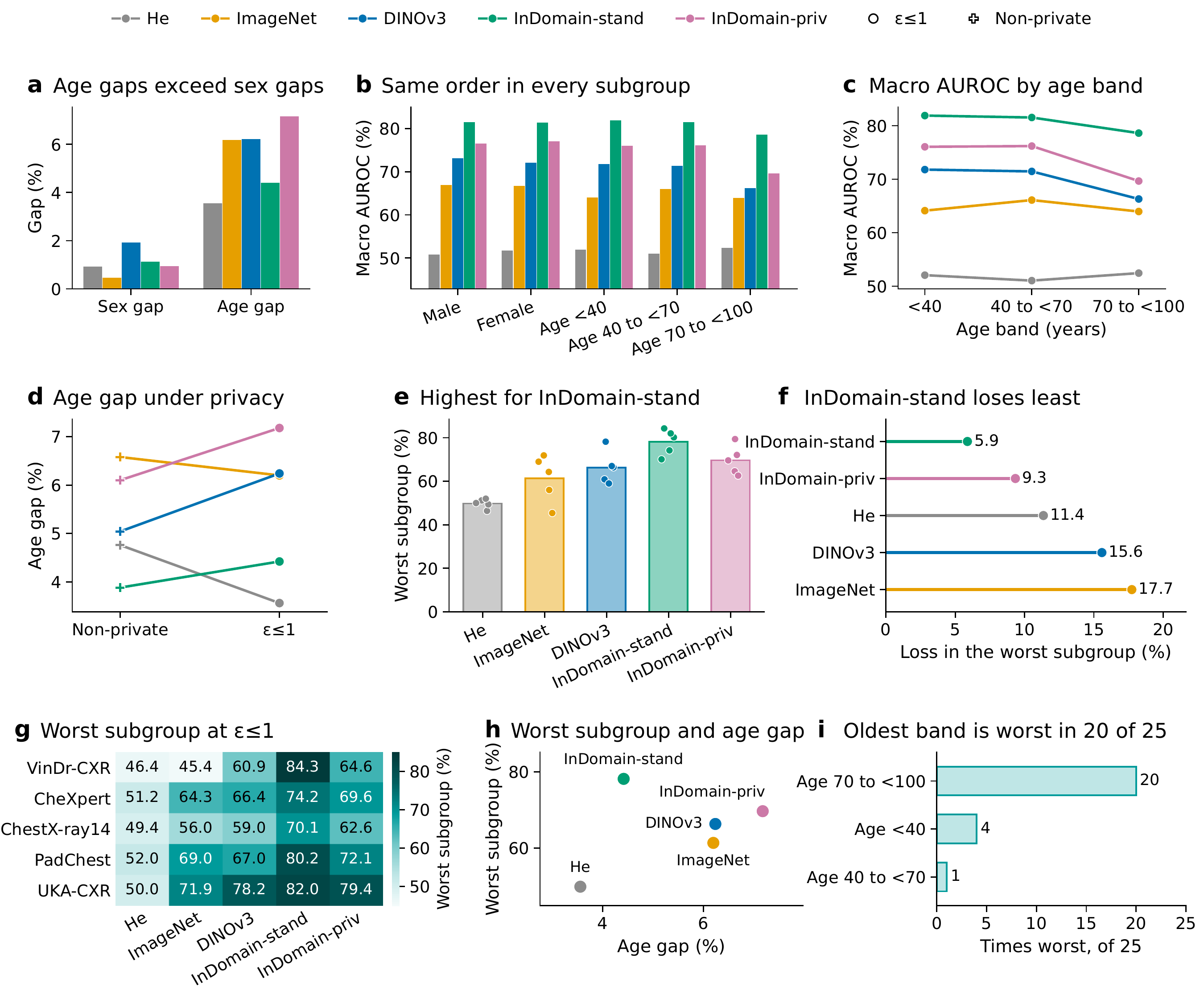}
\caption{Macro AUROC of ConvNeXt-Small~\cite{Liu2022ConvNet} classifiers finetuned with differentially private stochastic gradient descent~\cite{Abadi2016Deep} in demographic subgroups of the test sets of VinDr-CXR~\cite{Nguyen2022VinDr}, CheXpert~\cite{Irvin2019CheXpert}, ChestX-ray14~\cite{Wang2017ChestXray8}, PadChest~\cite{Bustos2020PadChest}, and UKA-CXR~\cite{Khader2022Artificial} ($n = 120{,}068$ test radiographs). The subgroups are male and female patients and the age bands from above 0 to below 40, from 40 to below 70, and from 70 to below 100 years. The sex gap is the absolute difference between the sexes, the age gap is the largest minus the smallest age-band value, and the worst subgroup has the smallest value of the five. Colors mark the initializations, and marker shapes mark the privacy settings in \textbf{d}. No panel has an error bar. Counts per panel: means over the five datasets (\textbf{a}--\textbf{f}, \textbf{h}), 5 datasets per initialization (dots in \textbf{e}), and 25 combinations of dataset and initialization (\textbf{g}, \textbf{i}). \textbf{a}, Sex gap and age gap at $\epsilon\le1$. \textbf{b}, Macro AUROC within each subgroup at $\epsilon\le1$. \textbf{c}, Macro AUROC within each age band at $\epsilon\le1$. \textbf{d}, Age gap without privacy and at $\epsilon\le1$. \textbf{e}, Macro AUROC of the worst subgroup at $\epsilon\le1$. \textbf{f}, Macro AUROC of the worst subgroup without privacy minus that at $\epsilon\le1$. \textbf{g}, Macro AUROC of the worst subgroup at $\epsilon\le1$ in each dataset. \textbf{h}, Macro AUROC of the worst subgroup against age gap at $\epsilon\le1$. \textbf{i}, Number of combinations of dataset and initialization in which each subgroup has the lowest macro AUROC at $\epsilon\le1$. AUROC, area under the receiver operating characteristic curve; $\epsilon$, privacy budget; He, He initialization~\cite{He2015Delving}; ImageNet, supervised pretraining on ImageNet~\cite{Deng2009ImageNet}; DINOv3, self-supervised DINOv3 pretraining~\cite{Simeoni2025DINOv3}; InDomain-stand, supervised pretraining on MIMIC-CXR~\cite{Johnson2019MIMIC}; InDomain-priv, differentially private pretraining on MIMIC-CXR.}
\label{fig:fairness}
\end{figure*}

Under DP, the largest losses in accuracy are known to occur in the smallest patient groups~\cite{Bagdasaryan2019Differential}. We therefore measured macro AUROC separately by sex and in three age bands in every test set. At $\epsilon\le1$ and averaged over the five datasets, the gap between the best and the worst age band was 3.6\% to 7.2\% macro AUROC for the five initializations. The average gap between male and female patients was 0.5\% to 1.9\%. The age gap was therefore larger than the sex gap for every initialization (Fig.~\ref{fig:fairness}a). For the four pretrained initializations, macro AUROC was lowest in the oldest age band, 70 to below 100 years (Fig.~\ref{fig:fairness}b,c). This band was the worst subgroup in 20 of the 25 combinations of dataset and initialization (Fig.~\ref{fig:fairness}i). From non-private training to $\epsilon\le1$, the average age gap increased for InDomain-stand (3.9\% to 4.4\%), InDomain-priv (6.1\% to 7.2\%), and DINOv3 (5.0\% to 6.2\%). It decreased for ImageNet (6.6\% to 6.2\%) and He initialization (4.8\% to 3.6\%) (Fig.~\ref{fig:fairness}d and Supplementary Fig.~\ref{sfig:subgroups}).

At $\epsilon\le1$, the worst subgroup averaged 78.2 macro AUROC with InDomain-stand, 69.7 with InDomain-priv, 66.3 with DINOv3, 61.3 with ImageNet, and 49.8 with He initialization. InDomain-stand had the highest worst-subgroup macro AUROC in all five datasets (Fig.~\ref{fig:fairness}e,g and Supplementary Fig.~\ref{sfig:subgroups}). Compared with non-private training, privacy lowered the worst-subgroup macro AUROC by 5.9\% for InDomain-stand, 9.3\% for InDomain-priv, 11.4\% for He initialization, 15.6\% for DINOv3, and 17.7\% for ImageNet (Fig.~\ref{fig:fairness}f). The two narrowest age gaps at $\epsilon\le1$ belonged to He initialization (3.6\%) and InDomain-stand (4.4\%). He initialization had a macro AUROC of 46.4 to 54.1 in every subgroup of every dataset. A narrow gap can therefore come from uniformly low performance. Over the five datasets, the lowest worst-subgroup macro AUROC was 70.1 for InDomain-stand and at most 62.6 for the other initializations (Fig.~\ref{fig:fairness}g). The age bands differ more in size and in finding prevalence than the two sexes (Supplementary Table~\ref{stab:demographics}). Per-sample clipping with added noise reduces the learning signal most for small subgroups with a skewed prevalence~\cite{Bagdasaryan2019Differential}. InDomain-stand had the smallest age gap of the pretrained initializations at $\epsilon\le1$ and the smallest loss in worst-subgroup macro AUROC under privacy. A domain-matched initialization may therefore lessen the uneven effect of privacy on subgroups.

\section*{Discussion}

We asked whether, in the differentially private finetuning of chest radiograph classifiers, the benefit of a pretrained initialization comes from its pretraining objective or from its pretraining images. On five datasets, InDomain-stand had the highest macro AUROC in 24 of the 25 combinations of dataset and privacy setting. At $\epsilon\le1$, the average macro AUROC was 82.1 for InDomain-stand, 73.2 for DINOv3, and 66.8 for ImageNet. The domain contrast exceeded the objective contrast in 24 of the 25 combinations. Under privacy, the mean domain contrast over the five datasets was 2.4 to 3.7 times the mean objective contrast. Pretraining on the same radiographs under DP lowered macro AUROC by about 5\% and left InDomain-priv ahead of DINOv3 and ImageNet in every dataset at $\epsilon\le1$. Generic pretraining still helped under privacy, since ImageNet and DINOv3 exceeded He initialization by 14.7\% to 22.5\% macro AUROC on average at every budget. Pretraining on the imaging modality of the task, however, reduced the loss of accuracy under private finetuning more than a self-supervised objective on natural images did.

Without privacy, ImageNet pretraining transfers well to chest radiographs~\cite{Ke2021CheXtransfer}. On large medical datasets, its benefit over training from scratch can be small~\cite{Raghu2019Transfusion}. Under privacy, the lead of InDomain-stand over ImageNet at $\epsilon\le1$ was about six times its lead without privacy. DP-SGD clips every per-sample gradient and adds Gaussian noise to their sum~\cite{Abadi2016Deep}. The privacy budget limits the number of these noisy updates. The error added by the noise does not grow with the number of parameters when the gradients concentrate in a few directions~\cite{Li2022When}. Public pretraining has also been proposed to help private training by selecting a good basin of the loss landscape, which is hard to do with private data~\cite{Ganesh2023Why}. The private data then refine the model within that basin. In this interpretation, an initialization pretrained on chest radiographs to recognize the same five findings needs smaller changes to fit the target task. The same noise then disturbs these changes less. The two in-domain initializations also converged before DINOv3 and ImageNet at every budget. In the main experiment, every private run that kept the model of its first epoch started from one of these two initializations.

Self-supervised pretraining has made medical imaging models more robust and more data-efficient in standard training~\cite{Azizi2023Robust,Krishnan2022SelfSupervised}. Self-supervised features also improve private learning~\cite{Tramer2021Differentially}. In our study, DINOv3 exceeded ImageNet in every dataset at every privacy setting. Under privacy, however, changing the objective improved macro AUROC less than changing the domain. Pretraining on radiological images instead of ImageNet has improved transfer most clearly on small target datasets~\cite{Mei2022RadImageNet}. DP lowers accuracy most in small cohorts~\cite{Mohammadi2026Differential}. Our largest in-domain advantage, on VinDr-CXR with its 12{,}000 training radiographs, is consistent with both observations. On natural-image benchmarks, private finetuning of models pretrained on large public image collections approaches non-private accuracy~\cite{De2022Unlocking}. Whether such benchmarks measure how well public pretraining transfers to sensitive domains has been questioned~\cite{Tramer2024Position}. We provide this measurement for chest radiographs. There, pretraining on natural images, with or without labels, left a larger loss of accuracy under privacy than pretraining on radiographs.

Reports of DP in medical imaging often state their privacy parameters incompletely~\cite{Mohammadi2026Differential,Ziller2024Reconciling}. A practical guide asks for the unit of privacy, the definition of neighboring datasets, the accountant and its assumptions, and the guarantee of the final model~\cite{Ponomareva2023How}. We report each of these, with the radiograph as the unit of privacy. Beside the budget of every run, we also report its achieved $\epsilon$, which is the budget spent up to the converged epoch. The guarantee of a final model remains that of its whole run, because the converged epoch was chosen after training had ended. At $\epsilon\le1$, InDomain-stand converged after spending two thirds of the budget on average. An initialization matched to the modality may therefore reach its best accuracy within a smaller budget than a generic initialization.

Private finetuning conventionally treats the pretraining data as public. This assumption is difficult to defend when the pretraining data are patient radiographs~\cite{Tramer2024Position}. Pretraining under DP removes it, because the guarantee then also covers each radiograph of the pretraining corpus. In our study, InDomain-priv exceeded DINOv3 in 18 of the 20 combinations of dataset and privacy budget and trailed it without privacy in every dataset. Its advantage therefore appeared only when the finetuning was private as well. Developers of open medical foundation models~\cite{Ma2025Fully} could release encoders pretrained under DP with a stated guarantee, for hospitals that finetune them privately.

On natural images, very large batches, long schedules, and very large public pretraining corpora recover much of the accuracy lost under DP-SGD~\cite{De2022Unlocking,Kurakin2022Toward}. A hospital that trains a model on its own patients rarely has these resources. In our study, the size of the backbone changed macro AUROC by 1.7\% on average between ConvNeXt-Tiny and ConvNeXt-Small. Within one backbone, the initialization changed it by up to 20.4\%. Matching the initialization to the modality needed no additional data from the target hospital, no additional compute, and no change to the privacy accounting. Chest radiograph classifiers lose accuracy at institutions that did not contribute training data~\cite{Khader2022Artificial,TayebiArasteh2024Securing}. When one dataset at a time was held out, InDomain-stand still led on four of the five held-out datasets under strict privacy.

Under privacy, LoRA improved on full finetuning in 11 of the 12 configurations, and head-only finetuning did so in 10. Parameter-efficient finetuning is known to improve the differentially private training of language models~\cite{Yu2022Differentially,Li2022Large}. On natural images, linear models trained privately on fixed features outperformed end-to-end private training at moderate budgets~\cite{Tramer2021Differentially}. In private few-shot classification, parameter-efficient adapters performed about as well as training the whole network~\cite{Tobaben2023Efficacy}. Our results extend these observations to chest radiograph classification. The gain from LoRA was largest for ImageNet and smaller for InDomain-stand, which already starts close to the target domain. He initialization gained 0.3\% on average and remained close to chance with every scheme. LoRA did not change the order of the initializations. Parameter-efficient finetuning therefore complemented a domain-matched initialization without replacing it.

Privacy can amplify differences in accuracy between groups~\cite{Bagdasaryan2019Differential}. A recent review found that few medical studies examine subgroup performance under DP and that several of them report wider subgroup gaps~\cite{Mohammadi2026Differential}. On a large chest radiograph cohort, private training largely did not increase disparities by age or sex~\cite{TayebiArasteh2024Preserving}. In our study, the disparities were larger between age bands than between the sexes. Chest radiograph classifiers are known to perform unevenly across age groups also without privacy~\cite{SeyyedKalantari2021Underdiagnosis}. Under strict privacy, the two narrowest age gaps belonged to InDomain-stand and to He initialization, whose worst subgroups differed by 28.4\% macro AUROC. A gap statistic does not distinguish these two cases, because a uniformly poor model and a uniformly good model both have a small gap. He initialization also had the lowest ECE at $\epsilon\le1$, although its macro AUROC was close to chance. A model can be well calibrated without separating the radiographs with a finding from those without it. Reporting the macro AUROC of the worst subgroup beside the gap would remove this ambiguity in evaluations of private medical models. A ranking metric reported beside a calibration metric would do the same for calibration.

Several limitations qualify these conclusions. First, the comparison of objective and domain lacks a self-supervised initialization pretrained on chest radiographs, because no public model of this kind was available for the ConvNeXt backbone. Self-supervised radiograph encoders exist for other architectures~\cite{Tiu2022Expert,Yao2025EVAX}. Pretraining such an encoder for ConvNeXt would complete the four combinations of objective and domain and allow their interaction to be estimated. Second, under privacy, the full finetuning of the two in-domain initializations used half the learning rate of the other initializations, since they were already adapted to chest radiographs. The comparison under privacy therefore does not separate the initialization from the learning rate. Repeating these private runs at the common learning rate would separate them. Third, at $\epsilon\le1$, InDomain-stand did not lead when UKA-CXR was held out. UKA-CXR was collected primarily in intensive care, while the pretraining corpus consists of the radiographs of patients who presented to an emergency department~\cite{Johnson2019MIMIC}. Pretraining corpora from several care settings would allow us to test whether the in-domain benefit depends on the care setting. Fourth, we studied one privacy mechanism, DP-SGD, with one backbone family at two capacities on one imaging modality. Because the in-domain initializations exist for ConvNeXt-Small only, the comparison of all five initializations is limited to this backbone. Repeating the design with vision transformers, with three-dimensional imaging, and with federated, synthetic-data, and hybrid approaches~\cite{Kaissis2021EndToEnd,Lotfinia2025Boosting} would show whether the initializations keep their order in these settings. Fifth, we analyzed fairness by sex and age only, because race and ethnicity are not recorded in the curated lists of these datasets. Intersectional and finer subgroups would need cohorts that record these attributes. Sixth, the guarantees bound the influence of each radiograph and not the influence of each patient, who can contribute several radiographs. Opacus samples and clips single training examples, and VinDr-CXR records no patient identifiers. A patient-level guarantee would require sampling and clipping all radiographs of a patient together. We leave this guarantee to future work. Finally, the datasets are retrospective and were labeled by expert annotation, by automated labeling of the radiology reports, or by structured clinical reporting, without a common reference standard. A prospective validation in clinical use would be needed before deployment.

Under DP, the domain of the pretraining images affected diagnostic accuracy more than the pretraining objective or the size of the backbone. A domain-matched initialization gave the highest macro AUROC in the worst subgroup and reached its best validation performance with less of the privacy budget. With LoRA, it recovered about half of the accuracy lost under strict privacy on the two datasets tested. Pretraining on the target modality under DP still outperformed the generic initializations in most private settings. With such pretraining, the guarantee covers the pretraining data as well as the finetuning data. For a hospital that trains a model on its own patients, choosing such an initialization improves private accuracy without additional data from its own patients. Openly released encoders that were pretrained under DP on medical images may help private medical AI more than generic self-supervised encoders.

\section*{Methods}
\subsection*{Ethics statement}

This study was conducted in accordance with the relevant guidelines and regulations. The Ethics Committee of the Medical Faculty of RWTH Aachen University approved the use of UKA-CXR~\cite{Khader2022Artificial} for this retrospective study (Reference No.\ EK 22-319) and waived individual informed consent. The other five datasets are previously collected, de-identified chest radiographs and labels released for research, some openly and some under credentialed access. MIMIC-CXR~\cite{Johnson2019MIMIC} was approved by the institutional review board of Beth Israel Deaconess Medical Center, which waived individual patient consent. CheXpert~\cite{Irvin2019CheXpert} was approved by the institutional review board of Stanford Hospital, which waived individual consent for the de-identified data. The institutional review boards of Hanoi Medical University Hospital and Hospital 108 approved VinDr-CXR~\cite{Nguyen2022VinDr} and waived patient consent. An institutional research committee approved PadChest~\cite{Bustos2020PadChest}. The NIH Clinical Center released ChestX-ray14~\cite{Wang2017ChestXray8} after removing all personally identifiable information. No approving committee and no consent procedure are reported for ChestX-ray14. Each dataset was used according to the conditions of its release. No new patient data were collected and no participants were recruited. No re-identification was attempted. No image or record is redistributed. All models ran on local hardware without any external service.

\subsection*{Datasets and preprocessing}

The models were finetuned and evaluated on five chest radiograph datasets with $n = 591{,}884$ radiographs in total. They are VinDr-CXR ($n = 18{,}000$)~\cite{Nguyen2022VinDr}, CheXpert ($n = 157{,}878$)~\cite{Irvin2019CheXpert}, ChestX-ray14 ($n = 112{,}120$)~\cite{Wang2017ChestXray8}, PadChest ($n = 110{,}525$)~\cite{Bustos2020PadChest}, and UKA-CXR ($n = 193{,}361$)~\cite{Khader2022Artificial}. A sixth dataset, MIMIC-CXR~\cite{Johnson2019MIMIC}, was used only to pretrain the two in-domain initializations. Every dataset was mapped to the same five findings, which are atelectasis, cardiomegaly, pleural effusion, pneumonia, and no finding. Each radiograph therefore has five labels in this order. The patient of each radiograph is known in CheXpert, ChestX-ray14, PadChest, and UKA-CXR. These four datasets were split into training, validation, and test sets by patient. No patient therefore appears in two splits. VinDr-CXR provides no patient identifiers and was split by radiograph. Supplementary Table~\ref{stab:datasets} lists the size of every split. Supplementary Table~\ref{stab:prevalence} lists the number of positive radiographs of every finding in every split. In Supplementary Note~\ref{snote:curation}, we give each curation step with its counts.

\paragraph{VinDr-CXR.} VinDr-CXR~\cite{Nguyen2022VinDr} was released with $n = 15{,}000$ training and $n = 3{,}000$ test radiographs and without a validation split. We drew 20\% of the official training split as the validation set and kept the official test split unchanged. The draw was stratified by the combination of the five labels and used seed 42. It gave $n = 12{,}000$ training, $n = 3{,}000$ validation, and $n = 3{,}000$ test radiographs. The labels are binary. Sex is recorded for $n = 8{,}608$ radiographs, and age is recorded for $n = 4{,}234$.

\paragraph{CheXpert.} Of the $n = 184{,}325$ radiographs of the curated CheXpert list~\cite{Irvin2019CheXpert}, we used the $n = 157{,}878$ frontal radiographs of 57{,}872 patients and discarded the lateral radiographs. They comprise $n = 115{,}458$ training, $n = 13{,}099$ validation, and $n = 29{,}321$ test radiographs. The labels come from a rule-based report labeler that marks each finding as present, absent, uncertain, or not mentioned. A finding not mentioned in the report was counted as absent. For atelectasis and pleural effusion, a radiograph with an uncertain label was excluded from the training loss, the validation, the decision thresholds, and the evaluation of that finding. The curated list contains no uncertain label for cardiomegaly, pneumonia, or no finding. Of the frontal radiographs, 59\% are from male patients. The median age is 61 years, with an interquartile range (IQR) of 49 to 74 years.

\paragraph{ChestX-ray14.} ChestX-ray14~\cite{Wang2017ChestXray8} contributed $n = 112{,}120$ radiographs of 30{,}805 patients. In the curated list, the official test split of $n = 25{,}596$ radiographs is unchanged. The official training and validation set is divided by patient into $n = 77{,}870$ training and $n = 8{,}654$ validation radiographs. The labels are binary. Its effusion label was used as pleural effusion. Of the radiographs, 56\% are from male patients, and the median age is 49 years (IQR 34 to 59).

\paragraph{PadChest.} Of the $n = 160{,}704$ radiographs of the curated PadChest list~\cite{Bustos2020PadChest}, we used the $n = 110{,}525$ posteroanterior, anteroposterior, and horizontal anteroposterior radiographs of 67{,}205 patients and discarded the others. They comprise $n = 79{,}697$ training, $n = 8{,}783$ validation, and $n = 22{,}045$ test radiographs. The labels are binary. Of the used radiographs, 50\% are from male patients, and the median age is 63 years (IQR 47 to 74).

\paragraph{UKA-CXR.} UKA-CXR~\cite{Khader2022Artificial} contributed $n = 193{,}361$ radiographs of 54{,}176 patients. The radiographs were collected primarily in intensive care at University Hospital RWTH Aachen. They comprise $n = 137{,}902$ training, $n = 15{,}353$ validation, and $n = 40{,}106$ test radiographs. The labels come from structured clinical reporting, which grades each finding on an ordinal scale. Cardiomegaly is graded for the whole heart, and atelectasis, pleural effusion, and pneumonic infiltrates are graded for each side of the chest. A finding was counted as present at grade 3 or 4, on either side for a finding graded per side. Pneumonic infiltrates were counted as pneumonia. A healthy examination was counted as no finding. Of the radiographs, 65\% are from male patients, and the median age is 68 years (IQR 58 to 77).

\paragraph{MIMIC-CXR.} The two in-domain initializations were pretrained on the $n = 153{,}255$ frontal radiographs of the training split of MIMIC-CXR~\cite{Johnson2019MIMIC}, which has $n = 215{,}187$ frontal radiographs in total. The MIMIC-CXR labels were produced by the same rule-based labeler as those of CheXpert.

\paragraph{Preprocessing.} Every radiograph was resized to $224\times224$ pixels~\cite{Sabottke2020Effect}, intensity-normalized per image, converted to 8-bit grayscale, and histogram-equalized. Each preprocessed radiograph was given to the network as a three-channel image with intensities scaled to the unit interval. In non-private training, the training radiographs were augmented by random horizontal flipping with probability 0.5 and by random rotation of up to 7 degrees. No augmentation was applied in private training, in validation, or in testing.

\subsection*{Network architecture and initializations}

Every model was a ConvNeXt classifier~\cite{Liu2022ConvNet}. Its four stages of ConvNeXt blocks have 96, 192, 384, and 768 channels. ConvNeXt-Small is the main backbone and has 3, 3, 27, and 3 blocks in these stages. ConvNeXt-Tiny is used in the capacity experiment and has 3, 3, 9, and 3 blocks. The last feature map is averaged over its spatial positions and layer-normalized into a 768-dimensional embedding. A linear head maps this embedding to one logit for each of the five findings. With the head, ConvNeXt-Small has 49.5 million parameters, and ConvNeXt-Tiny has 27.8 million.

He initialization is the random initialization of this study. By random, we mean that none of its weights comes from a trained model, and not that the weights are arbitrary. Every convolutional and linear weight of the network and of its head was drawn from the He (Kaiming) normal distribution~\cite{He2015Delving}. This standard initialization for training deep networks from scratch sets the variance of the weights of a layer with $n_{\mathrm{in}}$ inputs to $2/n_{\mathrm{in}}$. All biases were set to zero. Every layer normalization was reset to unit scale and zero shift. Every block multiplies its residual branch by a learnable vector called the layer scale. In He initialization, each layer scale was set to $10^{-6}$, as in the original ConvNeXt initialization~\cite{Liu2022ConvNet}. We applied this reinitialization to the ImageNet checkpoint of each backbone with the seed fixed at 42. No parameter kept its pretrained value. He initialization therefore has the architecture and the classification head of the ImageNet initialization and differs from it only in its starting weights.

ImageNet initialization used the ConvNeXt weights pretrained with supervision on ImageNet-1k at $224\times224$ pixels~\cite{Deng2009ImageNet}. DINOv3 initialization used the self-supervised DINOv3 ConvNeXt weights, which were distilled from a vision transformer pretrained on about 1.7 billion web images~\cite{Simeoni2025DINOv3}. For the in-domain standard initialization (InDomain-stand), we pretrained a ConvNeXt backbone by supervised multi-label classification of the same five findings on MIMIC-CXR. For the in-domain private initialization (InDomain-priv), we carried out the same pretraining under DP-SGD. The two in-domain initializations exist for ConvNeXt-Small only. ImageNet and DINOv3 received a new head with the default initialization of PyTorch. The two in-domain initializations kept the head pretrained on the five findings in the same order.

\subsection*{Optimization and model selection}

Every model minimized a weighted binary cross-entropy over the five findings,
\begin{equation}
\mathcal{L} = -\frac{1}{5|\mathcal{B}|}\sum_{i\in \mathcal{B}}\sum_{c=1}^{5} m_{ic}\Big[\, w_c\, y_{ic}\log \hat{p}_{ic} + (1-y_{ic})\log(1-\hat{p}_{ic}) \,\Big], \qquad w_c = \frac{N_c - n_c^{+}}{n_c^{+}},
\label{eq:posweight}
\end{equation}
where $\mathcal{B}$ is a batch of radiographs, $\hat{p}_{ic}=1/(1+e^{-z_{ic}})$ is the predicted probability of finding $c$ for radiograph $i$ with logit $z_{ic}$, $y_{ic}$ is its label, $m_{ic}$ is 0 for an excluded CheXpert label and 1 otherwise, $N_c$ is the number of training radiographs with a label for finding $c$, and $n_c^{+}$ is the number of these that are positive~\cite{Rezaei2020Weighted}. For a pooled training set, $N_c$ and $n_c^{+}$ were summed over its datasets. The optimizer was AdamW~\cite{Loshchilov2019Decoupled} with $\beta_1=0.9$, $\beta_2=0.999$, and a weight decay of 0.01 on the trainable parameters. The learning rate was constant during training. Full finetuning used a learning rate of $1\times10^{-5}$, except for the two in-domain initializations under privacy, which used $5\times10^{-6}$.

Non-private models were trained for 20 epochs on shuffled batches of 128 radiographs. The last incomplete batch of each epoch was dropped. Every model was validated after each epoch. The final model of a run is the model of its converged epoch, which is the epoch with the highest validation macro AUROC. For each finding, the decision threshold was the point of the validation receiver operating characteristic curve with the highest Youden's index, which is sensitivity plus specificity minus one~\cite{Unal2017Defining}. The thresholds of the converged epoch were fixed and applied to the test set. Each configuration was trained once. Python, NumPy, and PyTorch were seeded with 42, and cuDNN was restricted to deterministic algorithms.

\subsection*{Differentially private training}

DP bounds the influence of any single training example on the model. A randomized mechanism $\mathcal{M}$ satisfies $(\epsilon,\delta)$-DP if, for every pair of datasets $D$ and $D'$ that differ by the addition or removal of one example and every measurable set of outputs $S$,
\begin{equation}
\Pr[\mathcal{M}(D)\in S] \;\leq\; e^{\epsilon}\,\Pr[\mathcal{M}(D')\in S] + \delta.
\label{eq:dp}
\end{equation}
In this study, one example is one radiograph, both in the finetuning on each target dataset and in the private pretraining on MIMIC-CXR. Because one patient can contribute several radiographs, the guarantees bound the influence of each radiograph on the model and not the influence of each patient.

We trained with DP-SGD~\cite{Abadi2016Deep} as implemented in Opacus~\cite{Yousefpour2021Opacus}. At each step, the per-sample gradients of a sampled batch $\mathcal{B}$ were clipped to a maximum $L_2$ norm $C$, and Gaussian noise proportional to $C$ was added to their sum, giving the privatized gradient
\begin{equation}
\tilde{g} = \frac{1}{L}\left( \sum_{i\in \mathcal{B}} \frac{g_i}{\max\!\left(1,\, \lVert g_i \rVert_2 / C\right)} \;+\; \mathcal{N}\!\left(0,\, \sigma^2 C^2 \mathbf{I}\right) \right),
\label{eq:dpsgd}
\end{equation}
where $g_i$ is the gradient of the loss of radiograph $i$ with respect to all trainable parameters together, $\sigma$ is the noise multiplier, and $L$ is the number of training radiographs divided by the number of steps per epoch, rounded down. Each training radiograph entered a step independently with probability $q=1/\lfloor N/128\rfloor$ for a training set of $N$ radiographs. An epoch had $\lfloor N/128\rfloor$ steps on every training set except VinDr-CXR, where it had 92 steps, one fewer than $\lfloor N/128\rfloor$. The divisor $L$ was therefore 128 on every training set except VinDr-CXR, where it was 130. The privacy loss was tracked with the R\'enyi differential privacy accountant~\cite{Mironov2017Renyi} of Opacus, which used the reciprocal of the number of steps per epoch as the sampling rate. This rate equals $q$ on every training set except VinDr-CXR, where it was 1/92. For each run, $\sigma$ was calibrated by the binary search of Opacus from a target $\epsilon$, $\delta=6\times10^{-6}$, $q$, and a schedule of 150 epochs. With the calibrated $\sigma$, the budget computed for 150 epochs at the rate $q$ would be at most the target $\epsilon$ and within 0.01 of it. The clipping norm was $C=4$ for ConvNeXt-Small and $C=3.5$ for ConvNeXt-Tiny. Each sampled batch was processed in physical batches of at most 64 radiographs. The noise was added once per sampled batch. ConvNeXt contains no batch normalization. The network therefore needed no modification for per-sample gradients. The Gaussian noise was drawn from the standard pseudo-random generator of PyTorch, and the secure mode of Opacus was not enabled.

We used the target values $\epsilon=2$, 5, 8, and 11, which correspond to the privacy budgets $\epsilon\le1$, $\epsilon\le3$, $\epsilon\le6$, and $\epsilon\le9$, respectively. The spent budget was computed after every epoch. Training stopped once it was at least 0.99, 2.99, 5.99, or 8.99, just below the upper end of the respective budget. The achieved $\epsilon$ of a run is the budget spent up to its converged epoch. The converged epoch was chosen after training had ended. The final model therefore satisfies Eq.~\ref{eq:dp} with the $\epsilon$ spent by the whole run and $\delta=6\times10^{-6}$. Rounded to one decimal, as reported throughout, both the achieved $\epsilon$ and the budget spent by the whole run are at most the upper end of the budget of the run.

The reported budgets quantify the privacy of the finetuning stage with respect to the training radiographs of the target dataset. He initialization uses no pretraining, ImageNet and DINOv3 were pretrained on public natural images, and InDomain-stand was pretrained on MIMIC-CXR without a privacy guarantee. For these four initializations, the guarantee therefore covers the finetuning alone. The pretraining of InDomain-priv used the same Opacus implementation of DP-SGD and satisfies a separate guarantee of $\epsilon=7.93$ at $\delta=6\times10^{-6}$ for each MIMIC-CXR radiograph. Both its pretraining and its finetuning are therefore differentially private. Supplementary Table~\ref{stab:accounting} lists the achieved $\epsilon$ and the values of $\sigma$, $q$, and $C$ in Eq.~\ref{eq:dpsgd} for every private run of the main experiment. Supplementary Table~\ref{stab:steps} lists the optimization steps of these runs.

\subsection*{Experimental design}

The main experiment comprised 125 runs. Each finetuned ConvNeXt-Small from one of the five initializations on one of the five datasets, at one of the four privacy budgets or without privacy. Each run trained on one graphics processing unit (GPU). The 125 runs took 1{,}616 GPU hours in total. A private epoch took about twice as long as a non-private epoch on every dataset (Supplementary Table~\ref{stab:compute}). For each epoch time $t$ of a run, we counted the epoch times within $t$ plus or minus 5\% of the median epoch time. The time per epoch of the run is the mean of the epoch times in the window with the largest count.

The parameter-efficient finetuning experiment compared full finetuning with two schemes that train far fewer parameters. It was run for He initialization, ImageNet, and InDomain-stand on VinDr-CXR and CheXpert at $\epsilon\le1$ and $\epsilon\le6$. Head-only finetuning froze the backbone and trained the 3{,}845 parameters of the head, with a learning rate of $2\times10^{-3}$. LoRA~\cite{Hu2022LoRA} froze the backbone and added a trainable low-rank adapter to each of the 72 pointwise linear layers, 2 in each of the 36 blocks of ConvNeXt-Small. Each adapter replaced the output $W_0x$ of a frozen layer by $W_0x + (\alpha/r)\,B\,\mathrm{dropout}(Ax)$. The matrix $A\in\mathbb{R}^{r\times d_{\mathrm{in}}}$ was drawn uniformly from $[-1/\sqrt{d_{\mathrm{in}}},\,1/\sqrt{d_{\mathrm{in}}}]$, and $B\in\mathbb{R}^{d_{\mathrm{out}}\times r}$ was initialized to zero. We used $r=16$, $\alpha=32$, and a dropout probability of 0.05. LoRA trained 2{,}169{,}605 parameters including the head, with a learning rate of $5\times10^{-4}$. The 24 runs of the two schemes were compared with the 12 corresponding runs of the main experiment with full finetuning.

The leave-one-dataset-out experiment held out each dataset in turn, trained on the pooled training data of the other four, and tested on the held-out dataset, for all five initializations at $\epsilon\le1$ and without privacy. Checkpoints and thresholds were selected on the pooled validation data of the four training datasets. The held-out dataset was therefore used only as the test set. In the four folds that trained on VinDr-CXR, the official split of VinDr-CXR was used. Its 15{,}000 training radiographs were used for training. Its 3{,}000 test radiographs served as validation data, because the test set of those folds came from another dataset. The pooled training sets contained 288{,}025 to 410{,}927 radiographs. At $\epsilon\le1$, the models of the folds holding out VinDr-CXR were trained for one epoch, because the budget was spent within the first epoch on their pooled training set of 410{,}927 radiographs. The models of the folds holding out PadChest, ChestX-ray14, CheXpert, and UKA-CXR were trained for 2, 3, 7, and 10 epochs, respectively. The experiment comprised 50 runs.

The capacity experiment comprised 12 runs, which repeated He initialization, ImageNet, and DINOv3 with ConvNeXt-Tiny on CheXpert and VinDr-CXR at $\epsilon\le1$ and without privacy. The four experiments comprised 211 runs in total.

\subsection*{Evaluation metrics}

Each final model was evaluated once on the test set of its dataset, or on the held-out dataset in the leave-one-dataset-out experiment. For each finding, we computed eight metrics: the AUROC, the average precision, the area under the precision-recall curve, accuracy, sensitivity, specificity, the Brier score, and the ECE. Each metric was then averaged over the five findings. We report the average of the average precision as the mean average precision. Accuracy, sensitivity, and specificity used the validation thresholds of the model. The other five metrics do not depend on a threshold.

The average precision is the sum of the precision at each rank of the radiographs ordered by predicted probability, weighted by the increase in recall at that rank. The area under the precision-recall curve was computed by trapezoidal integration. The Brier score is the mean squared difference between the predicted probability and the label~\cite{Brier1950Verification}. The ECE was computed over $M=15$ equal-width bins of the predicted probability as $\sum_{m=1}^{M}(n_m/n)\,\lvert\bar{y}_m-\bar{p}_m\rvert$. Here, $n_m$ of the $n$ radiographs fall in bin $m$, and $\bar{y}_m$ and $\bar{p}_m$ are their fraction of positive labels and their mean predicted probability~\cite{Guo2017Calibration}.

\subsection*{Statistical analysis}

Uncertainty was estimated by a nonparametric bootstrap of the test set with 10{,}000 resamples drawn with seed 42~\cite{Konietschke2014Bootstrapping}. For each dataset, one set of resamples was drawn over the ordered test set and reused by every run. All comparisons on that dataset are therefore paired. In each resample, as many units as the test set contains were drawn with replacement. A unit drawn several times contributed all of its radiographs that many times. The unit was the patient for CheXpert, ChestX-ray14, PadChest, and UKA-CXR and the radiograph for VinDr-CXR. Each metric was summarized by the mean and the standard deviation of its 10{,}000 bootstrap values. This standard deviation, computed with the denominator 9{,}999, is the bootstrap standard error. The 2.5th and 97.5th percentiles of the bootstrap values form the 95\% CI. When a resample contained only one class of a finding, the AUROC, the average precision, and the area under the precision-recall curve of that finding were treated as missing in that resample. The macro average of the resample was then taken over the other findings. For CheXpert, the metrics of atelectasis and pleural effusion were computed over the radiographs with a label for that finding, in the same resamples. All metric values are reported as percentages with one decimal place. Every reported difference, mean over datasets, ratio, rank, and count was computed from these rounded values. The $p$-values were computed from the unrounded bootstrap values.

Initializations were compared on macro AUROC with a two-sided paired bootstrap test. We compared every pair of the five initializations in every dataset and privacy setting of the main experiment. There are 250 such comparisons. For a pair, the 10{,}000 paired differences of macro AUROC were centered by subtracting their mean. The $p$-value is the fraction of centered differences whose magnitude is at least the magnitude of the mean difference. With 10{,}000 resamples, the smallest $p$-value that can be resolved is 0.0001. Smaller values are therefore reported as $p<0.0001$. Other $p$-values below 0.001 are given in scientific notation and the rest with three decimals. The $p$-values were not corrected for multiple comparisons, and the significance level was 0.05. Differences are interpreted by their size and by the CIs of the two compared values. A $p$-value indicates statistical reliability and not clinical importance.

For one dataset and privacy setting, let $A_{\mathrm{ImageNet}}$, $A_{\mathrm{DINOv3}}$, and $A_{\mathrm{stand}}$ be the macro AUROC of ImageNet, DINOv3, and InDomain-stand, so that the objective contrast is $\Delta_{\mathrm{obj}}=A_{\mathrm{DINOv3}}-A_{\mathrm{ImageNet}}$ and the domain contrast is $\Delta_{\mathrm{dom}}=A_{\mathrm{stand}}-A_{\mathrm{ImageNet}}$. Because all initializations are evaluated on the same bootstrap resamples, the paired comparison of InDomain-stand with DINOv3 tests whether the domain contrast exceeds the objective contrast, since in every resample
\begin{equation}
\Delta_{\mathrm{dom}}-\Delta_{\mathrm{obj}} = A_{\mathrm{stand}}-A_{\mathrm{DINOv3}}.
\label{eq:contrasts}
\end{equation}

Demographic subgroups were defined by sex, male or female, and by three age bands, from above 0 to below 40 years, from 40 to below 70 years, and from 70 to below 100 years. For each subgroup, seven of the eight metrics were computed, all except the area under the precision-recall curve. They used the validation thresholds of the model and the same bootstrap resamples, restricted to the radiographs of the subgroup. This conditional bootstrap keeps each subgroup analysis paired with the analysis of the full test set. The age gap of a model is the macro AUROC of its best age band minus that of its worst age band. The sex gap is the absolute difference in macro AUROC between male and female patients. The worst subgroup is the subgroup with the lowest macro AUROC among the five. The subgroup sizes, the per-finding prevalence, and the number of radiographs excluded for a missing attribute are reported in Supplementary Table~\ref{stab:demographics}. These exclusions applied only to the subgroup analyses. Every other evaluation used the complete test set of each dataset. The excluded fraction was below 1\% for every dataset except VinDr-CXR. Of its test radiographs, 58\% lack a recorded sex and 84\% lack an age. The paired tests cover the comparisons of initializations in the main experiment. The comparisons in the parameter-efficient finetuning, leave-one-dataset-out, capacity, and subgroup analyses are descriptive.

\section*{Data availability}

Every dataset of this study except UKA-CXR comes from an existing public release. No images or derived records are redistributed here. VinDr-CXR~\cite{Nguyen2022VinDr,Nguyen2021VinDrPhysioNet} is available from PhysioNet under credentialed access at \url{https://physionet.org/content/vindr-cxr/1.0.0/}. CheXpert~\cite{Irvin2019CheXpert} is available from the Stanford Machine Learning Group at \url{https://stanfordmlgroup.github.io/competitions/chexpert/} on request. Its use is restricted to non-commercial purposes. ChestX-ray14~\cite{Wang2017ChestXray8}, the source of the radiograph in Fig.~\ref{fig:overview}, is openly available from the US National Institutes of Health Clinical Center at \url{https://nihcc.app.box.com/v/ChestXray-NIHCC}. PadChest~\cite{Bustos2020PadChest} is available from the BIMCV repository at \url{https://bimcv.cipf.es/bimcv-projects/padchest/} on request. Its research use agreement permits use for research only and prohibits sharing the data and any attempt at re-identification. UKA-CXR~\cite{Khader2022Artificial} is not publicly available because it contains patient data of University Hospital RWTH Aachen. Access requests can be directed to the corresponding author. A response may take up to 4 weeks. Access requires a written cooperation agreement that limits use to non-commercial academic research and prohibits redistribution and any attempt at re-identification. A subset of UKA-CXR is openly available at \url{https://huggingface.co/TLAIM}. We used MIMIC-CXR~\cite{Johnson2019MIMIC,Johnson2019MIMICCXRJPG} to pretrain the two in-domain initializations. It is available from PhysioNet under credentialed access at \url{https://physionet.org/content/mimic-cxr-jpg/2.1.0/}. Credentialed access to VinDr-CXR and MIMIC-CXR requires a signed data use agreement and completed training in the protection of human research subjects. The agreement permits use for scientific research only and prohibits sharing access to the data and any attempt to identify an individual or an institution. It also requires users who publish results to share the code that produced them in a repository open to the research community.

\section*{Code availability}

The code is publicly available under the MIT License on Zenodo~\cite{TayebiArasteh2026Code} at \url{https://doi.org/10.5281/zenodo.23269417}. The repository contains the code for reading the five datasets, drawing the validation split of VinDr-CXR, finetuning every model with and without DP, evaluating it, and computing the reported statistics. Its configuration fixes the training, splitting, and bootstrap seeds. It does not contain model weights or images.

All models were trained and evaluated on local hardware, without any cloud service or third-party application programming interface. The ImageNet and DINOv3 initializations are open-weight checkpoints on Hugging Face. The DINOv3 checkpoints require accepting their terms. We pretrained the two in-domain initializations on MIMIC-CXR. All models were accessed and all experiments were run between October 2025 and October 2026. The public checkpoints are available at the following addresses:

\begingroup
\footnotesize
\sloppy
\vspace{0.4em}
\textbf{Supervised ImageNet initialization:}
\begin{itemize}
  \item ConvNeXt-Small: \url{https://huggingface.co/facebook/convnext-small-224}
  \item ConvNeXt-Tiny: \url{https://huggingface.co/facebook/convnext-tiny-224}
\end{itemize}
\textbf{Self-supervised DINOv3 initialization:}
\begin{itemize}
  \item ConvNeXt-Small: \url{https://huggingface.co/facebook/dinov3-convnext-small-pretrain-lvd1689m}
  \item ConvNeXt-Tiny: \url{https://huggingface.co/facebook/dinov3-convnext-tiny-pretrain-lvd1689m}
\end{itemize}
\endgroup
The experiments used two Python environments. The first had Python~3.10 with PyTorch~2.5 (CUDA~12.4), torchvision~0.20, transformers~4.56, NumPy~2.2, SciPy~1.15, and scikit-learn~1.7. The second had Python~3.11 with PyTorch~2.9 (CUDA~13.0), torchvision~0.24, transformers~5.0, SciPy~1.17, and scikit-learn~1.8. Both environments used Opacus~1.5 for DP. The figures were produced with Matplotlib~3.10. Training and evaluation ran in 32-bit floating point precision on NVIDIA L40S GPUs with 48\,GB of memory and Intel Xeon Silver 4310 processors.

\section*{Funding}

S.T.A.\ is supported by the German Federal Ministry of Research, Technology and Space (ARISTOTLE, 01ZU2602B) and the Excellence Strategy of the German Federal Government, the L\"ander, and RWTH ERS (START\_526-26). S.N.\ is supported by the DFG (701010997, 517243167). D.T.\ is supported by the German Federal Ministry of Research, Technology and Space (TRANSFORM LIVER, 031L0312C; DECIPHER-M, 01KD2420B), the DFG (515639690), and the European Union (Horizon Europe, ODELIA, GA 101057091; ERC Starting Grant SAGMA, GA 101222556).

\section*{Author contributions}

The formal analysis and conceptualization were conducted by S.T.A.\ and D.T. The original draft was written by S.T.A., M.F., and M.L. The code was developed by S.T.A. The experiments were performed by S.T.A. The illustrations were designed by S.T.A. The statistical analyses were performed by S.T.A., M.F., S.N., and D.T. B.H.P., C.K., S.N., and D.T.\ provided clinical expertise. S.T.A., M.F., M.L., J.B., M.A., M.R., and D.T.\ provided technical expertise. The study was defined by S.T.A. All authors read the manuscript, agreed to the submission of this paper, and contributed to the editing.

\section*{Competing interests}

S.T.A.\ received honoraria for lectures from RWTH International Academy gGmbH, Germany and is on the editorial boards of Communications Medicine and of European Radiology Experimental, and on the trainee editorial board of Radiology: Artificial Intelligence. M.L.\ is employed by Generali Deutschland Services GmbH, Germany, and is on the editorial board of European Radiology Experimental. B.H.P.\ is an associate editor at the Journal of Medical Internet Research. D.T.\ received honoraria for lectures from Bayer, GE, Roche, AstraZeneca, and Philips and holds shares in StratifAI GmbH, Germany, and in Synagen GmbH, Germany. The authors declare no other competing financial or non-financial interests.

\bibliographystyle{splncs04}
\bibliography{bibliography}

\setcounter{table}{0}
\setcounter{figure}{0}
\setcounter{equation}{0}
\renewcommand{\tablename}{Supplementary Table}
\renewcommand{\figurename}{Supplementary Fig.}
\floatname{algorithm}{Supplementary Algorithm}
\renewcommand{\thealgorithm}{\arabic{algorithm}}
\renewcommand{\theequation}{S\arabic{equation}}
\newcounter{snote}
\renewcommand{\thesnote}{\arabic{snote}}

\section*{Supplementary information}

\refstepcounter{snote}
\section*{Supplementary Note \thesnote: Dataset curation and label harmonization}
\label{snote:curation}

\subsection*{Label harmonization}

Each dataset was reduced to five findings: atelectasis, cardiomegaly, pleural effusion, pneumonia, and no finding. ChestX-ray14~\cite{Wang2017ChestXray8} names pleural effusion as effusion. Pneumonic infiltrates in UKA-CXR~\cite{Khader2022Artificial} were counted as pneumonia. A healthy examination in UKA-CXR was counted as no finding. UKA-CXR grades cardiomegaly for the whole heart and grades atelectasis, pleural effusion, and pneumonic infiltrates separately for each side of the chest. Each of these findings was counted as present at grade 3 or 4. A finding graded per side was counted as present when either side had one of these grades. VinDr-CXR~\cite{Nguyen2022VinDr}, ChestX-ray14, and PadChest~\cite{Bustos2020PadChest} provide binary labels.

CheXpert~\cite{Irvin2019CheXpert} labels come from a rule-based report labeler that marks each finding as present, absent, uncertain, or not mentioned. A finding not mentioned in the report was counted as absent. The curated CheXpert list does not contain the 39{,}323 training radiographs of the release that have an uncertain label for cardiomegaly, lung opacity, lung lesion, edema, or pneumonia. Of these radiographs, 33{,}351 are frontal. Cardiomegaly and pneumonia therefore have no uncertain label in this study. For atelectasis and pleural effusion, a radiograph with an uncertain label was excluded from the training loss, the validation, the decision threshold, and the evaluation of that finding alone. This applied to 15{,}330 training, 1{,}641 validation, and 3{,}597 test radiographs for atelectasis and to 5{,}143, 597, and 1{,}387 radiographs for pleural effusion.

\subsection*{Projection filtering}

CheXpert and PadChest were filtered by projection before any split was used. The filter therefore applies to the training, validation, and test data alike. Of the 184{,}325 radiographs of the curated CheXpert list, we kept the 157{,}878 frontal radiographs and discarded the 26{,}447 lateral ones. Of the 160{,}704 radiographs of the curated PadChest list, we kept the 110{,}525 posteroanterior, anteroposterior, and horizontal anteroposterior radiographs and discarded the other 50{,}179, of which 49{,}538 are lateral. VinDr-CXR, ChestX-ray14, and UKA-CXR were not filtered by projection.

\subsection*{Splits}

The curated lists of CheXpert, ChestX-ray14, PadChest, and UKA-CXR assign every patient to one of the training, validation, and test splits. We used these assignments unchanged. ChestX-ray14 keeps its official test split. In each of these four datasets, the validation split contains 10\% of the radiographs outside the test split. No patient appears in two splits. VinDr-CXR was released with 15{,}000 training and 3{,}000 test radiographs, without a validation split and without patient identifiers. We drew 3{,}000 of its training radiographs as the validation set, stratified by the combination of the five labels with seed 42. The official test split was kept unchanged. Supplementary Table~\ref{stab:datasets} lists the size of every split.

\subsection*{Demographic subgroups}

Sex is recorded as male or female. A radiograph with another or no recorded sex belongs to no sex subgroup. VinDr-CXR records sex for 8{,}608 of its 18{,}000 radiographs. Age was grouped into three bands, from above 0 to below 40 years, from 40 to below 70 years, and from 70 to below 100 years. VinDr-CXR records an age of 0 for 13{,}766 of its radiographs. These radiographs fall into no age band. A radiograph outside a subgroup still contributes to every other result. Supplementary Table~\ref{stab:demographics} lists the size of every subgroup of every test set and the number of test radiographs without a sex or an age band.

\subsection*{Class prevalence}

Supplementary Table~\ref{stab:prevalence} lists the number of positive radiographs of every finding in every split. The training counts are the numbers $n_c^{+}$ of the positive weight in Eq.~\ref{eq:posweight}. For CheXpert, the percentage of atelectasis and pleural effusion is computed over the radiographs with a label for that finding.

\clearpage

\begin{figure*}[p]
\centering
\footnotesize
\includegraphics[width=\textwidth]{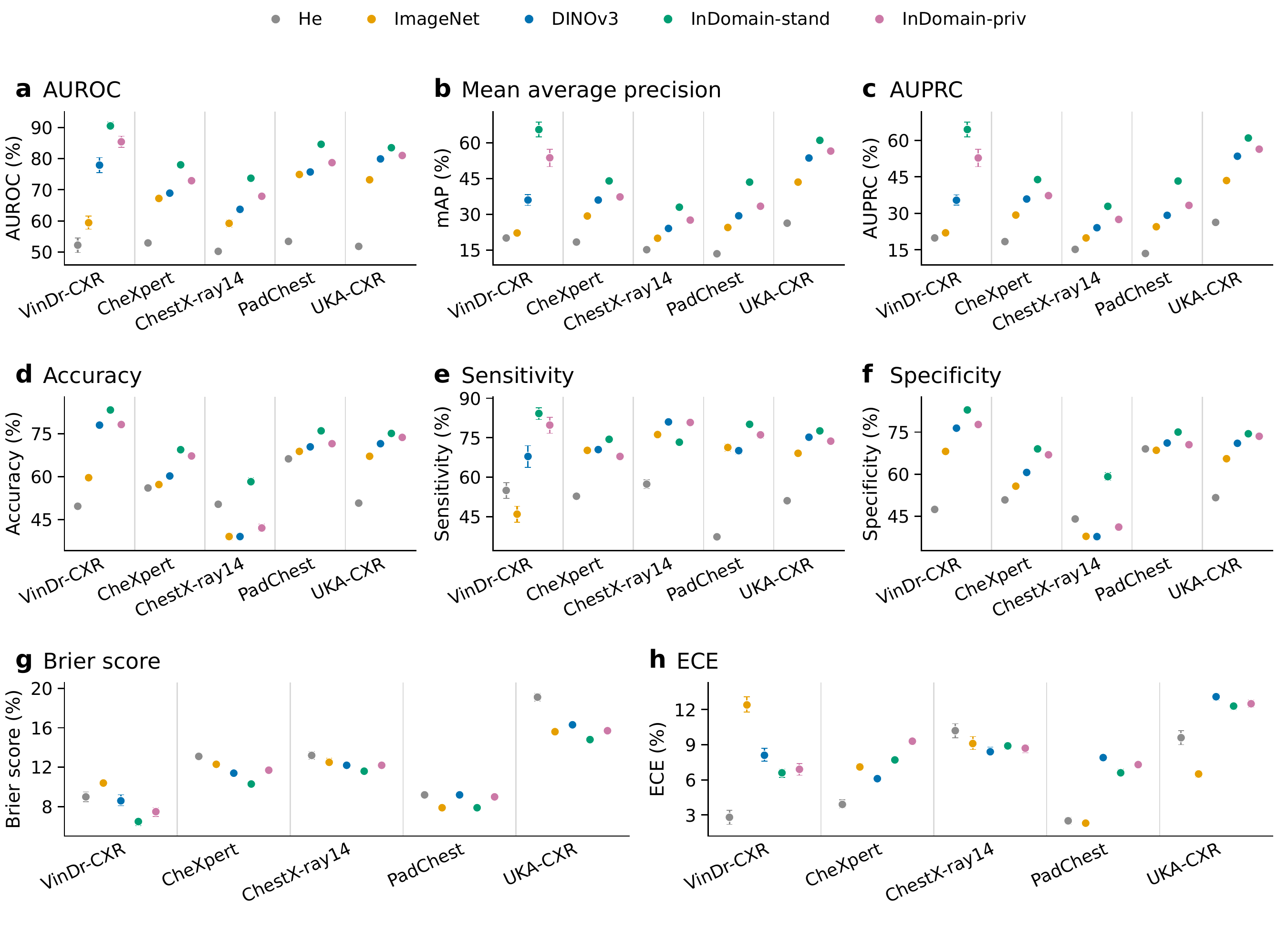}
\caption{Eight evaluation metrics of ConvNeXt-Small~\cite{Liu2022ConvNet} classifiers finetuned with differentially private stochastic gradient descent~\cite{Abadi2016Deep} at the strictest privacy budget, $\epsilon\le1$, on the test sets of VinDr-CXR~\cite{Nguyen2022VinDr}, CheXpert~\cite{Irvin2019CheXpert}, ChestX-ray14~\cite{Wang2017ChestXray8}, PadChest~\cite{Bustos2020PadChest}, and UKA-CXR~\cite{Khader2022Artificial} ($n = 120{,}068$ test radiographs). Each metric is averaged over the five findings. Color marks the initialization. A marker is the mean of 10{,}000 bootstrap resamples of the test set, and its error bar is the 95\% CI, from the 2.5th to the 97.5th percentile. Higher is better for every metric except the Brier score and ECE. Accuracy, sensitivity, and specificity use the per-finding thresholds selected on the validation set. Counts per panel: 25 markers, for 5 datasets and 5 initializations. \textbf{a}--\textbf{h}, AUROC, mean average precision, AUPRC, accuracy, sensitivity, specificity, Brier score, and ECE. Supplementary Fig.~\ref{sfig:metricsnp} plots the same metrics without privacy. AUPRC, area under the precision-recall curve; AUROC, area under the receiver operating characteristic curve; CI, confidence interval; ECE, expected calibration error; $\epsilon$, privacy budget; He, He initialization~\cite{He2015Delving}; ImageNet, supervised pretraining on ImageNet~\cite{Deng2009ImageNet}; DINOv3, self-supervised DINOv3 pretraining~\cite{Simeoni2025DINOv3}; InDomain-stand and InDomain-priv, supervised pretraining on MIMIC-CXR~\cite{Johnson2019MIMIC} without and with differential privacy; mAP, mean average precision.}
\label{sfig:metrics}
\end{figure*}

\begin{figure*}[p]
\centering
\footnotesize
\includegraphics[width=\textwidth]{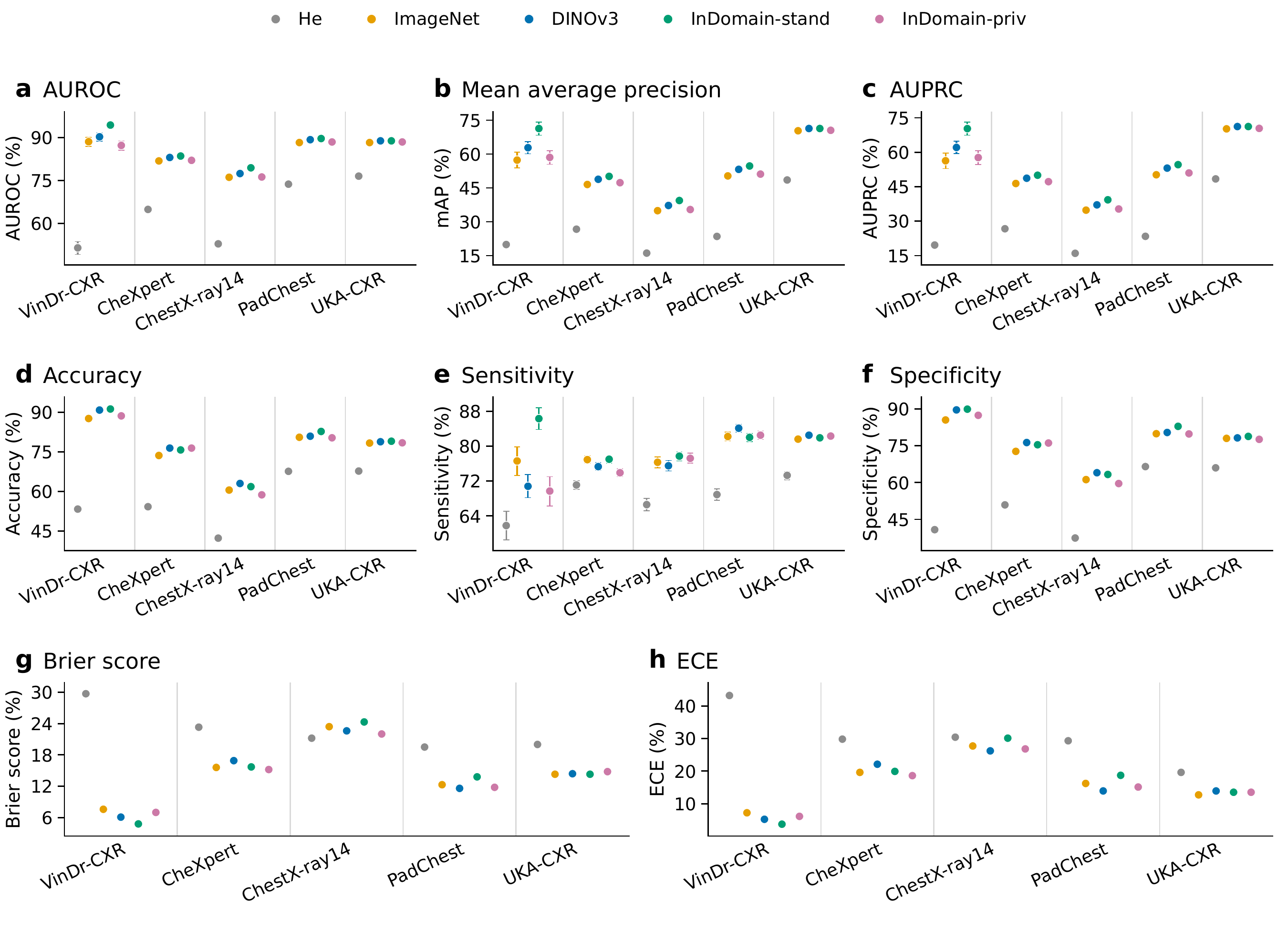}
\caption{Eight evaluation metrics of ConvNeXt-Small~\cite{Liu2022ConvNet} classifiers finetuned without privacy on the test sets of VinDr-CXR~\cite{Nguyen2022VinDr}, CheXpert~\cite{Irvin2019CheXpert}, ChestX-ray14~\cite{Wang2017ChestXray8}, PadChest~\cite{Bustos2020PadChest}, and UKA-CXR~\cite{Khader2022Artificial} ($n = 120{,}068$ test radiographs). Each metric is averaged over the five findings. Color marks the initialization. A marker is the mean of 10{,}000 bootstrap resamples of the test set, and its error bar is the 95\% CI, from the 2.5th to the 97.5th percentile. Higher is better for every metric except the Brier score and ECE. Accuracy, sensitivity, and specificity use the per-finding thresholds selected on the validation set. Counts per panel: 25 markers, for 5 datasets and 5 initializations. \textbf{a}--\textbf{h}, AUROC, mean average precision, AUPRC, accuracy, sensitivity, specificity, Brier score, and ECE. Supplementary Fig.~\ref{sfig:metrics} plots the same metrics at $\epsilon\le1$. AUPRC, area under the precision-recall curve; AUROC, area under the receiver operating characteristic curve; CI, confidence interval; ECE, expected calibration error; $\epsilon$, privacy budget; He, He initialization~\cite{He2015Delving}; ImageNet, supervised pretraining on ImageNet~\cite{Deng2009ImageNet}; DINOv3, self-supervised DINOv3 pretraining~\cite{Simeoni2025DINOv3}; InDomain-stand and InDomain-priv, supervised pretraining on MIMIC-CXR~\cite{Johnson2019MIMIC} without and with differential privacy; mAP, mean average precision.}
\label{sfig:metricsnp}
\end{figure*}

\begin{figure*}[p]
\centering
\footnotesize
\includegraphics[width=\textwidth]{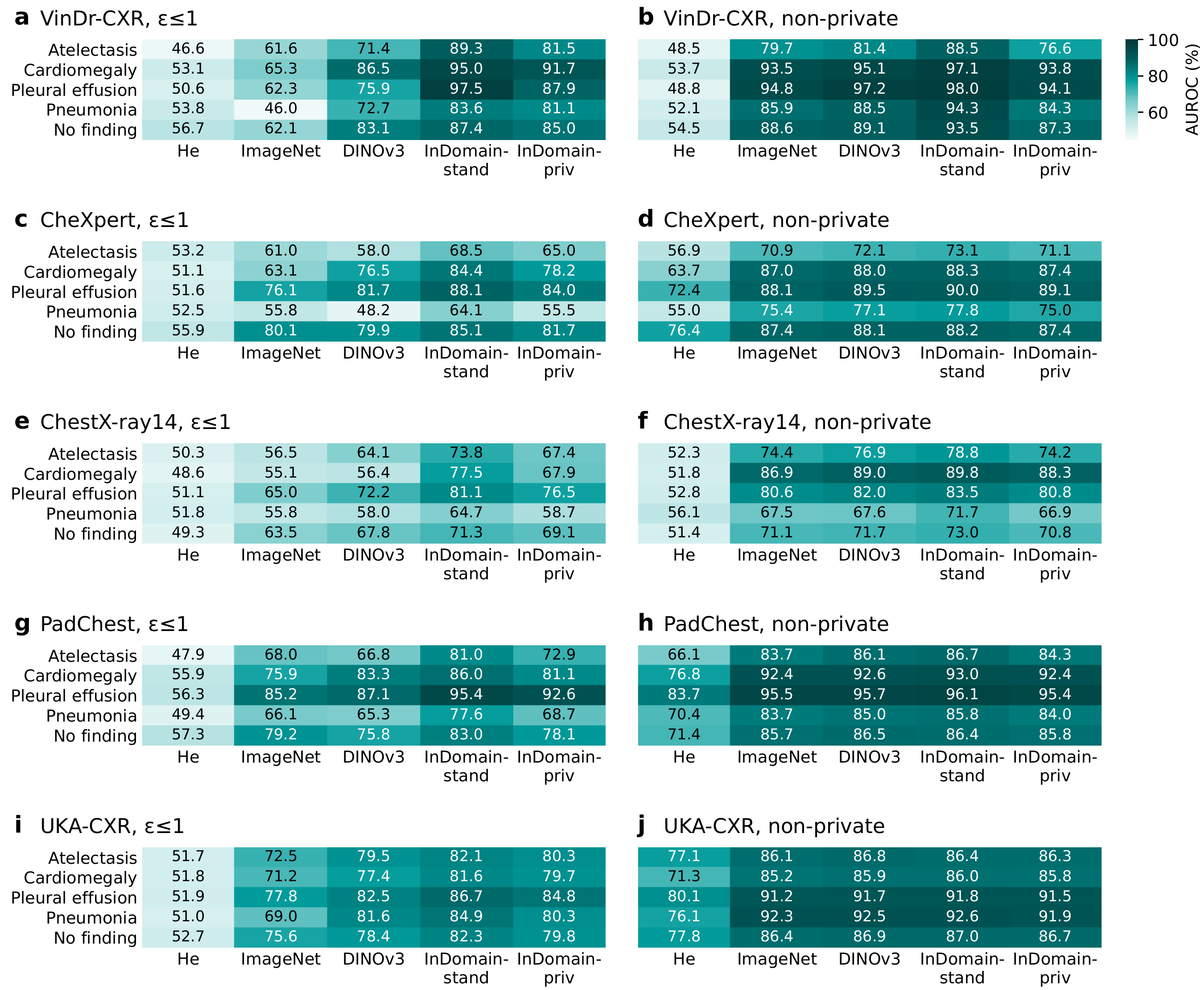}
\caption{AUROC of each finding for ConvNeXt-Small~\cite{Liu2022ConvNet} classifiers at the strictest privacy budget, $\epsilon\le1$, under differentially private stochastic gradient descent~\cite{Abadi2016Deep}, and without privacy, on the test sets of VinDr-CXR~\cite{Nguyen2022VinDr}, CheXpert~\cite{Irvin2019CheXpert}, ChestX-ray14~\cite{Wang2017ChestXray8}, PadChest~\cite{Bustos2020PadChest}, and UKA-CXR~\cite{Khader2022Artificial} ($n = 120{,}068$ test radiographs). Each entry is the mean AUROC in percent over 10{,}000 bootstrap resamples of the test set, and its color encodes the same value on a scale shared by all panels. Columns are the initializations, and rows are the findings. No panel has an error bar. For CheXpert, a radiograph with an uncertain label for atelectasis or pleural effusion does not enter the AUROC of that finding. Counts per panel: 25 entries, for 5 findings and 5 initializations. \textbf{a}, \textbf{c}, \textbf{e}, \textbf{g}, \textbf{i}, VinDr-CXR, CheXpert, ChestX-ray14, PadChest, and UKA-CXR at $\epsilon\le1$. \textbf{b}, \textbf{d}, \textbf{f}, \textbf{h}, \textbf{j}, The same datasets without privacy. AUROC, area under the receiver operating characteristic curve; $\epsilon$, privacy budget; He, He initialization~\cite{He2015Delving}; ImageNet, supervised pretraining on ImageNet~\cite{Deng2009ImageNet}; DINOv3, self-supervised DINOv3 pretraining~\cite{Simeoni2025DINOv3}; InDomain-stand and InDomain-priv, supervised pretraining on MIMIC-CXR~\cite{Johnson2019MIMIC} without and with differential privacy.}
\label{sfig:findings}
\end{figure*}

\begin{figure*}[p]
\centering
\footnotesize
\includegraphics[width=\textwidth]{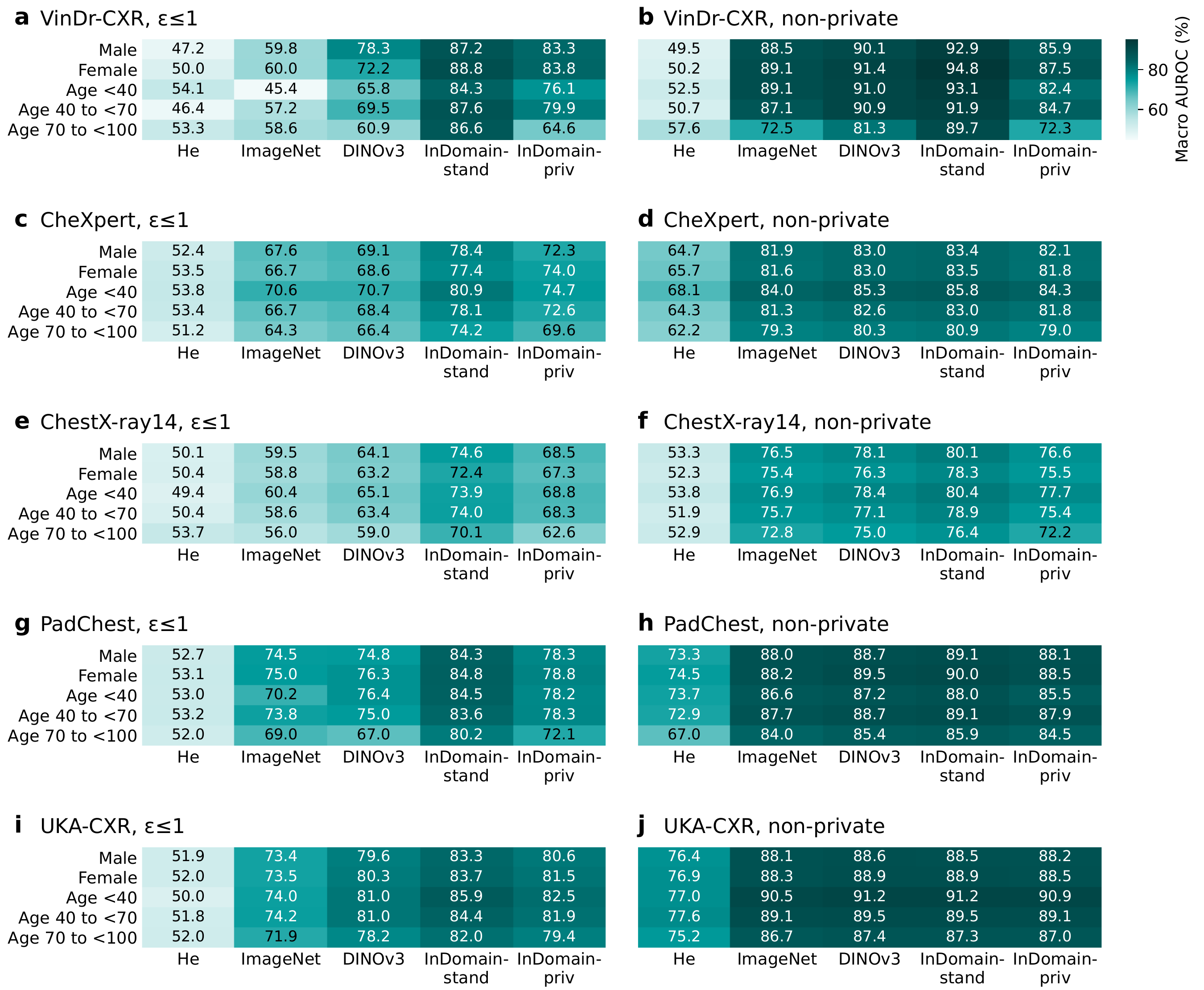}
\caption{Macro AUROC within each demographic subgroup for ConvNeXt-Small~\cite{Liu2022ConvNet} classifiers at the strictest privacy budget, $\epsilon\le1$, under differentially private stochastic gradient descent~\cite{Abadi2016Deep}, and without privacy, on the test sets of VinDr-CXR~\cite{Nguyen2022VinDr}, CheXpert~\cite{Irvin2019CheXpert}, ChestX-ray14~\cite{Wang2017ChestXray8}, PadChest~\cite{Bustos2020PadChest}, and UKA-CXR~\cite{Khader2022Artificial} ($n = 120{,}068$ test radiographs). Each entry is the mean macro AUROC in percent over 10{,}000 bootstrap resamples of the test set, restricted to the radiographs of the subgroup. Its color encodes the same value on a scale shared by all panels. Columns are the initializations, and rows are the subgroups. No panel has an error bar. The subgroups are male and female patients and the age bands from above 0 to below 40, from 40 to below 70, and from 70 to below 100 years. A radiograph without a recorded attribute belongs to no subgroup of that attribute. Supplementary Table~\ref{stab:demographics} lists the size of every subgroup. Counts per panel: 25 entries, for 5 subgroups and 5 initializations. \textbf{a}, \textbf{c}, \textbf{e}, \textbf{g}, \textbf{i}, VinDr-CXR, CheXpert, ChestX-ray14, PadChest, and UKA-CXR at $\epsilon\le1$. \textbf{b}, \textbf{d}, \textbf{f}, \textbf{h}, \textbf{j}, The same datasets without privacy. AUROC, area under the receiver operating characteristic curve; $\epsilon$, privacy budget; He, He initialization~\cite{He2015Delving}; ImageNet, supervised pretraining on ImageNet~\cite{Deng2009ImageNet}; DINOv3, self-supervised DINOv3 pretraining~\cite{Simeoni2025DINOv3}; InDomain-stand and InDomain-priv, supervised pretraining on MIMIC-CXR~\cite{Johnson2019MIMIC} without and with differential privacy.}
\label{sfig:subgroups}
\end{figure*}

\clearpage

\begin{table}[t]
\centering
\footnotesize
\caption{Glossary of the terms of this study, grouped by data, initialization, contrast, privacy, finetuning, metric, and reporting convention. Every metric is a percentage computed on the test set of a dataset. The tables omit the percent sign.}
\label{stab:glossary}
\scriptsize
\setlength{\tabcolsep}{3pt}
\renewcommand{\arraystretch}{1.0}
\begin{tabular}{@{}>{\raggedright\arraybackslash}p{0.25\textwidth}>{\raggedright\arraybackslash}p{0.73\textwidth}@{}}
\toprule
Term & Definition \\
\midrule
\multicolumn{2}{@{}l}{\textbf{Data}} \\
Datasets & VinDr-CXR~\cite{Nguyen2022VinDr}, CheXpert~\cite{Irvin2019CheXpert}, ChestX-ray14~\cite{Wang2017ChestXray8}, PadChest~\cite{Bustos2020PadChest}, and UKA-CXR~\cite{Khader2022Artificial}, with 591{,}884 radiographs in total. Each dataset is split into training, validation, and test sets. \\
MIMIC-CXR & The 153{,}255 frontal radiographs of the training split of MIMIC-CXR~\cite{Johnson2019MIMIC}, used only to pretrain the two in-domain initializations. \\
Findings & The five labels of every radiograph: atelectasis, cardiomegaly, pleural effusion, pneumonia, and no finding. \\
\midrule
\multicolumn{2}{@{}l}{\textbf{Initializations}} \\
He initialization (He) & Training from scratch. Every convolutional and linear weight is drawn from the He (Kaiming) normal distribution~\cite{He2015Delving}. No weight comes from a trained model. \\
ImageNet & ConvNeXt~\cite{Liu2022ConvNet} weights pretrained with labels on ImageNet-1k~\cite{Deng2009ImageNet}. \\
DINOv3 & Self-supervised DINOv3 ConvNeXt weights~\cite{Simeoni2025DINOv3}, distilled from a vision transformer pretrained without labels on about 1.7 billion web images. \\
InDomain-stand & A ConvNeXt-Small pretrained with labels on the five findings of MIMIC-CXR, without privacy. \\
InDomain-priv & The same pretraining under DP-SGD, with $\epsilon=7.93$ at $\delta=6\times10^{-6}$ for each MIMIC-CXR radiograph. \\
\midrule
\multicolumn{2}{@{}l}{\textbf{Contrasts}} \\
Objective contrast ($\Delta_{\mathrm{obj}}$) & Macro AUROC of DINOv3 minus that of ImageNet. Both initializations were pretrained on natural images. \\
Domain contrast ($\Delta_{\mathrm{dom}}$) & Macro AUROC of InDomain-stand minus that of ImageNet. Both initializations were pretrained with labels. \\
Difference of the contrasts & $\Delta_{\mathrm{dom}}-\Delta_{\mathrm{obj}}$, which equals InDomain-stand minus DINOv3 in every bootstrap resample (Eq.~\ref{eq:contrasts}). \\
\midrule
\multicolumn{2}{@{}l}{\textbf{Privacy}} \\
Differential privacy (DP) & A guarantee that adding or removing one training radiograph changes the probability of any output by at most a factor of $e^{\epsilon}$, plus $\delta$ (Eq.~\ref{eq:dp}). \\
Differentially private stochastic gradient descent (DP-SGD) & Training in which the gradient of each radiograph is clipped to the norm $C$ and Gaussian noise of standard deviation $\sigma C$ is added to their sum~\cite{Abadi2016Deep} (Eq.~\ref{eq:dpsgd}). \\
Privacy budget & The upper end of the $\epsilon$ of a private run, written $\epsilon\le1$, $\epsilon\le3$, $\epsilon\le6$, or $\epsilon\le9$, at $\delta=6\times10^{-6}$. \\
Target $\epsilon$ & The value of 2, 5, 8, or 11 for which the noise multiplier $\sigma$ was calibrated over 150 epochs. \\
Converged epoch & The epoch with the highest validation macro AUROC. Its model is the final model of the run. \\
Achieved $\epsilon$ & The privacy budget spent up to the converged epoch, given at one decimal. \\
\midrule
\multicolumn{2}{@{}l}{\textbf{Finetuning}} \\
ConvNeXt-Small and ConvNeXt-Tiny & The main backbone with 49.5 million parameters and the smaller backbone with 27.8 million, which is used in the capacity experiment. \\
Full finetuning & Training of every parameter of the network and its classification head. \\
Head-only finetuning & Training of the 3{,}845 parameters of the classification head of a frozen backbone. \\
Low-rank adaptation (LoRA) & Training of low-rank adapters~\cite{Hu2022LoRA} on the 72 pointwise linear layers of a frozen backbone, together with the head, 2{,}169{,}605 parameters in total. \\
Leave-one-dataset-out & Training on the pooled training data of four datasets and testing on the fifth. \\
\midrule
\multicolumn{2}{@{}l}{\textbf{Metrics}} \\
Macro AUROC & The area under the receiver operating characteristic curve (AUROC) averaged over the five findings, in percent. \\
Sex gap & The absolute difference in macro AUROC between male and female patients. \\
Age gap & The macro AUROC of the best minus that of the worst of the age bands from above 0 to below 40, from 40 to below 70, and from 70 to below 100 years. \\
Worst subgroup & The subgroup with the lowest macro AUROC among male patients, female patients, and the three age bands. \\
\midrule
\multicolumn{2}{@{}l}{\textbf{Reporting}} \\
Mean $\pm$ standard deviation [95\% confidence interval (CI)] & The mean, the standard deviation, and the 2.5th to 97.5th percentiles of a metric over 10{,}000 bootstrap resamples of the test set, drawn with seed 42. \\
Resampling unit & The patient for CheXpert, ChestX-ray14, PadChest, and UKA-CXR, and the radiograph for VinDr-CXR. \\
Paired bootstrap test & A two-sided test of the 10{,}000 paired differences in macro AUROC at the 0.05 level, without correction for multiple comparisons. The smallest $p$-value that can be resolved is 0.0001. \\
\bottomrule
\end{tabular}
\end{table}

\begin{table}[t]
\centering
\footnotesize
\caption{Summary of the findings, in the order in which they are reported. Values are taken unchanged from the main text. Every value is a macro AUROC or a difference in macro AUROC in percent, unless the row names another quantity. The $p$-values are from the two-sided paired bootstrap test. The datasets are VinDr-CXR~\cite{Nguyen2022VinDr}, CheXpert~\cite{Irvin2019CheXpert}, ChestX-ray14~\cite{Wang2017ChestXray8}, PadChest~\cite{Bustos2020PadChest}, and UKA-CXR~\cite{Khader2022Artificial}. AUPRC, area under the precision-recall curve; AUROC, area under the receiver operating characteristic curve; ECE, expected calibration error; $\epsilon$, privacy budget; He, He initialization~\cite{He2015Delving}; ImageNet, supervised pretraining on ImageNet~\cite{Deng2009ImageNet}; DINOv3, self-supervised DINOv3 pretraining~\cite{Simeoni2025DINOv3}; InDomain-stand and InDomain-priv, supervised pretraining on MIMIC-CXR~\cite{Johnson2019MIMIC} without and with differential privacy; LoRA, low-rank adaptation~\cite{Hu2022LoRA}.}
\label{stab:findings}
\scriptsize
\setlength{\tabcolsep}{3pt}
\renewcommand{\arraystretch}{1.0}
\begin{tabular}{@{}>{\raggedright\arraybackslash}p{0.115\textwidth}>{\raggedright\arraybackslash}p{0.155\textwidth}>{\raggedright\arraybackslash}p{0.48\textwidth}>{\raggedright\arraybackslash}p{0.19\textwidth}@{}}
\toprule
Analysis & Comparison & Result & Interpretation \\
\midrule
Privacy-utility frontier & Five initializations on five datasets at every privacy setting & InDomain-stand highest in 24 of 25 combinations and equal to DINOv3 on UKA-CXR without privacy (88.8 each, $p=0.562$). Lead over ImageNet and DINOv3 of 2.6 and 1.4 without privacy and of 15.3 and 8.8 at $\epsilon\le1$ & In-domain pretraining sets the frontier. Its lead grows under privacy \\
\midrule
Loss under privacy & Non-private training against $\epsilon\le1$ & Decrease of 5.1 for InDomain-stand, 7.3 for InDomain-priv, 12.5 for DINOv3, and 17.8 for ImageNet. He initialization at 50.2 to 56.5 at every privacy budget & The domain-matched initialization loses the least \\
\midrule
Dataset size & InDomain-stand minus ImageNet in each dataset & 31.1 at $\epsilon\le1$ and 5.8 without privacy on VinDr-CXR, the smallest dataset. Between 9.7 and 14.5 at $\epsilon\le1$ and between 0.6 and 3.3 without privacy on the other four datasets & The in-domain lead is largest on the smallest dataset \\
\midrule
Other metrics & Eight metrics at $\epsilon\le1$, averaged over the datasets & Best AUROC, AUPRC, mean average precision, accuracy, sensitivity, specificity, and Brier score for InDomain-stand. Its ECE of 8.4 is above those of ImageNet (7.5) and He initialization (5.8) & The advantage holds for seven of the eight metrics \\
\midrule
Objective and domain & The two contrasts and their difference & Both contrasts positive in all 25 combinations. Domain contrast larger in 24 of 25, with a mean 2.2 times the mean objective contrast without privacy and 2.4 to 3.7 times under privacy. Difference of 2.1 to 12.6 under privacy, each of the 20 with $p<0.0001$ & Changing the domain improves macro AUROC more than changing the objective at every privacy budget \\
\midrule
Privacy spent & Converged epoch and achieved $\epsilon$ at $\epsilon\le1$ & Converged epoch of 18.6, 22.2, 37.0, 38.4, and 39.4, and achieved $\epsilon$ at 67\%, 77\%, 96\%, 98\%, and 99\% of the upper end of the budget, for InDomain-stand, InDomain-priv, DINOv3, He initialization, and ImageNet & The domain-matched initializations converge earlier and spend less \\
\midrule
Private pretraining & InDomain-priv against InDomain-stand and the generic initializations & 4.9 below InDomain-stand at $\epsilon\le1$, with $p<0.0001$ in all 25 comparisons. At $\epsilon\le1$, above DINOv3, ImageNet, and He initialization by 4.0, 10.4, and 25.1. Above DINOv3 in 18 of the 20 private combinations and 1.3 below it without privacy & Private pretraining outperforms generic pretraining in most private settings \\
\midrule
Parameter-efficient finetuning & Head-only finetuning and LoRA against full finetuning & LoRA higher in 11 of 12 configurations (mean 4.3, largest 16.3) and head-only finetuning in 10 of 12 (mean 2.1). At $\epsilon\le1$, InDomain-stand 1.8 and 2.5 below its non-private reference on VinDr-CXR and CheXpert with LoRA and 3.8 and 5.5 below it with full finetuning & LoRA reduces the loss in macro AUROC of InDomain-stand by about half and keeps the order of the initializations \\
\midrule
Unseen institutions & Leave-one-dataset-out at $\epsilon\le1$ & Pretrained initializations 3.9 to 6.6 lower on a held-out dataset than after training on it. InDomain-stand first on four of the five held-out datasets, with a mean of 76.5. With UKA-CXR held out, ImageNet at 67.4 and InDomain-stand at 66.2 & The advantage persists at unseen institutions, except UKA-CXR \\
\midrule
Backbone capacity & ConvNeXt-Small against ConvNeXt-Tiny & Backbones differ by 1.7 on average and by at most 6.6. Within ConvNeXt-Tiny, the initializations differ by up to 20.4 on VinDr-CXR and 19.0 on CheXpert at $\epsilon\le1$ & Initialization changes macro AUROC more than capacity \\
\midrule
Demographic subgroups & Sex and three age bands at $\epsilon\le1$ & Age gaps of 3.6 to 7.2 and sex gaps of 0.5 to 1.9. Worst-subgroup macro AUROC of 78.2 for InDomain-stand and at most 69.7 for the other initializations, averaged over the datasets. InDomain-stand highest in the worst subgroup of every dataset. Oldest band the worst subgroup in 20 of 25 combinations & In-domain pretraining gives the highest macro AUROC in the worst subgroup under privacy \\
\bottomrule
\end{tabular}
\end{table}

\begin{table}[t]
\centering
\footnotesize
\caption{Privacy accounting of the 100 differentially private runs of the main experiment. The Poisson sampling rate, the noise multiplier, and the per-sample clipping norm depend on the dataset and the privacy budget and are identical for the five initializations. They are therefore listed once. On VinDr-CXR, the accountant computed the achieved $\epsilon$ at a sampling rate of 1/92, the reciprocal of its 92 steps per epoch. The achieved $\epsilon$ is the budget spent up to the epoch whose model was kept and is given for each initialization at one decimal. Rows are grouped by dataset and ordered by privacy budget, from the strictest to the loosest. Every run used $\delta=6\times10^{-6}$ and a noise multiplier calibrated for 150 epochs. Every model is a ConvNeXt-Small~\cite{Liu2022ConvNet} finetuned with differentially private stochastic gradient descent~\cite{Abadi2016Deep} on VinDr-CXR~\cite{Nguyen2022VinDr}, CheXpert~\cite{Irvin2019CheXpert}, ChestX-ray14~\cite{Wang2017ChestXray8}, PadChest~\cite{Bustos2020PadChest}, and UKA-CXR~\cite{Khader2022Artificial}. $\epsilon$, privacy budget; He, He initialization~\cite{He2015Delving}; ImageNet, supervised pretraining on ImageNet~\cite{Deng2009ImageNet}; DINOv3, self-supervised DINOv3 pretraining~\cite{Simeoni2025DINOv3}; InDomain-stand and InDomain-priv, supervised pretraining on MIMIC-CXR~\cite{Johnson2019MIMIC} without and with differential privacy.}
\label{stab:accounting}
\scriptsize
\setlength{\tabcolsep}{4pt}
\begin{tabular}{lccc ccccc}
\toprule
 & & & & \multicolumn{5}{c}{Achieved $\epsilon$} \\
\cmidrule(l){5-9}
Privacy budget & \begin{tabular}[b]{@{}c@{}}Sampling\\rate\end{tabular} & \begin{tabular}[b]{@{}c@{}}Noise\\multiplier\end{tabular} & \begin{tabular}[b]{@{}c@{}}Clipping\\norm\end{tabular} & He & ImageNet & DINOv3 & InDomain-stand & InDomain-priv \\
\midrule
\multicolumn{9}{l}{\textbf{VinDr-CXR}} \\
$\epsilon\le1$ & 0.01075 & 2.910 & 4.0 & 1.0 & 1.0 & 1.0 & 0.2 & 1.0 \\
$\epsilon\le3$ & 0.01075 & 1.426 & 4.0 & 3.0 & 3.0 & 2.9 & 0.6 & 3.0 \\
$\epsilon\le6$ & 0.01075 & 1.067 & 4.0 & 6.0 & 6.0 & 5.7 & 5.7 & 5.7 \\
$\epsilon\le9$ & 0.01075 & 0.908 & 4.0 & 9.0 & 9.0 & 8.3 & 9.0 & 7.8 \\
\addlinespace
\multicolumn{9}{l}{\textbf{CheXpert}} \\
$\epsilon\le1$ & 0.00111 & 1.123 & 4.0 & 1.0 & 1.0 & 1.0 & 0.7 & 0.5 \\
$\epsilon\le3$ & 0.00111 & 0.721 & 4.0 & 3.0 & 3.0 & 3.0 & 2.1 & 1.6 \\
$\epsilon\le6$ & 0.00111 & 0.622 & 4.0 & 6.0 & 5.8 & 6.0 & 3.4 & 2.5 \\
$\epsilon\le9$ & 0.00111 & 0.571 & 4.0 & 9.0 & 8.7 & 8.7 & 4.6 & 3.2 \\
\addlinespace
\multicolumn{9}{l}{\textbf{ChestX-ray14}} \\
$\epsilon\le1$ & 0.00164 & 1.294 & 4.0 & 1.0 & 1.0 & 0.9 & 1.0 & 0.9 \\
$\epsilon\le3$ & 0.00164 & 0.787 & 4.0 & 2.7 & 2.9 & 2.5 & 3.0 & 2.8 \\
$\epsilon\le6$ & 0.00164 & 0.666 & 4.0 & 5.8 & 5.9 & 5.7 & 5.7 & 6.0 \\
$\epsilon\le9$ & 0.00164 & 0.606 & 4.0 & 8.0 & 8.9 & 7.9 & 7.8 & 8.3 \\
\addlinespace
\multicolumn{9}{l}{\textbf{PadChest}} \\
$\epsilon\le1$ & 0.00161 & 1.282 & 4.0 & 1.0 & 1.0 & 1.0 & 0.4 & 0.4 \\
$\epsilon\le3$ & 0.00161 & 0.782 & 4.0 & 3.0 & 3.0 & 2.9 & 3.0 & 2.7 \\
$\epsilon\le6$ & 0.00161 & 0.663 & 4.0 & 6.0 & 5.9 & 6.0 & 5.8 & 5.6 \\
$\epsilon\le9$ & 0.00161 & 0.604 & 4.0 & 8.7 & 9.0 & 8.9 & 8.6 & 7.7 \\
\addlinespace
\multicolumn{9}{l}{\textbf{UKA-CXR}} \\
$\epsilon\le1$ & 0.00093 & 1.057 & 4.0 & 1.0 & 1.0 & 0.9 & 1.0 & 1.0 \\
$\epsilon\le3$ & 0.00093 & 0.696 & 4.0 & 2.9 & 3.0 & 2.9 & 3.0 & 3.0 \\
$\epsilon\le6$ & 0.00093 & 0.605 & 4.0 & 5.9 & 6.0 & 5.3 & 6.0 & 6.0 \\
$\epsilon\le9$ & 0.00093 & 0.557 & 4.0 & 8.8 & 9.0 & 7.3 & 9.0 & 9.0 \\
\addlinespace
\bottomrule
\end{tabular}
\end{table}

\begin{table}[t]
\centering
\footnotesize
\caption{Optimization steps of the 100 differentially private runs of the main experiment. Each entry is the number of optimization steps taken up to the epoch whose model was kept, which is the epoch of highest validation macro AUROC. This count determines the achieved $\epsilon$ in Supplementary Table~\ref{stab:accounting}. Rows are grouped by dataset and ordered by privacy budget, from the strictest to the loosest, and columns are ordered from no pretraining to in-domain pretraining. The datasets are VinDr-CXR~\cite{Nguyen2022VinDr}, CheXpert~\cite{Irvin2019CheXpert}, ChestX-ray14~\cite{Wang2017ChestXray8}, PadChest~\cite{Bustos2020PadChest}, and UKA-CXR~\cite{Khader2022Artificial}. AUROC, area under the receiver operating characteristic curve; $\epsilon$, privacy budget; He, He initialization~\cite{He2015Delving}; ImageNet, supervised pretraining on ImageNet~\cite{Deng2009ImageNet}; DINOv3, self-supervised DINOv3 pretraining~\cite{Simeoni2025DINOv3}; InDomain-stand and InDomain-priv, supervised pretraining on MIMIC-CXR~\cite{Johnson2019MIMIC} without and with differential privacy.}
\label{stab:steps}
\scriptsize
\setlength{\tabcolsep}{5pt}
\begin{tabular}{lccccc}
\toprule
Privacy budget & He & ImageNet & DINOv3 & InDomain-stand & InDomain-priv \\
\midrule
\multicolumn{6}{l}{\textbf{VinDr-CXR}} \\
$\epsilon\le1$ & 3{,}772 & 3{,}772 & 3{,}772 & 184 & 3{,}680 \\
$\epsilon\le3$ & 5{,}428 & 5{,}428 & 5{,}060 & 92 & 5{,}428 \\
$\epsilon\le6$ & 8{,}188 & 8{,}188 & 7{,}452 & 7{,}544 & 7{,}360 \\
$\epsilon\le9$ & 9{,}568 & 9{,}660 & 8{,}372 & 9{,}660 & 7{,}360 \\
\addlinespace
\multicolumn{6}{l}{\textbf{CheXpert}} \\
$\epsilon\le1$ & 32{,}472 & 36{,}080 & 36{,}080 & 12{,}628 & 902 \\
$\epsilon\le3$ & 45{,}100 & 45{,}100 & 44{,}198 & 12{,}628 & 902 \\
$\epsilon\le6$ & 73{,}964 & 68{,}552 & 73{,}062 & 12{,}628 & 902 \\
$\epsilon\le9$ & 90{,}200 & 83{,}886 & 82{,}984 & 12{,}628 & 902 \\
\addlinespace
\multicolumn{6}{l}{\textbf{ChestX-ray14}} \\
$\epsilon\le1$ & 25{,}536 & 24{,}928 & 21{,}280 & 25{,}536 & 20{,}672 \\
$\epsilon\le3$ & 25{,}536 & 32{,}224 & 21{,}280 & 33{,}440 & 28{,}576 \\
$\epsilon\le6$ & 48{,}032 & 49{,}856 & 46{,}816 & 45{,}600 & 51{,}680 \\
$\epsilon\le9$ & 48{,}032 & 60{,}192 & 46{,}816 & 45{,}600 & 51{,}680 \\
\addlinespace
\multicolumn{6}{l}{\textbf{PadChest}} \\
$\epsilon\le1$ & 23{,}636 & 24{,}258 & 24{,}880 & 622 & 622 \\
$\epsilon\le3$ & 33{,}588 & 34{,}210 & 30{,}478 & 34{,}210 & 26{,}124 \\
$\epsilon\le6$ & 52{,}248 & 51{,}626 & 52{,}248 & 49{,}760 & 45{,}406 \\
$\epsilon\le9$ & 59{,}090 & 62{,}200 & 60{,}956 & 56{,}602 & 45{,}406 \\
\addlinespace
\multicolumn{6}{l}{\textbf{UKA-CXR}} \\
$\epsilon\le1$ & 37{,}695 & 38{,}772 & 31{,}233 & 36{,}618 & 37{,}695 \\
$\epsilon\le3$ & 47{,}388 & 49{,}542 & 46{,}311 & 48{,}465 & 49{,}542 \\
$\epsilon\le6$ & 81{,}852 & 87{,}237 & 63{,}543 & 87{,}237 & 87{,}237 \\
$\epsilon\le9$ & 101{,}238 & 105{,}546 & 63{,}543 & 106{,}623 & 106{,}623 \\
\addlinespace
\bottomrule
\end{tabular}
\end{table}

\begin{table}[t]
\centering
\footnotesize
\caption{Leave-one-dataset-out generalization to unseen institutions ($n = 120{,}068$ test radiographs). Each dataset was held out in turn, a ConvNeXt-Small~\cite{Liu2022ConvNet} was trained on the pooled training data of the other four, and macro AUROC in percent was measured on the test set of the held-out dataset. Values are the bootstrap mean $\pm$ standard deviation with the 95\% CI in brackets from 10{,}000 resamples. Rows are grouped by privacy setting, and columns are ordered from no pretraining to in-domain pretraining. The held-out test sets are those of Table~\ref{tab:frontier}, from VinDr-CXR~\cite{Nguyen2022VinDr}, CheXpert~\cite{Irvin2019CheXpert}, ChestX-ray14~\cite{Wang2017ChestXray8}, PadChest~\cite{Bustos2020PadChest}, and UKA-CXR~\cite{Khader2022Artificial}. AUROC, area under the receiver operating characteristic curve; CI, confidence interval; $\epsilon$, privacy budget; He, He initialization~\cite{He2015Delving}; ImageNet, supervised pretraining on ImageNet~\cite{Deng2009ImageNet}; DINOv3, self-supervised DINOv3 pretraining~\cite{Simeoni2025DINOv3}; InDomain-stand and InDomain-priv, supervised pretraining on MIMIC-CXR~\cite{Johnson2019MIMIC} without and with differential privacy.}
\label{stab:lodo}
\scriptsize
\begin{tabular}{lccccc}
\toprule
Held-out dataset & He & ImageNet & DINOv3 & InDomain-stand & InDomain-priv \\
\midrule
\multicolumn{6}{l}{\textbf{$\epsilon\le1$}} \\
VinDr-CXR & \begin{tabular}[t]{@{}c@{}}51.4 $\pm$ 1.2\\{[}49.1, 53.6{]}\end{tabular} & \begin{tabular}[t]{@{}c@{}}61.0 $\pm$ 1.1\\{[}58.8, 63.1{]}\end{tabular} & \begin{tabular}[t]{@{}c@{}}73.6 $\pm$ 1.2\\{[}71.2, 75.8{]}\end{tabular} & \begin{tabular}[t]{@{}c@{}}88.6 $\pm$ 0.6\\{[}87.3, 89.8{]}\end{tabular} & \begin{tabular}[t]{@{}c@{}}79.6 $\pm$ 0.9\\{[}77.9, 81.3{]}\end{tabular} \\
CheXpert & \begin{tabular}[t]{@{}c@{}}52.0 $\pm$ 0.3\\{[}51.5, 52.6{]}\end{tabular} & \begin{tabular}[t]{@{}c@{}}62.6 $\pm$ 0.3\\{[}62.0, 63.2{]}\end{tabular} & \begin{tabular}[t]{@{}c@{}}66.4 $\pm$ 0.3\\{[}65.8, 67.0{]}\end{tabular} & \begin{tabular}[t]{@{}c@{}}74.8 $\pm$ 0.3\\{[}74.2, 75.4{]}\end{tabular} & \begin{tabular}[t]{@{}c@{}}69.3 $\pm$ 0.3\\{[}68.7, 69.9{]}\end{tabular} \\
ChestX-ray14 & \begin{tabular}[t]{@{}c@{}}50.3 $\pm$ 0.4\\{[}49.4, 51.2{]}\end{tabular} & \begin{tabular}[t]{@{}c@{}}58.4 $\pm$ 0.6\\{[}57.2, 59.5{]}\end{tabular} & \begin{tabular}[t]{@{}c@{}}61.3 $\pm$ 0.6\\{[}60.1, 62.5{]}\end{tabular} & \begin{tabular}[t]{@{}c@{}}73.0 $\pm$ 0.5\\{[}72.0, 74.0{]}\end{tabular} & \begin{tabular}[t]{@{}c@{}}65.6 $\pm$ 0.6\\{[}64.4, 66.7{]}\end{tabular} \\
PadChest & \begin{tabular}[t]{@{}c@{}}51.1 $\pm$ 0.4\\{[}50.3, 51.9{]}\end{tabular} & \begin{tabular}[t]{@{}c@{}}65.2 $\pm$ 0.4\\{[}64.4, 65.9{]}\end{tabular} & \begin{tabular}[t]{@{}c@{}}73.3 $\pm$ 0.3\\{[}72.7, 74.0{]}\end{tabular} & \begin{tabular}[t]{@{}c@{}}80.0 $\pm$ 0.3\\{[}79.4, 80.7{]}\end{tabular} & \begin{tabular}[t]{@{}c@{}}76.7 $\pm$ 0.3\\{[}76.1, 77.3{]}\end{tabular} \\
UKA-CXR & \begin{tabular}[t]{@{}c@{}}50.0 $\pm$ 0.3\\{[}49.4, 50.6{]}\end{tabular} & \begin{tabular}[t]{@{}c@{}}67.4 $\pm$ 0.3\\{[}66.8, 68.0{]}\end{tabular} & \begin{tabular}[t]{@{}c@{}}61.7 $\pm$ 0.2\\{[}61.2, 62.1{]}\end{tabular} & \begin{tabular}[t]{@{}c@{}}66.2 $\pm$ 0.2\\{[}65.8, 66.6{]}\end{tabular} & \begin{tabular}[t]{@{}c@{}}61.6 $\pm$ 0.2\\{[}61.2, 62.1{]}\end{tabular} \\
\addlinespace
\multicolumn{6}{l}{\textbf{Non-private}} \\
VinDr-CXR & \begin{tabular}[t]{@{}c@{}}73.9 $\pm$ 1.0\\{[}71.9, 75.9{]}\end{tabular} & \begin{tabular}[t]{@{}c@{}}89.1 $\pm$ 0.5\\{[}88.1, 90.2{]}\end{tabular} & \begin{tabular}[t]{@{}c@{}}90.1 $\pm$ 0.5\\{[}89.1, 91.0{]}\end{tabular} & \begin{tabular}[t]{@{}c@{}}90.1 $\pm$ 0.5\\{[}89.1, 91.0{]}\end{tabular} & \begin{tabular}[t]{@{}c@{}}89.0 $\pm$ 0.5\\{[}87.9, 90.0{]}\end{tabular} \\
CheXpert & \begin{tabular}[t]{@{}c@{}}67.3 $\pm$ 0.3\\{[}66.7, 67.9{]}\end{tabular} & \begin{tabular}[t]{@{}c@{}}75.2 $\pm$ 0.3\\{[}74.6, 75.7{]}\end{tabular} & \begin{tabular}[t]{@{}c@{}}76.6 $\pm$ 0.3\\{[}76.0, 77.2{]}\end{tabular} & \begin{tabular}[t]{@{}c@{}}77.6 $\pm$ 0.3\\{[}77.1, 78.2{]}\end{tabular} & \begin{tabular}[t]{@{}c@{}}77.1 $\pm$ 0.3\\{[}76.5, 77.6{]}\end{tabular} \\
ChestX-ray14 & \begin{tabular}[t]{@{}c@{}}63.7 $\pm$ 0.6\\{[}62.5, 64.7{]}\end{tabular} & \begin{tabular}[t]{@{}c@{}}75.8 $\pm$ 0.5\\{[}74.9, 76.7{]}\end{tabular} & \begin{tabular}[t]{@{}c@{}}77.9 $\pm$ 0.4\\{[}77.1, 78.7{]}\end{tabular} & \begin{tabular}[t]{@{}c@{}}77.7 $\pm$ 0.4\\{[}76.9, 78.5{]}\end{tabular} & \begin{tabular}[t]{@{}c@{}}75.6 $\pm$ 0.5\\{[}74.7, 76.5{]}\end{tabular} \\
PadChest & \begin{tabular}[t]{@{}c@{}}74.7 $\pm$ 0.3\\{[}74.0, 75.3{]}\end{tabular} & \begin{tabular}[t]{@{}c@{}}81.3 $\pm$ 0.3\\{[}80.7, 81.9{]}\end{tabular} & \begin{tabular}[t]{@{}c@{}}81.7 $\pm$ 0.3\\{[}81.1, 82.3{]}\end{tabular} & \begin{tabular}[t]{@{}c@{}}83.5 $\pm$ 0.3\\{[}82.9, 84.0{]}\end{tabular} & \begin{tabular}[t]{@{}c@{}}80.7 $\pm$ 0.3\\{[}80.1, 81.3{]}\end{tabular} \\
UKA-CXR & \begin{tabular}[t]{@{}c@{}}64.0 $\pm$ 0.3\\{[}63.5, 64.6{]}\end{tabular} & \begin{tabular}[t]{@{}c@{}}71.3 $\pm$ 0.2\\{[}70.8, 71.8{]}\end{tabular} & \begin{tabular}[t]{@{}c@{}}71.4 $\pm$ 0.3\\{[}70.9, 71.9{]}\end{tabular} & \begin{tabular}[t]{@{}c@{}}72.6 $\pm$ 0.2\\{[}72.1, 73.0{]}\end{tabular} & \begin{tabular}[t]{@{}c@{}}70.2 $\pm$ 0.3\\{[}69.7, 70.7{]}\end{tabular} \\
\addlinespace
\bottomrule
\end{tabular}
\end{table}

\begin{table}[t]
\centering
\footnotesize
\caption{Effect of backbone capacity on macro AUROC in percent on two test sets ($n = 32{,}321$ test radiographs). ConvNeXt-Tiny~\cite{Liu2022ConvNet} has 27.8 million parameters and ConvNeXt-Small 49.5 million. Values are the bootstrap mean $\pm$ standard deviation with the 95\% CI in brackets from 10{,}000 resamples, and $\Delta$ is ConvNeXt-Small minus ConvNeXt-Tiny. The two in-domain initializations exist for ConvNeXt-Small only. The comparison therefore covers the three other initializations. The test sets comprise 3{,}000 (VinDr-CXR~\cite{Nguyen2022VinDr}) and 29{,}321 (CheXpert~\cite{Irvin2019CheXpert}) radiographs. AUROC, area under the receiver operating characteristic curve; CI, confidence interval; $\epsilon$, privacy budget; He, He initialization~\cite{He2015Delving}; ImageNet, supervised pretraining on ImageNet~\cite{Deng2009ImageNet}; DINOv3, self-supervised DINOv3 pretraining~\cite{Simeoni2025DINOv3}.}
\label{stab:capacity}
\small
\begin{tabular}{lccc}
\toprule
Initialization & ConvNeXt-Tiny & ConvNeXt-Small & $\Delta$ \\
\midrule
\multicolumn{4}{l}{\textbf{VinDr-CXR, $\epsilon\le1$}} \\
He & \begin{tabular}[t]{@{}c@{}}50.9 $\pm$ 1.1\\{[}48.6, 53.1{]}\end{tabular} & \begin{tabular}[t]{@{}c@{}}52.2 $\pm$ 1.2\\{[}49.8, 54.5{]}\end{tabular} & +1.3 \\
ImageNet & \begin{tabular}[t]{@{}c@{}}62.9 $\pm$ 1.1\\{[}60.7, 65.0{]}\end{tabular} & \begin{tabular}[t]{@{}c@{}}59.4 $\pm$ 1.0\\{[}57.3, 61.5{]}\end{tabular} & -3.5 \\
DINOv3 & \begin{tabular}[t]{@{}c@{}}71.3 $\pm$ 1.2\\{[}68.9, 73.6{]}\end{tabular} & \begin{tabular}[t]{@{}c@{}}77.9 $\pm$ 1.2\\{[}75.5, 80.3{]}\end{tabular} & +6.6 \\
\addlinespace
\multicolumn{4}{l}{\textbf{VinDr-CXR, non-private}} \\
He & \begin{tabular}[t]{@{}c@{}}50.5 $\pm$ 1.1\\{[}48.3, 52.5{]}\end{tabular} & \begin{tabular}[t]{@{}c@{}}51.5 $\pm$ 1.2\\{[}49.2, 53.7{]}\end{tabular} & +1.0 \\
ImageNet & \begin{tabular}[t]{@{}c@{}}88.7 $\pm$ 0.7\\{[}87.2, 90.0{]}\end{tabular} & \begin{tabular}[t]{@{}c@{}}88.5 $\pm$ 0.8\\{[}86.9, 90.0{]}\end{tabular} & -0.2 \\
DINOv3 & \begin{tabular}[t]{@{}c@{}}87.4 $\pm$ 0.8\\{[}85.7, 89.0{]}\end{tabular} & \begin{tabular}[t]{@{}c@{}}90.2 $\pm$ 0.7\\{[}88.7, 91.6{]}\end{tabular} & +2.8 \\
\addlinespace
\multicolumn{4}{l}{\textbf{CheXpert, $\epsilon\le1$}} \\
He & \begin{tabular}[t]{@{}c@{}}50.9 $\pm$ 0.3\\{[}50.3, 51.5{]}\end{tabular} & \begin{tabular}[t]{@{}c@{}}52.9 $\pm$ 0.3\\{[}52.3, 53.4{]}\end{tabular} & +2.0 \\
ImageNet & \begin{tabular}[t]{@{}c@{}}66.5 $\pm$ 0.3\\{[}65.9, 67.1{]}\end{tabular} & \begin{tabular}[t]{@{}c@{}}67.2 $\pm$ 0.3\\{[}66.6, 67.8{]}\end{tabular} & +0.7 \\
DINOv3 & \begin{tabular}[t]{@{}c@{}}69.9 $\pm$ 0.3\\{[}69.3, 70.5{]}\end{tabular} & \begin{tabular}[t]{@{}c@{}}68.9 $\pm$ 0.3\\{[}68.3, 69.4{]}\end{tabular} & -1.0 \\
\addlinespace
\multicolumn{4}{l}{\textbf{CheXpert, non-private}} \\
He & \begin{tabular}[t]{@{}c@{}}64.2 $\pm$ 0.3\\{[}63.6, 64.9{]}\end{tabular} & \begin{tabular}[t]{@{}c@{}}64.9 $\pm$ 0.3\\{[}64.3, 65.5{]}\end{tabular} & +0.7 \\
ImageNet & \begin{tabular}[t]{@{}c@{}}81.8 $\pm$ 0.2\\{[}81.3, 82.3{]}\end{tabular} & \begin{tabular}[t]{@{}c@{}}81.8 $\pm$ 0.3\\{[}81.3, 82.3{]}\end{tabular} & +0.0 \\
DINOv3 & \begin{tabular}[t]{@{}c@{}}82.5 $\pm$ 0.2\\{[}82.0, 83.0{]}\end{tabular} & \begin{tabular}[t]{@{}c@{}}83.0 $\pm$ 0.2\\{[}82.5, 83.4{]}\end{tabular} & +0.5 \\
\addlinespace
\bottomrule
\end{tabular}
\end{table}

\begin{table}[t]
\centering
\footnotesize
\caption{Composition of the demographic subgroups of the five test sets ($n = 120{,}068$ test radiographs). For every subgroup, the table gives the number of radiographs and the prevalence in percent of each finding within it. The age bands are from above 0 to below 40, from 40 to below 70, and from 70 to below 100 years. For CheXpert, the prevalence of atelectasis and pleural effusion is computed over the radiographs with a label for that finding. The heading row of each dataset lists the number of test radiographs without a recorded sex and the number with an age outside the three age bands. These radiographs belong to no subgroup of that attribute. Rows are ordered by dataset and within a dataset by sex and then by age band. The datasets are VinDr-CXR~\cite{Nguyen2022VinDr}, CheXpert~\cite{Irvin2019CheXpert}, ChestX-ray14~\cite{Wang2017ChestXray8}, PadChest~\cite{Bustos2020PadChest}, and UKA-CXR~\cite{Khader2022Artificial}.}
\label{stab:demographics}
\scriptsize
\begin{tabular}{lrccccc}
\toprule
Subgroup & Radiographs & Atelectasis & Cardiomegaly & Pleural effusion & Pneumonia & No finding \\
\midrule
\multicolumn{7}{l}{\textbf{VinDr-CXR (1{,}746 without a recorded sex, 2{,}532 outside the age bands)}} \\
Male & 702 & 4 & 8 & 6 & 11 & 56 \\
Female & 552 & 3 & 22 & 4 & 7 & 56 \\
Age $<$40 & 149 & 1 & 4 & 1 & 5 & 84 \\
Age 40 to $<$70 & 250 & 6 & 15 & 4 & 10 & 46 \\
Age 70 to $<$100 & 69 & 4 & 46 & 6 & 13 & 6 \\
\addlinespace
\multicolumn{7}{l}{\textbf{CheXpert (0 without a recorded sex, 0 outside the age bands)}} \\
Male & 17{,}885 & 18 & 13 & 41 & 3 & 12 \\
Female & 11{,}436 & 17 & 13 & 41 & 3 & 12 \\
Age $<$40 & 4{,}391 & 12 & 10 & 28 & 3 & 23 \\
Age 40 to $<$70 & 16{,}421 & 18 & 12 & 40 & 2 & 12 \\
Age 70 to $<$100 & 8{,}509 & 19 & 18 & 49 & 3 & 6 \\
\addlinespace
\multicolumn{7}{l}{\textbf{ChestX-ray14 (0 without a recorded sex, 1 outside the age bands)}} \\
Male & 14{,}882 & 13 & 4 & 18 & 2 & 38 \\
Female & 10{,}714 & 13 & 5 & 18 & 2 & 39 \\
Age $<$40 & 8{,}410 & 10 & 5 & 16 & 3 & 41 \\
Age 40 to $<$70 & 15{,}495 & 14 & 4 & 19 & 2 & 37 \\
Age 70 to $<$100 & 1{,}690 & 19 & 5 & 26 & 1 & 35 \\
\addlinespace
\multicolumn{7}{l}{\textbf{PadChest (1 without a recorded sex, 142 outside the age bands)}} \\
Male & 11{,}408 & 6 & 8 & 8 & 5 & 27 \\
Female & 10{,}636 & 5 & 10 & 5 & 4 & 38 \\
Age $<$40 & 3{,}612 & 2 & 1 & 3 & 8 & 60 \\
Age 40 to $<$70 & 10{,}788 & 5 & 5 & 5 & 3 & 38 \\
Age 70 to $<$100 & 7{,}503 & 8 & 18 & 10 & 5 & 13 \\
\addlinespace
\multicolumn{7}{l}{\textbf{UKA-CXR (0 without a recorded sex, 15 outside the age bands)}} \\
Male & 25{,}536 & 14 & 51 & 13 & 16 & 35 \\
Female & 14{,}570 & 13 & 40 & 13 & 13 & 44 \\
Age $<$40 & 2{,}325 & 14 & 28 & 10 & 17 & 51 \\
Age 40 to $<$70 & 19{,}383 & 14 & 43 & 12 & 16 & 41 \\
Age 70 to $<$100 & 18{,}383 & 14 & 53 & 14 & 13 & 34 \\
\addlinespace
\bottomrule
\end{tabular}
\end{table}

\begin{table}[t]
\centering
\footnotesize
\caption{Size and composition of the five datasets after projection filtering ($n = 591{,}884$ radiographs). Columns give the number of radiographs in total and in the training, validation, and test splits, the number of unique patients, and the projections retained. VinDr-CXR records no patient identifiers, and its validation split is 20\% of its official training split. The datasets are VinDr-CXR~\cite{Nguyen2022VinDr}, CheXpert~\cite{Irvin2019CheXpert}, ChestX-ray14~\cite{Wang2017ChestXray8}, PadChest~\cite{Bustos2020PadChest}, and UKA-CXR~\cite{Khader2022Artificial}. AP, anteroposterior; N/A, not applicable; PA, posteroanterior.}
\label{stab:datasets}
\setlength{\tabcolsep}{6pt}
\renewcommand{\arraystretch}{1.2}
\footnotesize
\begin{tabular}{@{}l r r r r r l@{}}
\toprule
Dataset & Total & Training & Validation & Test & Patients & Projections \\
\midrule
VinDr-CXR & 18{,}000 & 12{,}000 & 3{,}000 & 3{,}000 & N/A & All \\
CheXpert & 157{,}878 & 115{,}458 & 13{,}099 & 29{,}321 & 57{,}872 & Frontal \\
ChestX-ray14 & 112{,}120 & 77{,}870 & 8{,}654 & 25{,}596 & 30{,}805 & All \\
PadChest & 110{,}525 & 79{,}697 & 8{,}783 & 22{,}045 & 67{,}205 & PA, AP, horizontal AP \\
UKA-CXR & 193{,}361 & 137{,}902 & 15{,}353 & 40{,}106 & 54{,}176 & All \\
\addlinespace
All five datasets & 591{,}884 & & & & & \\
\bottomrule
\end{tabular}
\end{table}

\begin{table}[t]
\centering
\footnotesize
\caption{Number of positive radiographs of each finding in every split of the five datasets ($n = 591{,}884$ radiographs), with the percentage of the split in parentheses. The findings are not mutually exclusive. The percentages of a split therefore do not sum to 100. The training counts set the positive weights of the loss in Eq.~\ref{eq:posweight}. For CheXpert, the percentage of atelectasis and pleural effusion is taken over the radiographs with a label for that finding. The datasets are VinDr-CXR~\cite{Nguyen2022VinDr}, CheXpert~\cite{Irvin2019CheXpert}, ChestX-ray14~\cite{Wang2017ChestXray8}, PadChest~\cite{Bustos2020PadChest}, and UKA-CXR~\cite{Khader2022Artificial}.}
\label{stab:prevalence}
\scriptsize
\begin{tabular}{lrccccc}
\toprule
Split & Radiographs & Atelectasis & Cardiomegaly & Pleural effusion & Pneumonia & No finding \\
\midrule
\multicolumn{7}{l}{\textbf{VinDr-CXR}} \\
Training & 12{,}000 & 51 ($<$1) & 1{,}453 (12) & 507 (4) & 377 (3) & 8{,}481 (71) \\
Validation & 3{,}000 & 11 ($<$1) & 364 (12) & 127 (4) & 94 (3) & 2{,}120 (71) \\
Test & 3{,}000 & 86 (3) & 309 (10) & 111 (4) & 246 (8) & 2{,}051 (68) \\
\addlinespace
\multicolumn{7}{l}{\textbf{CheXpert}} \\
Training & 115{,}458 & 19{,}488 (19) & 14{,}317 (12) & 48{,}439 (44) & 2{,}851 (2) & 12{,}088 (10) \\
Validation & 13{,}099 & 2{,}302 (20) & 1{,}629 (12) & 5{,}265 (42) & 297 (2) & 1{,}372 (10) \\
Test & 29{,}321 & 4{,}523 (18) & 3{,}944 (13) & 11{,}438 (41) & 816 (3) & 3{,}540 (12) \\
\addlinespace
\multicolumn{7}{l}{\textbf{ChestX-ray14}} \\
Training & 77{,}870 & 7{,}392 (9) & 1{,}549 (2) & 7{,}833 (10) & 782 (1) & 45{,}415 (58) \\
Validation & 8{,}654 & 888 (10) & 158 (2) & 826 (10) & 94 (1) & 5{,}085 (59) \\
Test & 25{,}596 & 3{,}279 (13) & 1{,}069 (4) & 4{,}658 (18) & 555 (2) & 9{,}861 (39) \\
\addlinespace
\multicolumn{7}{l}{\textbf{PadChest}} \\
Training & 79{,}697 & 4{,}413 (6) & 7{,}083 (9) & 5{,}071 (6) & 3{,}793 (5) & 26{,}074 (33) \\
Validation & 8{,}783 & 513 (6) & 808 (9) & 540 (6) & 437 (5) & 2{,}858 (33) \\
Test & 22{,}045 & 1{,}240 (6) & 1{,}954 (9) & 1{,}373 (6) & 992 (4) & 7{,}216 (33) \\
\addlinespace
\multicolumn{7}{l}{\textbf{UKA-CXR}} \\
Training & 137{,}902 & 19{,}063 (14) & 64{,}447 (47) & 17{,}372 (13) & 18{,}976 (14) & 53{,}233 (39) \\
Validation & 15{,}353 & 2{,}260 (15) & 7{,}143 (47) & 1{,}950 (13) & 2{,}301 (15) & 5{,}858 (38) \\
Test & 40{,}106 & 5{,}650 (14) & 18{,}758 (47) & 5{,}107 (13) & 5{,}895 (15) & 15{,}364 (38) \\
\addlinespace
\bottomrule
\end{tabular}
\end{table}

\begin{table}[t]
\centering
\footnotesize
\caption{Training cost of the 125 runs of the main experiment. For each dataset, the table gives the number of training radiographs, the mean time of one epoch without privacy and under DP-SGD~\cite{Abadi2016Deep}, the ratio of the two, and the total training time summed over the 25 runs of that dataset. The epoch time of a run is the mean of its epoch times in the most populated window of plus or minus 5\% of their median. The last row gives the total number of training radiographs and the total training time of the 125 runs. The datasets are VinDr-CXR~\cite{Nguyen2022VinDr}, CheXpert~\cite{Irvin2019CheXpert}, ChestX-ray14~\cite{Wang2017ChestXray8}, PadChest~\cite{Bustos2020PadChest}, and UKA-CXR~\cite{Khader2022Artificial}. DP-SGD, differentially private stochastic gradient descent.}
\label{stab:compute}
\small
\begin{tabular}{lrcccr}
\toprule
Dataset & Training radiographs & Non-private (min) & Private (min) & Ratio & Total (h) \\
\midrule
VinDr-CXR & 12{,}000 & 0.9 & 1.9 & 2.12 & 49 \\
CheXpert & 115{,}458 & 8.6 & 18.5 & 2.16 & 439 \\
ChestX-ray14 & 77{,}870 & 6.1 & 12.8 & 2.08 & 317 \\
PadChest & 79{,}697 & 6.3 & 13.2 & 2.08 & 325 \\
UKA-CXR & 137{,}902 & 10.8 & 22.6 & 2.08 & 486 \\
\addlinespace
All five datasets & 422{,}927 & & & & 1{,}616 \\
\bottomrule
\end{tabular}
\end{table}

\end{document}